%% file: main.tex
\documentclass[10pt,twocolumn,letterpaper]{article}

\usepackage{iccv}      %

\usepackage{graphicx}
\usepackage{times}
\usepackage{epsfig}
\usepackage{graphicx}
\usepackage{amsmath}
\usepackage{amssymb}
\usepackage[usenames,dvipsnames]{xcolor}

\usepackage{array}
\usepackage{xspace}
\usepackage{enumitem}
\usepackage{float}

\usepackage{dblfloatfix}

\usepackage{caption}
\usepackage{subcaption}

\usepackage[breaklinks=true,letterpaper=true,colorlinks,bookmarks=false]{hyperref}
\usepackage{makecell}
\usepackage{placeins}
\usepackage[capitalize]{cleveref}
\crefname{section}{Sec.}{Secs.}
\Crefname{section}{Section}{Sections}
\Crefname{table}{Table}{Tables}
\crefname{table}{Tab.}{Tabs.}

\usepackage{multirow}

\newcommand{\inner}[1]{\left\langle#1\right\rangle}

\def\R{\mathbb{R}}

\newcommand{\norm}[1]{\left\|#1\right\|}

\def\ones{\mathop{\rm 1}\nolimits}

\def\argmax{\mathop{\rm arg\,max}\limits}%

\def\ones{\mathbf{1}}

\usepackage[normalem]{ulem}

\def\argmax{\mathop{\rm argmax}\nolimits}

\def\red{\color{BrickRed}}
\def\blue{\color{blue}}

\def\change#1{\textcolor{black}{#1}}

\iccvfinalcopy

\def\NPFV{\textrm{NPFV}}
\def\XXX{322 } %

\def\iccvPaperID{10797} %

\newcolumntype{?}[1]{!{\vrule width #1}}

\ificcvfinal\fi

\date{October 2022}

\begin{document}
\title{Spurious Features Everywhere - Large-Scale Detection of Harmful Spurious Features in ImageNet}

\author{Yannic Neuhaus %
\and 
Maximilian Augustin%
\and 
Valentyn Boreiko %
\and 
Matthias Hein\and 
T\"ubingen AI Center --
University of T\"ubingen
}
\maketitle

\begin{abstract}
Benchmark performance of deep learning classifiers alone is not a reliable predictor for the performance of a deployed model. In particular, if the image classifier has picked up spurious features in the training data, its predictions can fail in unexpected ways. In this paper, we develop a framework that allows us to systematically identify spurious features in large datasets like ImageNet. It is based on our neural PCA components and their visualization. Previous work on spurious features often operates in toy settings or requires costly pixel-wise annotations.  In contrast, we work with ImageNet and validate our results by showing that presence of the harmful spurious feature of a class alone is sufficient to trigger the prediction of that class. We introduce the novel dataset ``Spurious ImageNet'' which allows to measure the reliance of any ImageNet classifier on harmful spurious features. Moreover, we introduce SpuFix as a simple mitigation method to reduce the dependence of any ImageNet classifier on previously identified harmful spurious features without requiring additional labels or retraining of the model. We provide code and data at \url{https://github.com/YanNeu/spurious_imagenet}. %
\end{abstract}

\section{Introduction}
\label{sec:intro}

\thispagestyle{empty} %

\input{teaser_figure.tex}
Deep learning has led to tremendous progress in image classification \cite{KurGooBen2016a,recht2019imagenet} and natural language processing \cite{Brown2020GPT3}. %
Over the years, however, it has become apparent, that evaluating predictive performance on a fixed test set is not necessarily indicative of the performance when image classifiers are deployed in the wild. Several potential failure cases have been discovered. This starts with a lack of robustness due to image corruptions \cite{HenDie2019}, adversarial perturbations \cite{SzeEtAl2014}, and arbitrary predictions on out-of-distribution inputs \cite{NguYosClu2015,HenGim2017,HeiAndBit2019}. In this paper, we consider the problem of identifying and debugging image classifiers from spurious features \cite{Adebayo2020}. Spurious features in image classification are features that co-occur with the actual class object and are picked up by the classifier.
In the worst case, they lead to shortcut learning \cite{Geirhos2020shortcut}, where only the spurious but not the correct feature is associated with the class, e.g., \cite{Zech2018variable} found that a pneumonia detector's bad generalization across hospitals was 
caused by the neural network learning to identify the hospital where the training data originated from.
A weaker form of spurious feature (at least from a learning perspective) is the case when the classifier picks up the correct class features, e.g., of a hummingbird, but additionally associates a spurious feature, e.g., a bird feeder, with the class as {\change{they appear together on a subset of the training set}}. %
This becomes a harmful spurious feature if \emph{only} the spurious feature \emph{without} the class feature is sufficient to trigger the classification of that class, see Fig.~\ref{fig:teaser} for an illustration of such spurious features found via our method. Harmful spurious features are difficult to detect
and thus can easily go unnoticed, leading to unexpected behaviour of a deployed image classifier.\newline
\noindent In this paper we make the following key contributions:
\begin{itemize}[leftmargin=*,nosep]
\item
we develop a pipeline for the detection of harmful spurious features with little human supervision based on our class-wise neural PCA (NPCA) components of an adversarially robust classifier together with their  Neural PCA Feature Visualization (\NPFV). 
\item  unlike prior work, which used masking images or pixel-wise annotations, we validate our found spurious features by using our NPCA components to find real images containing only the spurious feature but not the class object.
\item using these images we create the dataset ``Spurious ImageNet''  and propose a measure for dependence on spurious features. We do a large-scale evaluation of state-of-the-art (SOTA) ImageNet models. We show that the spurious features found for the robust model generalize to non-robust classifiers.
Moreover, we analyze the influence of different training setups, e.g. pre-training on ImageNet21k or larger datasets like LAION.
\item 
we develop SpuFix, a technique to mitigate the dependence on identified harmful spurious features without requiring new labels or retraining, and show how to transfer it to any ImageNet classifier. SpuFix consistently improves the dependence on harmful spurious features even for SOTA models with negligible impact on test accuracy.

\end{itemize}

\section{Related work}\label{sec:related}

When classifiers in safety-critical systems such as healthcare or autonomous driving are deployed in the wild  \cite{AmoEtAl2016},  it is important to discover potential failure cases before release. Prior work has focused on corruption \cite{HenDie2019}, adversarial robustness \cite{BigEtAl13,SzeEtAl2014,MadEtAl2018}, and out-of-distribution detection \cite{NguYosClu2015,HenGim2017,HeiAndBit2019}. There is less work on spurious features, although their potential harm might be higher.\\
\textbf{Spurious features:} It has been noted early on that classifiers %
show reliance on spurious features \cite{CHOI2012853} e.g., a cow on the beach is not recognized \cite{beery2018recognition} due to the missing spurious feature of grass. Other forms of spurious features have been reported in the classification of skin lesions \cite{Bissoto2020Debiasing}, pneumonia \cite{Zech2018variable}, traffic signs \cite{stock2018convnets}, and object recognition \cite{Zhu2017Object}.
Moreover, it has been shown that deep neural networks are biased towards texture \cite{GeiEtAl2019} and background context \cite{xiao2021noise}, see \cite{Geirhos2020shortcut} for an overview. \cite{Shah2020pitfalls} argues theoretically that spurious features are picked up due to a simplicity bias. 

Detection of spurious features has been achieved using human label-intense pixel-wise annotations \cite{Plumb2022finding,shetty2019not, singh2020don}. In \cite{Wong2021Leveraging}, they use sparsity regularization to enforce a more interpretable model and %
use it for finding
 spurious features. 
\cite{Anders2022finding} propose a complex pipeline to detect spurious features. While they scale to ImageNet, their analysis is limited to a few spurious features for a subset of 100 classes.  \cite{Singla2022salient,moayeri2022hard,singla2022core} do a search on full ImageNet based on class-weighted ``neural maps''. The neural maps are used to add noise to ``spurious'' resp. ``core'' features but no significant difference in classification performance is observed. 
It remains unclear if their found spurious features are harmful, that is the feature alone triggers the decision for that class. 

\noindent\textbf{Interpretability methods:} In recent years several interpretability methods %
have been proposed e.g., attribution methods such as GradCAM \cite{Selvaraju2017GradCAM}, Shapley values \cite{SHAPley_explanations}, Relevance Propagation \cite{BaeEtAl2010}, and  LIME \cite{Ribeiro-Lime}. The use of these methods for the detection of spurious features has been analyzed in \cite{Adebayo2020,Adebayo2022} with mixed success and it has been argued that interpretability methods are not robust \cite{Dombrowski2019, slack2020fooling, ghorbani2019interpretation}. However, attribution methods work better for robust classifiers due to more interpretable gradients \cite{etmann2019connection}. Another technique is counterfactual explanations \cite{wachter2018counterfactual, verma2020counterfactual} which are difficult to generate for images due to the similarity to adversarial examples \cite{SzeEtAl2014}. Thus visual counterfactual explanations are realized via manipulation of a latent space \cite{schutte2021using} or in image space \cite{santurkar2019image,augustin2020,boreiko2022sparse} for an adversarially robust classifier.  Visual counterfactuals for non-robust classifiers  using diffusion models \cite{ho2020denoising, song2021scorebased, 
ho2021classifierfree, dhariwal2021diffusion} have been proposed in \cite{Augustin2022Diffusion}.

\noindent\textbf{ImageNet:} ImageNet \cite{ILSVRC15} suffers from several shortcomings: apart from an inherent dataset bias \cite{torralba2011unbiased}, semantically overlapping or even identical class pairs were reported \cite{hooker2020selective,tsipras2020imagenet,beyer2020we}, e.g., two classes ``maillot'', ``sunglass'' vs ``sunglasses'', ``notebook'' vs ``laptop'' etc. We disregard such trivial cases of dataset contamination and focus on classes with harmful spurious features, in particular ones where only a small portion of the training set is contaminated.

\section{Spurious features}
\label{sec:spurious}
\begin{figure}[t]
    \centering
    \includegraphics[width=\columnwidth]{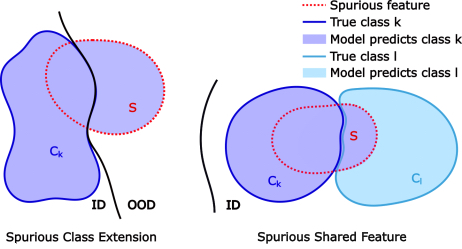}
    \caption{\label{fig:spurious-illustration}
    \textbf{Type of harmful Spurious Feature:} Left: The spurious feature $s$ is taken up by the classifier which predicts class $k$ on $C_k \cup S$ instead only on $C_k$. Right: The spurious feature $s$ is shared between classes but appears more often in class $k$. The classifier associates $S$ with class $k$ and thus predicts class $k$ also on $C_l \cap S$ instead of class $l$.}
    \vspace{-3mm}
\end{figure}

A proper definition of spurious features is difficult. We describe two settings of harmful spurious features which appear in this paper. We denote by $C_k$ the set of all images containing objects belonging to class $k$ (assuming for simplicity that we have a deterministic problem and ignoring multi-labels). Let $S$ be the set of all images containing a feature $s$ (e.g. a bird feeder). It is a \emph{correlated feature} for class $k$ when $C_k\cap S$ and $S \setminus C_k$ are non-empty, i.e. the feature occurs frequently with the class object but there is no causal implication that appearance of $s$ implies the appearance of the class object (a bird feeder in the image does not imply presence of a hummingbird). A \emph{correlated feature} becomes a \emph{spurious feature} when the classifier picks it up as feature of this class. Not every spurious feature is immediately harmful, even humans use context information \cite{Geirhos2020shortcut} to get more confident in a decision. However, a spurious feature is \emph{harmful} if the spurious feature alone is enough to trigger the decision for the corresponding class without the class object being present in the image. We consider two scenarios for a \emph{harmful spurious feature} shown in Fig.~\ref{fig:spurious-illustration}.

\indent \textbf{Spurious Class Extension:} For this type of spurious feature (left in Fig.~\ref{fig:spurious-illustration}) the classifier picks up the spurious feature $s$ for class $k$ and predicts the class $k$ even on $S \setminus C_k$ with high confidence (prediction of ``hummingbird'' for images showing a bird feeder but no hummingbird). The classifier predicts class $k$ beyond its actual domain $C_k$ and thus we call this a spurious class extension. While this spurious feature does not necessarily hurt in terms of test performance it can easily lead to completely unexpected behavior when the classifier is deployed in the wild.\newline
\indent\textbf{Spurious Shared Feature:} Here, two classes $C_k$ and $C_l$ share a spurious feature $s$ (e.g., ``water jet'' for the classes ``fireboat'' and ``fountain''). As there are more training images with feature $s$ in $C_k$ than in $C_l$ the classifier associates $S$ with class $k$ and predicts class $k$ for $S \cap C_l$.\newline
\indent The two types of harmful spurious features are not exclusive. A shared spurious feature $s$ can at the same time lead to a spurious class extension, 
e.g., the object bird feeder leads to a spurious class extension of the class ``hummingbird'' (see Fig.~\ref{fig:spurious-exp}) to images of bird feeders \textbf{without} hummingbirds. In the training set the  hummingbird feeder appears only in images of class ``hummingbird'' but the hummingbird feeder has parts which mimic flowers and flowers are a shared spurious feature with bees. In Fig.~\ref{fig:spurious-exp} right top row, images of bees on a bird feeder are classified as ``hummingbird'' instead of ``bee'', so the spurious feature is strong enough to override the decision for the true class ``bee'' (spurious shared feature).\newline

\section{Finding spurious features via neural PCA and associated feature visualizations}\label{sec:method}
First, we define our class-wise neural PCA (NPCA) which allows us to find diverse subpopulations in the training data, e.g., we checked that the bird feeder for ``hummingbird'' is visible in 15\% of the training images (component 2), while another 15\% contain a  part of it (component 3), see Fig.~\ref{fig:PCA-vs-TopFeature} {\change{or, for more examples, App.~\ref{app:qualitative}}}. Then we introduce our neural PCA feature visualization (which requires an adversarially robust model) and how we select NPCA components for human inspection. The identification of spurious features requires minimal human supervision, 
and our effective setup allows us to screen all ImageNet classes.

\textbf{Adversarially robust model:}
Similar to \cite{Singla2022salient}, we use an adversarially robust model to find spurious features in ImageNet. The reason for this is that robust models have generative properties \cite{TsiEtAl18,santurkar2019image,augustin2020,Singla2022salient,boreiko2022sparse} in the sense that maximizing the predicted probability of a class in a neighborhood of an image leads to semantically meaningful changes. They also have more informative gradients \cite{etmann2019connection} and thus attribution maps such as GradCAM \cite{Selvaraju2017GradCAM} work better. We use the multiple-norm robust model of \cite{boreiko2022sparse} as they claim it has the best generative properties. %
The generative properties of robust models are mainly used for the neural PCA feature visualization where we maximize the NPCA component of a class starting from a gray image. If a spurious feature appears without the class object, this is a strong indicator of a harmful spurious feature. A non-robust model would only produce semantically meaningless adversarial noise, hence we need a robust model for this part of our detection pipeline.
Our detected spurious features are not specific to the robust model. We show that SOTA ImageNet models share the same spurious features (Fig.~\ref{fig:spurious-exp} and Sec.~\ref{sec:exp-dataset}).%

\textbf{Class-wise neural PCA:} %
Let $(x_i,y_i)_{i=1}^N$ be the training set, where $y_i \in \{1,\ldots,K\}$ and $K$ is the number of classes.
We consider features of the penultimate layer $\phi(x) \in \R^D$ of a neural network for an input $x$. For a given class $k$ and its associated weights $w_k \in \R^D$ in the final layer, we define
\begin{equation}
\psi_k(x)=w_k \odot \phi(x).
\end{equation}
where $\odot$ is the componentwise product.
Let $b \in \R^K$ be the bias vector of the final layer then the logit $f_k$ of class $k$ is:
\begin{equation}\label{eq:logit}
f_k(x) = \sum\nolimits_{j=1}^D w_{kj} \,\phi(x)_j + b_k =\inner{\ones,\psi_k(x)} + b_k.
\end{equation}
Let $I_k$ be the index set of the training set of class $k$ and $\bar{\psi}_k$ the class-wise mean, $\bar{\psi}_k= \frac{1}{|I_k|}
    \sum\nolimits_{s \in I_k}\psi_k(x_s)$.
    The \emph{class-wise neural PCA} allows us to identify variations in the set $\{\psi(x_r)\}_{r \in I_k}$ arising due to small subpopulations in the training set. In the class-wise neural PCA, we compute eigenvectors of the class-wise covariance matrix,
\begin{equation} 
C  = \sum_{s \in I_k} \big(\psi_k(x_s) - \bar{\psi_k}\big) \big(\psi_k(x_s) - \bar{\psi_k}\big)^T.
\end{equation}
The eigenvectors, $v_1,\ldots,v_D$ form an orthonormal basis of $\R^D$ and we write $\psi_k(x)-\bar{\psi}_k$ in this basis, 
\begin{equation}
\psi_k(x)-\bar{\psi}_k=\sum\nolimits_{l=1}^D v_l\inner{\psi_k(x)-\bar{\psi}_k,v_l},
\end{equation}
and define
\begin{equation} \alpha^{(k)}_l(x) = \inner{\ones,v_l}\inner{\psi_k(x)-\bar{\psi}_k,v_l},
\end{equation}
The logit $f_k(x)$ of the $k$-th class can then be written as
\begin{equation} f_k(x) = \sum\nolimits_{l=1}^D  \alpha^{(k)}_l(x) + \inner{\ones,\bar{\psi}_k} + b_k.
\end{equation}
Thus for a given $x$, we can interpret $\alpha^{(k)}_l(x)$
as the contribution of the neural PCA component $l$ of class $k$ to the logit $f_k(x)$ of class $k$ since the term $\inner{\ones,\bar{\psi}_k} + b_k$ is constant for all inputs. Based on this we introduce a
mitigation technique for spurious features without retraining in Sec. \ref{sec:spufix}. 
\input{top5_compare_hummingbird_figure.tex}

\textbf{Neural PCA Feature Visualization:} %
To identify semantic features corresponding to our neural PCA component $l$%
, we show the training images which attain the maximal values of $\alpha^{(k)}_l(x)$. Additionally, we generate an image $z^{(k)}_l$, which we call the \textbf{Neural PCA Feature Visualization (NPFV)} of feature $l$ of class $k$,  by maximizing
$\alpha^{(k)}_l(x)$:
\[ z^{(k)}_l = g + \argmax_{\norm{\delta}_2\leq \epsilon} \,\alpha^{(k)}_l(g+\delta),\]
where $g$ is a gray image (all channels equal to $0.5$). Thus we maximize the feature $\alpha^{(k)}_l$ outgoing from a non-informative and unbiased initialization $g$.  The optimization problem is solved using adaptive projected gradient descent (APGD) \cite{croce2020reliable} with 200 steps. The budget for changes, $\epsilon=30$,  is small to avoid the overactivation of feature attacks maximizing the output of individual neurons \cite{engstrom2019adversarial,singla2021understanding,Singla2022salient}, see %
Fig.~\ref{fig:PCA-vs-TopFeature-extra}. 
In Fig.~\ref{fig:spurious-exp} we show for each identified spurious feature,  the corresponding NPFV%
, e.g., for ``hummingbird'' one can see the bird feeder but no hummingbird. The NPCA components together with the maximally activating training images are in principle sufficient to identify spurious features, but the NPFV is very useful to judge how \emph{harmful} a spurious feature is. Therefore, we use an adversarially robust model, since a non-robust model would yield semantically meaningless adversarial noise as NPFV.

\textbf{Selection of neural PCA components for human inspection:}
The penultimate layer of the robust ResNet50 we are using has 2048 neurons. Thus it is infeasible (and unnecessary) to investigate all neural PCA components. A strong criterion that one has found a harmful spurious feature is i) the NPFV shows mainly the spurious feature and not the class, and ii) the NPFV has high confidence. If ii) is not satisfied, then the NPFV is a spurious feature the classifier may have picked up, but it is not harmful in the sense that this feature alone causes the classifier to choose that class. Moreover, we noticed that the eigenvalues of the neural PCA, and the corresponding $\alpha$ values, decay quickly. Thus we compute the NPFV for the top 128 neural PCA components (having maximal variance) and then select the ten components which realize the highest confidence for their NPFV in the corresponding class. Note that we do not optimize the confidence when generating the NPFV but only $\alpha^{(k)}_l(x)$ which is part of the logit of the $k$-th class.   

\textbf{Identification of spurious neural PCA components via human supervision:} %
For each ImageNet class $k$ we show the human labeler the top 10 components. For each component $l$ we show the NPFV $z^{(k)}_l$ and the 5 training images $x_r$ of class $k$ with the largest values of $\alpha^{(k)}_l(x_r)$. Moreover, we compute GradCAM \cite{Selvaraju2017GradCAM} images for the NPFV and the five training images using the NPCA component $\alpha^{(k)}_l$ as score. The human marks a component as spurious if i) the NPFV shows dominantly an object not belonging to the class %
ii) the five training images show consistently this object, iii) the GradCAM activations are primarily not on the class object. The setup shown to the human labeler can be seen in App.~\ref{app:labeling}. The labeling of one class takes on average about 45 seconds, so the full labeling of all ImageNet classes took about 13 hours. The human labeler (researcher in machine learning) found in total 337 spurious components. Another human labeler checked all of them and removed spurious features in case of disagreement, resulting in \XXX spurious features from 233 ImageNet classes. 
\input{spurious_examples.tex}
\section{SpuFix - Mitigation of spurious features}\label{sec:spufix}
Once the spurious features are identified, the question is how one can mitigate that the classifier relies on them. One way is to identify the training images containing the spurious feature and then discard or downweight them during training. However, this would require relabeling all spurious ImageNet classes which is not feasible. We could %
order the training set according to the value $\alpha^{(k)}_l$ of the corresponding neural PCA component %
which indicates how much of the spurious feature an image contains. While this would speed up the process significantly, it would still require a significant amount of manual relabeling. Can one do it also without any additional labeling? Yes, as described in Sec. \ref{sec:method} we can rewrite the logit of the $k$-th class as
\begin{equation} \label{eq:logit-fix}
f_k(x) = \sum\nolimits_{l=1}^D  \alpha^{(k)}_l(x) + \inner{\ones,\bar{\psi}_k} + b_k.
\end{equation}
For a spurious component $l$ of class $k$, we use $\min\{\alpha^{(k)}_l,0\}$ instead of $\alpha^{(k)}_l$ to remove its positive contribution from the logit (negative contributions are semantically different). After removal of the spurious features, the new logit becomes
\begin{equation}\label{eq:SpuFix-org}
    f_k^{SpuFix}(x) = f_k(x) - \sum_{l\in\mathcal{S}_k} \max\{\alpha^{(k)}_l(x),0\}
\end{equation}
where $\mathcal{S}_k$ is the set of spurious NPCA components of class $k$. We denote this method as \textbf{SpuFix}. It significantly reduces dependence on spurious features, see Sec. \ref{sec:exp-mitigation}.
\paragraph{Transfer of SpuFix to any ImageNet classifer:}

As described in Sec.~\ref{sec:spurious}, harmful spurious features are a result of subpopulations in the training data. While not every spurious correlation will be picked up by every model, most of our detected spurious features generalize to a wide range of classifiers (see Sec.~\ref{sec:exp}). In the following, we show how SpuFix can be transferred to any given ImageNet classifier $\Tilde{f}$ for which $\Tilde{f}_k$ denotes the logit and $\Tilde{\psi}_k$ the weighted penultimate layer of class $k$. The goal is to find %
a direction $b$ in the weighted feature space $\Tilde{\psi}_k$ of $\Tilde{f}$ for every spurious NPCA component $l$ corresponding to the eigenvector $v_l$ of the original model, resp. $\alpha_l^{(k)}(x)$, and then truncate its positive component. To find this direction we maximize the covariance of the projection onto $b$ and $\alpha^{(k)}_l$ over the training images of class $k$: %
\begin{equation}
    b^{(k)}_l=\argmax\limits_{\norm{b}_2 = 1} \sum_{s\in I_k} \inner{b,\tilde{\psi}_k(x_s) - \bar{\tilde{\psi}}_k}\alpha_l^{(k)}(x_s).
\end{equation}
which has a  closed form solution
\begin{equation}
    b^{(k)}_l = \frac{\sum_{s\in I_k} \left(\tilde{\psi}_k(x_s) - \bar{\tilde{\psi}}_k\right)\alpha_l^{(k)}(x_s)}{\norm{\sum_{s\in I_k} \left(\tilde{\psi}_k(x_s) - \bar{\tilde{\psi}}_k\right)\alpha_l^{(k)}(x_s)}_2}.
\end{equation}
In contrast to the eigenvectors $v_l$, the matched vectors $b^{(k)}_l$ are not necessarily orthogonal. 
Thus before truncation the centered features $\tilde{\psi}_k(x)-\bar{\tilde{\psi}}_k$ need to be projected onto the subspace spanned by the $b^{(k)}_l$ and  represented in the non-orthogonal basis
$\{b^{(k)}_l\}_{l\in\mathcal{S}_k}$. %
We denote this representation by $P^{(k)}(x)$ (details in \ref{app:non-orth}). %
The logit of class $k$ of the SpuFix version of $\tilde{f}$ is then:
\begin{equation}
    \Tilde{f}_k^{SpuFix}(x) = \Tilde{f}_k(x) - \sum_{l\in\mathcal{S}_k} \max\big\{\big<\mathbf{1}, b^{(k)}_l\big>P^{(k)}_l(x),0\big\}.
\end{equation}
In the case where $\Tilde{f} = f$ (the robust model), we recover the original SpuFix truncation in \eqref{eq:SpuFix-org} (see \ref{app:transfer-recover}).  It turns out that SpuFix is even effective when the architecture is quite different from the ResNet50 we used for detection, e.g. ViT or VOLO, see Sec.~ \ref{sec:exp-mitigation} and Table~\ref{tab:quantitative} and ~\ref{tab:app_extended}.

\section{Comparison to neural features of \cite{Singla2022salient}}\label{sec:comparison}
We compare our NPCA framework to the method of \cite{Singla2022salient} to detect spurious features for ImageNet. As model, they use a $\ell_2$-robust ResNet50. %
Let $J_k$ be the set of training images classified as class $k$. \cite{Singla2022salient} define the $j$-th component  $m^{(k)}_j$ of the class-wise mean over predictions,
\begin{equation}
    m^{(k)}= \frac{1}{|J_k|}
    \sum\nolimits_{x_s \in J_k}\psi_k(x_s),
\end{equation} 
as the importance of the $j$-th neuron for class $k$.\footnote{
The top-5 neurons do not change when using $I_k$ instead of $J_k$.
} Then, they order the neurons of the penultimate layer according to the score $m^{(k)}_j$ and consider the top-5 neurons of each class.  The main difference to our approach is that they assume single neurons with maximal influence on the mean are capturing spurious features whereas our NPCA components are linear combinations of neurons that capture the variance \emph{around} the mean. Given that the ResNet50 has only 2048 neurons for 1000 classes, some neurons are labeled as core \textbf{and} spurious feature for multiple classes simultaneously, even though the images are quite different. A major advantage of NPCA is that due to the orthogonality of the PCA components, we identify \emph{diverse} subpopulations in the training data. As \cite{Singla2022salient} use no constraints for the neurons, they often find very similar subpopulations. Hence, one may miss spurious subpopulations when only checking the top-5 components, see Fig.~\ref{fig:comparison}.
Another difference is that they maximize the score $m^{(k)}_j$ for the training images $x_r$ 
\[ w^{(k)}_j= x_r +\argmax_{\norm{\delta}_2\leq \epsilon} m^{(k)}_j(x_r+\delta),\]
whereas we maximize our NPCA component $\alpha^{(k)}_l(x)$ starting from a gray image to introduce no bias. Thus, we check if the classifier produces the spurious feature and not only enhances it on an image showing it already.
\input{top5_compare_their_resnet50_hummingbird_figure.tex}

The third difference is that \cite{Singla2022salient} want to identify \emph{any} spurious feature while our goal is to find \emph{harmful} ones. Thus, they use weaker criteria for deciding if a neuron shows a spurious feature: %
main criterion is if the neural activation map based on $m^{(k)}_j$ is off the class object according to %
a majority vote of 5 human labelers. The visualizations $w^{(k)}_j$ are only used if the activation maps are inconclusive. In contrast, we require i) our NFPV to show the spurious feature, ii) the GradCAM based on $\alpha^{(k)}_l$ highlights mainly the spurious
feature, and iii) our human labelers have to \emph{agree} that the NPCA component is a harmful spurious feature.
\newline
\indent As our criteria are more strict (in particular, that the NFPV shows the spurious feature is a strong criterion), it is not surprising that \cite{Singla2022salient} find more spurious features (630 in 357 classes) than we do (\XXX in 233 classes). Moreover, the employed models are different and we examine top-10 NPCA components whereas they check top-5 neurons. Thus, the comparison is difficult and our novel dataset ``SpuriousImageNet'' could be biased towards spurious features which only we found. Hence, we compute our top-5 NPCA components for their robust ResNet50 for a direct comparison to their top-5 neurons and their found spurious features. In  Fig.~\ref{fig:comparison} we compare them for the class hummingbird where they do not find the spurious feature ``bird feeder''.
In general, we observe that our found subpopulations are more diverse and thus we find more spurious features than they do when using their weaker criteria. In App.~\ref{app:comparison} we do an extensive comparison for %
all classes.

\begin{figure*}[ht!]
    \centering
    \includegraphics[width=\textwidth]{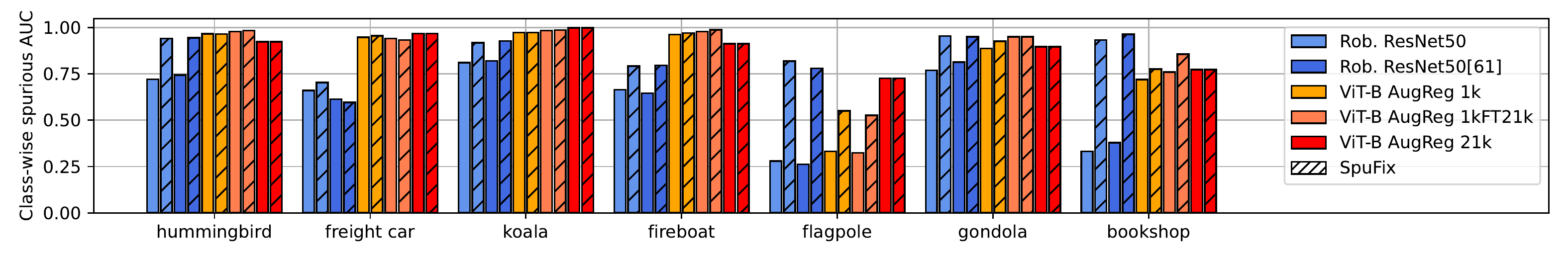}
\caption{\label{fig:barplot} \textbf{Spurious Score:} we plot the AUC of different models  for 7 out of 100 classes in ``Spurious ImageNet'' (see Fig \ref{fig:app_bar1}-\ref{fig:app_bar3}): hummingbird (bird feeder/red flowers), freight car (graffiti), koala (plants/trees), fireboat (water jet), flagpole (US flag), gondola (house/river), and bookshop (storefront). The 
spurious features for hummingbird, freight car and koala which were not detected in \cite{Singla2022salient,singla2022core} are also spurious for their robust ResNet50 \cite{Singla2022salient}. %
Training on ImageNet 21k or Fine-tuning from 21k (1kFT21k) decreases dependence on harmful spurious features but classes like flagpole and bookshop remain strongly affected.
Our cheap mitigation strategy SpuFix improves the AUCs significantly for both robust ResNet50 but also improves the ViT-B variants trained/fine-tuned on ImageNet1k, especially for the difficult classes flagpole and bookshop.}
\end{figure*}

\begin{figure}
    \centering
    \begin{tabular}{ccc}
        hummingbird & freight car & koala\\
        (red flowers) & (graffiti) & (plants/trees) \\
        \includegraphics[width=0.25\columnwidth]{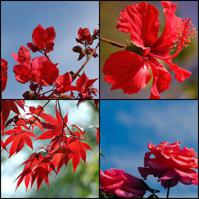}
        &
        \includegraphics[width=0.25\columnwidth]{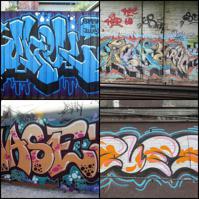}
        &
        \includegraphics[width=0.25\columnwidth]{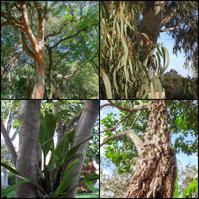}
        \\
        fireboat & flagpole & gondola \\
        (water jet) & (US flag) & (house/river)\\
        \includegraphics[width=0.25\columnwidth]{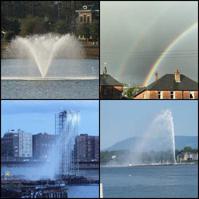}
        &
        \includegraphics[width=0.25\columnwidth]{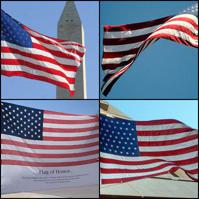}
        &
        \includegraphics[width=0.25\columnwidth]{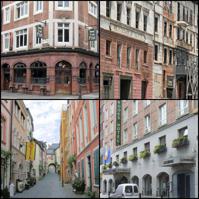}
        \\ 
    \end{tabular} \vspace{-2mm}
    \caption{\label{fig:dataset}\textbf{Spurious ImageNet:} sample images from the dataset for 6 out of 100 classes showing the spurious feature but not the class object, see also App.~\ref{app:dataset}.}\vspace{-2mm}
    \end{figure}

\section{Experiments}\label{sec:exp}
In this section, we provide a qualitative and quantitative evaluation of our  \XXX detected spurious features in ImageNet, see Sec. \ref{sec:method}. For the quantitative evaluation, we create the dataset ``Spurious ImageNet'', which allows checking the reliance of a given ImageNet classifier on spurious features. We also evaluate our mitigation strategy ``SpuFix'' which does not require additional labels or retraining of the classifier and can be transferred to other image classifiers.

\subsection{Qualitative evaluation}\label{sec:exp-qualitative}
For the qualitative evaluation, we visualize some of our found \XXX spurious features, see Fig.~\ref{fig:spurious-exp}. For each class, we show five random training images, the \NPFV, and the four most activating training images of the neural PCA component labeled to be spurious. Additionally, we always show ten images which \textbf{only} show the spurious feature but \textbf{not} the actual class e.g., only the bird feeder (spurious) but no hummingbird (class). All ten images are classified as the corresponding class for the robust classifier we have used to compute the NPCA components and three non-robust ImageNet classifiers (ResNext101, Eff.Net B5, ConvNext-B, see also Tab.~\ref{tab:app_extended}). This shows that our spurious features generalize from the robust classifier to SOTA ImageNet classifiers, indicating that the found spurious features are mainly due to the design of the training set, rather than failures in model training. Our novel validation by collecting real images with the spurious feature but without the class object which are consistently classified as this class directly shows the impact of harmful spurious features and has the advantage that it does not introduce artifacts via masking nor requires expensive pixel-wise segmentations.  \newline
\indent \textbf{Image Collection:} The images showing only the spurious feature were obtained by sorting the 9 million images of OpenImages \cite{OpenImages} by the value $\alpha^{(k)}_l(x)$ of the neural PCA component. We check the top 625 retrieved images classified by the robust classifier as the corresponding class if they are all classified as the same class by the additional non-robust classifiers \emph{and} do not show the corresponding class. This is a quite strict criterion as spurious features can be shared across classes, e.g., twigs for birds, and thus agreement of classifiers is not granted and the images can show the spurious feature \emph{and} the true class. Nevertheless, this procedure yields between 77 (``hummingbird'') and 179 (``freight car'') images of which we show a selection. For ``hummingbird'', a lot of these images show red flowers (without a hummingbird) which makes sense as the NPFV displays features of red flowers and due to the bias that OpenImages does not contain many images of hummingbird feeders. In these cases, we additionally retrieve Flickr images with appropriate text queries e.g., ``hummingbird feeder'' and filter them. \newline
\indent \textbf{Spurious Class Extension:} For ``hummingbird'', ``freight car'', and ``koala'' the spurious features significantly extend the predictions beyond the actual class (see Fig.~\ref{fig:spurious-illustration}). Bird feeders are classified as hummingbirds, graffiti as freight cars, and (eucalyptus) plants as koalas. This class extension cannot be detected by monitoring test performance and thus is likely to be noticed only after deployment. For ``hummingbird'', we see in Fig.~\ref{fig:spurious-exp} two images with bees on the bird feeder where ``bee'' is an ImageNet class (also a grasshopper for ``koala''). Nevertheless, the spurious ``bird feeder'' feature of ``hummingbird'' overrules ``bee'' even though no hummingbird is present. \newline
\indent \textbf{Spurious Shared Feature:} The spurious feature ``water jet'' is shared among the classes ``fireboat'' and ``fountain''. It appears more frequently for ``fireboat'' (see Fig.~\ref{fig:spurious-illustration}) which leads to an in-distribution shift where now a large number of images of the ``fountain''-class with a water jet are wrongly classified as ``fireboat''. %
The spurious feature ``water jet'' for ``fireboat'' has been found also in the Salient ImageNet dataset \cite{Singla2022salient,singla2022core}. However, they did not find 
spurious features for freight car %
and koala (in App.~\ref{app:koala-PCA} we do a comparison). 
More examples are in App.~\ref{app:qualitative}.

\subsection{The Spurious ImageNet dataset}
\label{sec:exp-dataset}
A key contribution of this paper is our novel evaluation of spurious features for image classifiers without requiring pixel-wise annotations \cite{Plumb2022finding,shetty2019not} or having to rely on the validity of neural heatmaps \cite{Singla2022salient}. Instead, we use images from OpenImages to show that images only containing the spurious feature but not the class object are classified as this class. This has the advantage that we consider real images and thus provide a realistic impression of the performance of ImageNet classifiers in the wild. Adding noise \cite{Singla2022salient} or masking \cite{Plumb2022finding,moayeri2022hard} image regions requires pixel-wise accurate annotations which are labor-expensive, masking only the object still contains shape information, and using masks avoiding this, e.g., a bounding box around the object, can hide a significant portion of the image which is unrealistic.\newline 
\indent To allow for a quantitative analysis of the influence of spurious features on ImageNet classifiers, we collected images similar to the ones shown to illustrate the spurious features in Fig.~\ref{fig:spurious-exp}. The images are chosen such that they show the spurious feature but not the class object. The only difference is that we relax the classification condition and only require two of the four classifiers (robust ResNet50, ResNext101, EfficientNet-B5, ConvNext-B) to predict the corresponding class. We select 100 of our spurious features and for each collect 75 images from the top-ranked images in OpenImages according to the value of $\alpha^{(k)}_l$ for which two human labelers agree that they contain the spurious feature but not class $k$ and two out of four classifiers predict class $k$. We call the dataset \textbf{Spurious ImageNet} as it allows to check the dependence on spurious features with real images for ImageNet classifiers, see Fig.~\ref{fig:dataset} and App.~\ref{app:dataset} for samples.

\indent \textbf{Spurious Score:} A classifier $f$ not relying on the spurious feature should predict a low probability for class $k$ for the Spurious ImageNet samples, especially compared to ImageNet test set images of class $k$. Thus, for each class, we measure the AUC (area under the curve) for the separation of images with the spurious features but not showing class $k$ versus test set images of class $k$ according to the predicted probability for class $k$. A classifier not depending on the spurious feature should attain a perfect AUC of 1, whereas a value significantly below 1 shows strong reliance. We report the mean AUC (mAUC) over all 100 classes in Tab.~\ref{tab:quantitative}. All ImageNet models trained only on ImageNet1k are heavily influenced by spurious features. Thus, spurious features are mainly a problem of the training set rather than the classifier, and spurious features found with an adversarially robust model transfer to other ImageNet classifiers. \newline
\indent \textbf{Pre-training on larger datasets:} Some spurious features such as flag (flag pole), bird feeder (hummingbird), and eucalyptus (koala) are classes in ImageNet21k. Therefore, they should no longer be spurious for the other classes. Thus, we test if ImageNet1k-classifiers fine-tuned from an ImageNet21k model are less reliant on spurious features. The results in Tab.~\ref{tab:quantitative} and Fig.~\ref{fig:barplot} suggest that the influence of spurious features is damped but they are far from being free of them. To check how much is lost due to fine-tuning we evaluate a ViT-B trained on ImageNet21k which has a mean AUC of 0.931 whereas the fine-tuned model has 0.917. This shows that fine-tuning does not hurt much. While finetuning from ImageNet21k improves the mean AUC, for several classes the dependence on spurious features is still significant, see also Fig.~\ref{fig:quill} how one has to be careful in the interpretation of higher AUC values. In addition to ImageNet21k, we also evaluate models trained on other large image datasets (JFT-300M\cite{gupta2017revisiting}, YFFC-100M, 1B Instagram\cite{yalniz2019billion}, MIM\cite{EVA},  LAION-2B and LAION-400M\cite{schuhmann2022laion}) using self-supervised learning or which are based on CLIP \cite{radford2021learning}. However, these models also do not achieve better spurious scores (Tab.~\ref{tab:quantitative} and Tab.~\ref{tab:app_extended}). We evaluate a large number of SOTA models in App.~\ref{app:quant_eval}.

\begin{table}
   \setlength{\tabcolsep}{0.2em}
    \centering
   \begin{tabular}{l|c|c?{0.35mm}c|c}

    & \multicolumn{2}{c?{0.35mm}}{\textbf{Original}} & \multicolumn{2}{c}{\textbf{SpuFix}}\\
     &  INet & SpurIN & INet & SpurIN\\
    Model & Acc. $\uparrow$& mAUC $\uparrow$& Acc. $\uparrow$& mAUC  $\uparrow$\\
    \hline
    \multicolumn{5}{c}{ImageNet1k}\\
    \hline
     Rob. ResNet50 & 57.4\% & 0.630 & 56.8\%&\textbf{0.763} \\
     \hline
     Rob. ResNet50\cite{Singla2022salient} & 57.9\% & 0.651 & 57.2\% &\textbf{0.764}\\  
     ConvNeXt-L\cite{liu2022convnet} & 84.8\% &  0.803  & 84.8\% &\textbf{0.819}\\
     ViT-B AugReg\cite{steiner2021train} &81.1\% &  0.850 & 81.1\% &\textbf{0.859}\\
     VOLO-D5 512\cite{yuan2022volo} & 87.1\% & 0.882 & 87.1\% &\textbf{0.907}\\
     \hline
     \multicolumn{5}{c}{ImageNet21kFT1k}\\
     \hline
     EfficientNetv2-L\cite{tan2021efficientnetv2}  & 86.8\%  & 0.893 & 86.8\% &\textbf{0.898}\\
     ConvNeXt-L\cite{liu2022convnet} & 87.0\%   & 0.910 & 87.0\%  &\textbf{0.913}\\
     ViT-B AugReg\cite{steiner2021train}  & 86.0\%  &  0.917& 85.9\% &\textbf{0.925}\\
     BEiT-L\textbackslash16\cite{bao2021beit}  &88.6\% & 0.921 & 88.6\% &\textbf{0.927}\\
    \hline 
     \multicolumn{5}{c}{LAION-2B}\\
     \hline
     {\small CNeXt-L CLIP 384\cite{radford2021learning}} & 87.8\% & 0.879  & 87.9\% &\textbf{0.884}\\
     {\small ViT-L\textbackslash14 CLIP\cite{radford2021learning}} & 88.2\% & 0.912 & 88.2\% &\textbf{0.914} \\
     \hline
     \multicolumn{5}{c}{MIM}\\
     \hline
     {\small EVA-G\textbackslash14 CLIP 560\cite{EVA}} & 89.8\% & 0.919 &89.8\% &\textbf{0.925}\\
      \hline
     \multicolumn{5}{c}{ImageNet21k}\\
     \hline
     ConvNeXt-L\cite{liu2022convnet} & - & 0.943 & -&0.943 \\
      ViT-B  AugReg\cite{steiner2021train} & - & 0.931 & -&0.931\\
     \hline
     \end{tabular}
    \caption{\label{tab:quantitative}\textbf{Quantitative Evaluation on Spurious ImageNet:}
    ImageNet classifiers of different training modalities depend on spurious features in varying strength. The mAUC is the mean of AUCs for the separation of images containing the spurious feature but not class $k$ versus test images of class $k$ with the predicted probability of class $k$ as score.  %
    }
    \vspace{-3mm}
\end{table}
\subsection{Evaluation of mitigation technique SpuFix}\label{sec:exp-mitigation}
Fixing spurious features is a non-trivial task and can require a substantial labeling effort. We evaluate our SpuFix from Sec.~\ref{sec:spufix} that does not require retraining or additional labels. The positive effect of this fix of spurious features (SpuFix) can be seen in Tab.~\ref{tab:quantitative} and Fig.~\ref{fig:barplot}. Compared to the original robust ResNet50 with a spurious mAUC of 0.630, the SpuFix version has a significantly better spurious mAUC of 0.763. Test set accuracy reduces by $0.6\%$ but this is a rather positive effect, as several of the additional errors arise since the robust ResNet50 uses spurious features for its decision, e.g., for classes like ``balance beam'' or ``puck'' the class object is often not visible in the cropped test set images. In Table \ref{tab:app_extended} we provide a large scale evaluation of the transfer of SpuFix to SOTA ImageNet models. We observe a consistently better mAUC on Spurious ImageNet, even for very large models fine-tuned from 21k or trained on other large datasets, e.g. SpuFix improves the mAUC of VOLO-D5 (87.1\% acc.) trained only on 1k by 2.5\%, or by 0.6\% for EVA-G\textbackslash 14 CLIP 560 trained on MIM (91.9\% acc.), as well as BeiT-L\textbackslash 16 fine-tuned from 21k (88.6\% acc.) by 0.6\%. Thus even SOTA models profit from our SpuFix with negligible difference in accuracy $(\leq 0.1\%)$ and thus the use of SpuFix is recommended for any ImageNet model.

\section{Conclusion}
We have shown that large-scale identification of spurious features is feasible with our neural PCA components and neural PCA feature visualizations. With ``Spurious ImageNet'' we introduced a novel dataset to evaluate the dependence of ImageNet classifiers on spurious features based on real images. We demonstrated that our SpuFix method mitigates the dependence on harmful spurious features for any ImageNet classifier without costly labeling or re-training.%

\paragraph{Acknowledgements}
We acknowledge support by the DFG, Project number 390727645, the Carl Zeiss Foundation,   project “Certification and Foundations of Safe Machine Learning Systems in Healthcare”. The authors thank the IMPRS-IS for supporting YN.

{\small
\bibliographystyle{ieee_fullname}
\bibliography{egbib}
}
\clearpage
\let\cleardoublepage\clearpage

\appendix
\section*{Overview of Appendix}
In the following, we provide a brief overview of the additional experiments reported in the Appendix.

\begin{itemize}
 \item In App.~\ref{app:koala-PCA}, we compare our top-5 Neural PCA components for our robust model vs. the  neural features of the Top-5 neurons of \cite{Singla2022salient} for their robust model for classes ``Koala'', ``Indigo-Bunting'', and ``Mountain Bike'', for which \cite{Singla2022salient} do not find spurious features and we do (see App.~\ref{app:comparison} for a direct comparison of NPCA for their robust model to their found components as done in Fig. \ref{fig:comparison}) 
 \item In App.~\ref{app:labeling}, we explain our labeling setup to create the dataset ``Spurious ImageNet'' in more detail.
\item In App.~\ref{app:non-orth} and App.~\ref{app:transfer-recover}, we present the details of transferring SpuFix to other models. In particular, we define the orthogonal projection $P^{(k)}$ onto the subspace spanned by non-orthogonal vectors and show that the transfer recovers the original SpuFix method when applied to the original model. We validate that the SpuFix improvement is independent of the image collection procedure in Fig.~\ref{fig:testset}.%
 \item In App.~\ref{app:quant_eval}, we use our ``Spurious ImageNet'' dataset to quantitatively analyze the dependence of the classifiers on spurious components. By doing so, we show that pre-training on larger datasets like  ImageNet21k helps to reduce this dependence. We also discuss the empirical results of the SpuFix method.
 \item In App.~\ref{app:comparison} we continue the comparison to \cite{Singla2022salient} from Section \ref{sec:comparison}. As in Fig.~\ref{fig:comparison}. we do a direct comparison to their found top-5 neurons by computing the top-5 NPCA components for their robust model. We observe that their top-5 neurons are less diverse than our top-5 NPCA components, see Fig.~\ref{fig:hist-matched} and Fig.~\ref{fig:hist-zeros}.
 \item In App.~\ref{app:qualitative}, we extend our qualitative evaluation of the spurious components from Figure \ref{fig:spurious-exp}.
 \item In App.~\ref{app:dataset}, we show random samples from all 100 spurious features in our ``Spurious ImageNet'' dataset.
\item In App.~\ref{app:dvces_flips}, we show how we change the predicted class for an image by introducing only spurious features of the target class. To do this automatically, we adapt the Diffusion Visual Counterfactual Explanations (DVCEs) of \cite{Augustin2022Diffusion}.
\end{itemize}
\section{Neural PCA Components}\label{app:koala-PCA}
We illustrate in Fig.~\ref{fig:PCA-vs-TopFeature-extra} that our neural PCA components capture the different subpopulations in the training set better compared to the neural features of \cite{Singla2022salient}. We find three spurious features: eucalyptus/plants for the class koala, twigs for the class indigo bunting, and forest for mountain bike, which were not found by \cite{Singla2022salient}. Please see the caption of Fig.~\ref{fig:PCA-vs-TopFeature-extra} for more details. Note that for this comparison, we consider the NPCA components computed on our robust ResNet50 which differs from the one used in \cite{Singla2022salient}. See Fig.~\ref{fig:worst12_compare_figure_1},\ref{fig:worst12_compare_figure_2} and App.~\ref{app:comparison} for a comparison using the same model.

For 46 out of 100 classes in our ``Spurious ImageNet'' dataset, no spurious feature is reported in \cite{Singla2022salient}. The 46 classes are: tench, indigo bunting, American alligator, black grouse, ptarmigan, ruffed grouse, 
s.-c. cockatoo, hummingbird, koala, leopard, walking stick, gar, bakery, barbershop, barn, bathtub, beer bottle, bikini, bulletproof vest, bullet train, chain mail, cradle, dam, dumbbell, fountain pen, freight car, hair spray, hamper, hard disc, mountain bike, neck brace, nipple, obelisk, ocarina, pencil box, pill bottle, plastic bag, plunger, pole, pop bottle, quill, radio telescope, shoe shop, shovel, steel drum, cheeseburger. However, note that even if for the same class their and our method report a spurious feature, this need not be the same.
\input{top5_compare_figure}

\begin{figure*}
    \centering
    \includegraphics[width=0.8\textwidth]{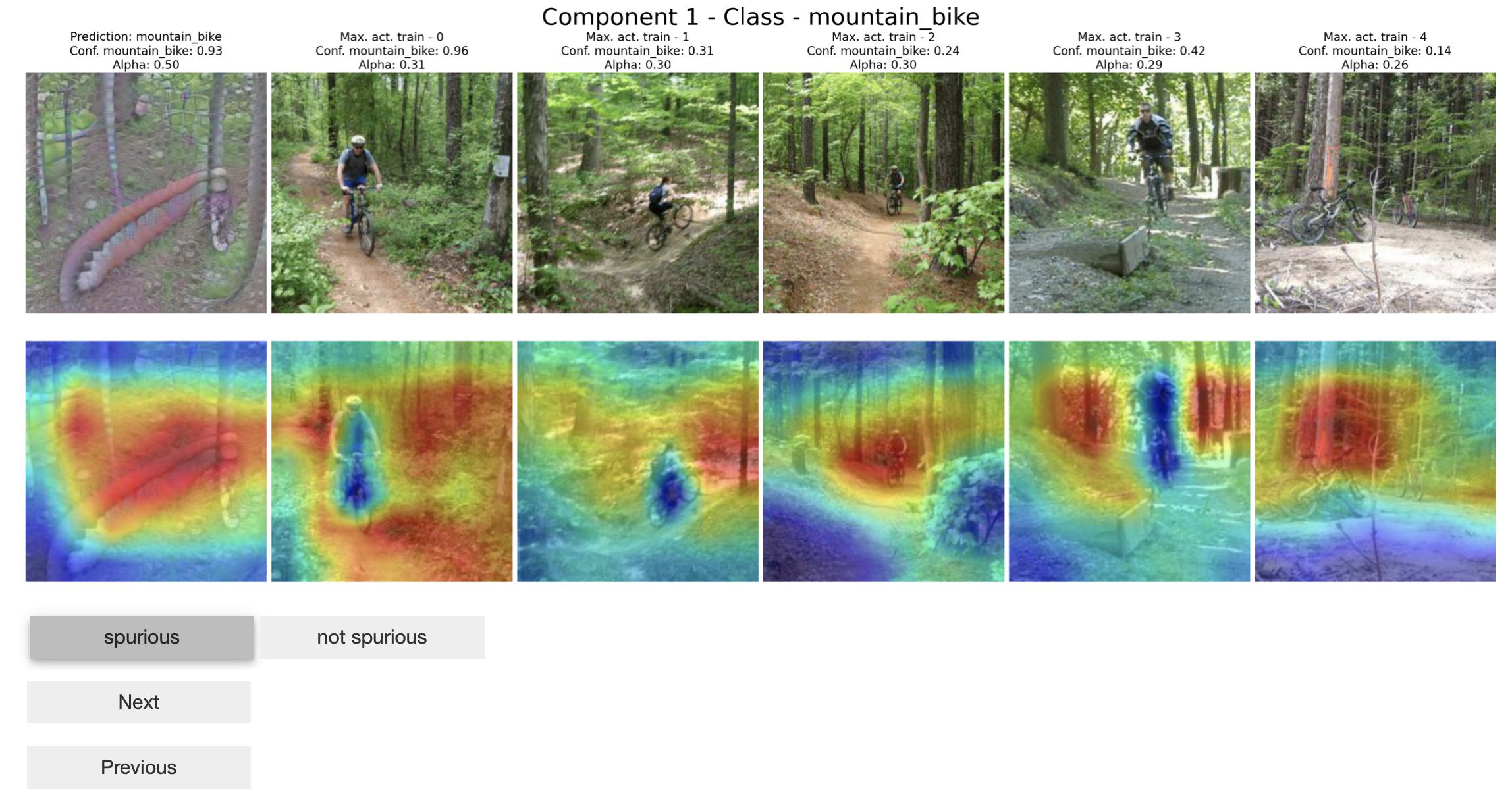}
    \caption{\label{fig:PCA-labeling}\textbf{Illustration of the information shown for labeling neural PCA components:} The illustration shows the visualization of the first neural PCA component of the class ``mountain bike''. The first image on the left shows the NPFV, the prediction of the robust ResNet50, its probability for the class ``mountain bike'', and the corresponding value of $\alpha^{(k)}_l$. The other five images shown, along with the corresponding probabilities and $\alpha^{(k)}_l$, are the maximally activating training images of this component. The second row shows GradCAM heatmaps with respect to the component $\alpha^{(k)}_l(x)$. Below the visualization, the labeler can select one of the two possible labels (\emph{spurious} and \emph{not spurious}) and navigate through the next or last neural PCA component.}
    \label{fig:labeling_screenshot}
\end{figure*}

\begin{figure*}
    \centering
    \includegraphics[width=0.5\textwidth]{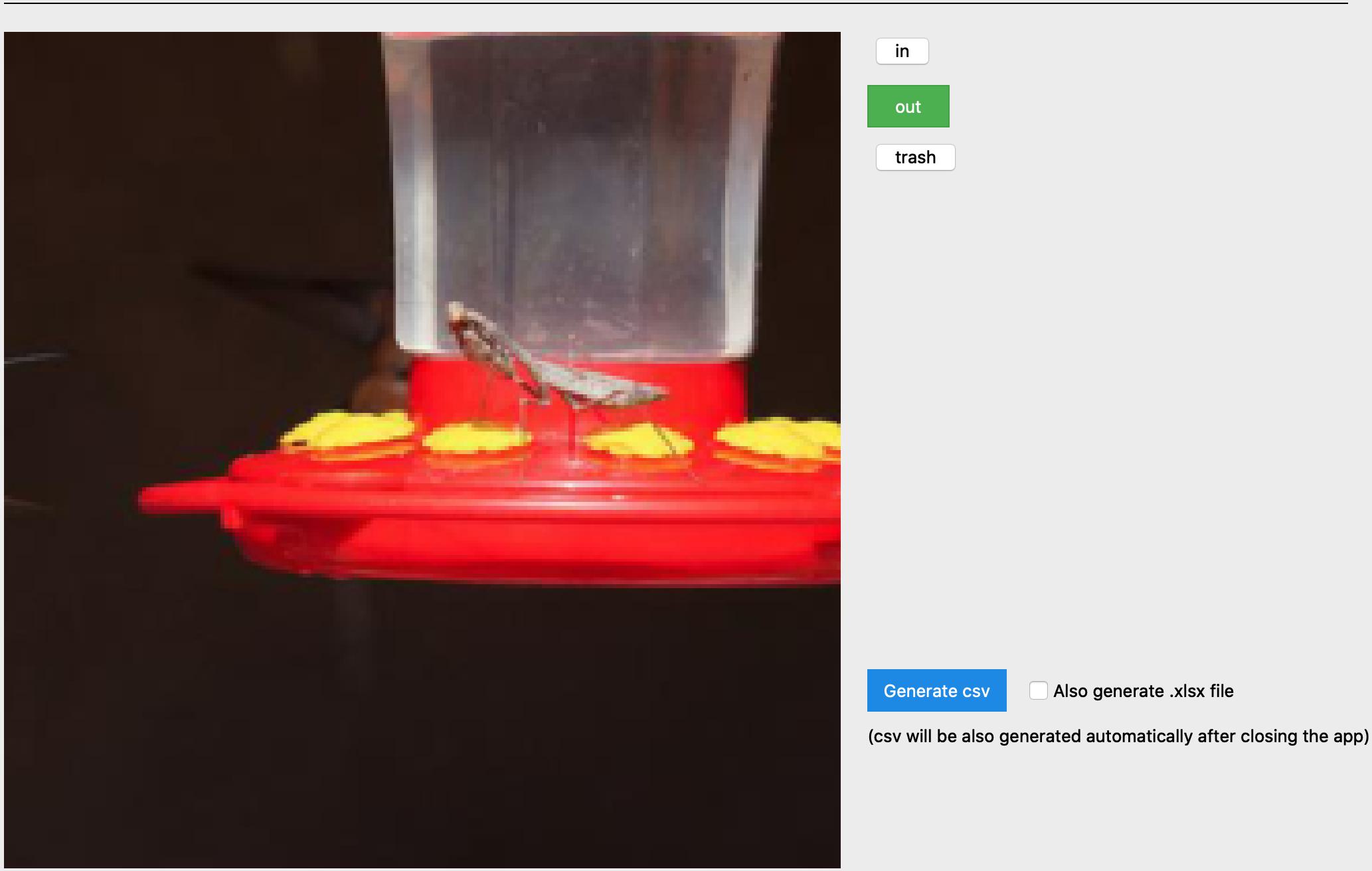}
    \caption{\label{fig:labeling_dataset_screenshot}\textbf{Illustration of the information shown for labeling images to create our ``Spurious ImageNet'':} This screenshot illustrates a tool that we used to create labels for our dataset and an example of the image that is chosen to be in our dataset, in class ``hummingbird'', as it contains the spurious feature bird feeder of the class ``hummingbird'' but no hummingbird.}
    \hspace{2cm}
\end{figure*}

\section{Labeling Setup for Spurious features}\label{app:labeling}
In our paper, we have two labeling tasks for two objectives: i) identifying spurious components, and ii) creating ``Spurious Imagenet''.

\textbf{Identifying spurious components}. Fig.~\ref{fig:labeling_screenshot} illustrates the information shown to the human labeler to identify neural PCA components corresponding to spurious features. This includes the NPFV, the 5 most activating training images, and GradCAM heatmaps, as well as the corresponding class probabilities and $\alpha^{(k)}_l$ values. The decision, of whether a neural PCA component corresponds to a spurious feature has been made only based on the visualization as shown in Fig.~\ref{fig:labeling_screenshot}.

\textbf{Creating ``Spurious Imagenet''.}
To create our ``Spurious ImageNet'' dataset, 
\begin{itemize}
    \item we selected 100 (spurious component $l$, class $k$) pairs, such that for each class we have only one selected spurious component, and sorted all images from OpenImages for which at least two of our four classifiers predict class $k$ according to the value $\alpha^{(k)}_l$ of the respective neural PCA component $l$;
    \item we have used the open-source tool\footnote{https://github.com/robertbrada/PyQt-image-annotation-tool} for labeling images and created three labels ``in'' (the images that contain the class features), \textbf{``out''} (the images that contain only the spurious and no class features), ``trash'' (images that are too far from the distribution of the spurious features and contain no class features) for each image as can be seen in the Fig.~\ref{fig:labeling_dataset_screenshot};
    \item for each component, 75 images that are guaranteed to contain the spurious feature but not the class object of class $k$ were selected by two human labelers. Images were only accepted into the dataset if both labelers assigned the label ``out''.
\end{itemize}

\section{SpuFix}
\subsection{SpuFix - Orthogonal projection onto a non-orthogonal basis}\label{app:non-orth}
Let $L=|\mathcal{S}_k|$. The projection can be written as a least squares problem: Let $b_1, \ldots, b_L$ be the matched directions and $B\in\mathbb{R}^{\tilde{D}\times L}$ the matrix containing them as columns. Now, the projection onto the subspace spanned by the $b_l, l\in\{1, \ldots, L\}$ is given by
\begin{equation}
    \min_{P^{(k)} \in \mathbb{R}^L} \norm{\tilde{\psi}_k(x) - \bar{\tilde{\psi}}_k - BP^{(k)}}_2^2
\end{equation}
with closed-form solution
\begin{equation}
    P^{(k)}(x) = (B^TB)^{-1}B^T\left(\tilde{\psi}_k(x) - \bar{\tilde{\psi}}_k\right).
\end{equation}
\subsection{SpuFix - Recovering the original method}\label{app:transfer-recover}
When using our robust ResNet50, it holds $\Tilde{f} = f$. Then the matched directions are 
\begin{align}
     b_l^* &= \frac{\sum_{s\in I_k} \left(\psi_k(x_s) - \bar{\psi}_k\right)\alpha_l^{(k)}(x_s)}{\bigg\lVert\sum_{s\in I_k} \left(\psi_k(x_s) - \bar{\psi}_k\right)\alpha_l^{(k)}(x_s)\bigg\rVert_2} \\
     &=\frac{C v_l\inner{\mathbf{1}, v_l}}{\bigg\lVert C v_l\inner{\mathbf{1}, v_l}\bigg\rVert_2}\\
     &=\frac{\lambda_l v_l \inner{\mathbf{1}, v_l}}{\bigg\lVert\lambda_l v_l\inner{\mathbf{1}, v_l}\bigg\rVert_2} = v_l
\end{align}
where $\lambda_l$ is the eigenvalue corresponding to the eigenvector $v_l$ and we have used: 
\begin{align*} &\sum_{s\in I_k} \left(\psi_k(x_s) - \bar{\psi}_k\right)\alpha_l^{(k)}(x_s)\\
=&  \sum_{s\in I_k} \left(\psi_k(x_s) - \bar{\psi}_k\right) \inner{\psi_k(x_s) - \bar{\psi}_k,v_l}\inner{v_l,\mathbf{1}}\\
=& Cv_l \inner{v_l,\mathbf{1}}
\end{align*}
Thus, as the $v_l$ are an orthonormal basis it holds $P^{(k)}_l(x)=\inner{v_l,\tilde{\psi}_k(x) - \bar{\tilde{\psi}}_k}$ and we get
\begin{align}
    &\Tilde{f}_k^{SpuFix}(x)\\ 
    =& \Tilde{f}_k(x) - \sum_{l\in\mathcal{S}_k} \max\{\inner{\mathbf{1}, v_l}P^{(k)}_l(x),0\} \\
    =& f_k(x) - \sum_{l\in\mathcal{S}_k} \max\{\alpha_l^{(k)},0\} = f_k^{SpuFix}(x).
\end{align}

\input{appendix_mauc_table_new.tex}

\section{Quantitative Evaluation}\label{app:quant_eval}
In this section, we extend the quantitative results given in the main paper in Tab.~\ref{tab:quantitative} for a large number of ImageNet models. In Tab.~\ref{tab:app_extended} we show ILSVRC-2012 test accuracies and the mean spurious AUC (mAUC) for a wide selection of ImageNet models with different architectures and training configurations. Again, our spurious AUC is computed classwise using the predicted probability for that class as a score where we compare the images corresponding to this class of ``Spurious ImageNet'' (not showing the class object, but just the spurious feature) vs the ImageNet validation set images of that class. Finally, we take the mean of all classwise AUCs to get the final mAUC. All models except for our multiple-norm robust ResNet50 and the robust ResNet50 from \cite{Singla2022salient} are taken from PyTorch Image Models \cite{rw2019timm}.%
We further distinguish between models trained on ImageNet1k only (in1k), models pre-trained on ImageNet21k and then fine-tuned on ImageNet1k (in1kFT21k), ImageNet21k classifiers (in21k) and models trained using semi-supervised training techniques on large datasets containing 100M or more images \cite{xie2020selftraining,yalniz2019billion} and multimodal CLIP \cite{radford2021learning} models that are pre-trained on large datasets containing text and image pairings \cite{schuhmann2021laion, schuhmann2022laion}. Note that we do not report accuracies for ImageNet21k models as no test set for in21k is available. All models pre-trained on other datasets than ImageNet21k are Imagenet 1k models with a classification head containing 1000 classes after potential fine-tuning. In Fig.~\ref{fig:app_acc_vs_auc}, we also plot mean spurious AUC against ImageNet-1k test accuracy and color code the models based on the dataset used during training.

\textbf{Pre-training and fine-tuning:} Overall, the trend seems to be that better models (in terms of accuracy) improve in mAUC and are less vulnerable to spurious features. It is also easily observable that pre-training on larger datasets such as ImageNet-21k can help to decrease vulnerability to spurious features, which can be seen best from the EfficientNetv2-M/L, ViT-B/L AugReg and ConvNeXt-L models for which we can evaluate the difference between in1k, in1kFT21k, and in21k training. The in1k ViT-B\textbackslash16 AugReg achieves an mAUC of 0.850 whereas the same in21k model achieves an mAUC of 0.931 before and 0.917 after in1k fine-tuning. Similar trends are also visible for the EfficientNetv2-M and ConvNeXt-L models, where all ImageNet21k models (in1kFT21k and in21k) perform better than pure in1k models, however, parts of the improvement of the in21k models is lost during fine-tuning.  While we do not have pure in1k models for them to compare to, other in21k pre-trained models such as the Big Transfer models, as well as the standard ViT's without AugReg, the BEiT and Swin architecture-based models show the same behavior and decrease spurious mAUC during fine-tuning. It thus remains an open question how one can use the benefits of pre-training on massive datasets with fine-grained class structures to preserve or even improve mAUC during fine-tuning to smaller datasets such as ImageNet1k. 

\textbf{Different architectures:} In terms of architecture, there is no easily observable trend. On pure in1k models, VOLO D5 achieves the best mAUC of 0.882, however, it is also the  most accurate model. The best overall model in terms of mAUC is the BEiTV2-L in21k with an mAUC of 0.951, however, after fine-tuning, the mAUC decreases to 0.921 where it achieves similar values as some other models like the ViT-B Augreg (0.917), ConvNextV2-H (0.919) and BEiT-L (0.921) which are the best models with ImageNet-1k classification head in terms of mAUC. In summary, attention-based transformers do not seem to yield strong benefits over convolutional neural networks in terms of vulnerability to spurious features.  
From Table \ref{tab:app_extended}, we also see that semi-supervised training approaches like Noisy Student self-training\cite{xie2020selftraining} can help to improve mAUC over pure ImageNet-1k training. However, there the smallest EfficientNet B5 actually achieves better mAUC than all other models, even the EfficientNet L2 which achieves much better clean accuracy. Pre-training using CLIP on large image/text datasets can also yield models with mAUC above 0.9 and is comparable to in21k pre-training. For example, the ViT-L achieves an mAUC of 0.914 (with AugReg) after in21k pre-training and 0.912 after CLIP pre-training on LAION-2B (without AugReg).  

\textbf{SpuFix on the robust ResNet50:} The robust ResNet50 shows a substantial improvement in mAUC from 0.630 to 0.763. Fig.~\ref{fig:app_bar1} to ~\ref{fig:app_bar3} show the class-wise values. In particular, it raises the class-wise spurious AUC from 0.332 to 0.932 for bookshop (+60.0\%), from 0.279 to 0.819 for flagpole (+54.0\%) and from 0.246 to 0.778 for Band Aid (+53.3\%). Overall, the mAUC increases for 95 of 100 classes and achieves an improvement of at least $0.1$ for 49 of them. Both the SpuFix method and the image collection procedure for the Spurious ImageNet benchmark are based on the values $\alpha_l^{(k)}$ for spurious NPCA components $l$. Thus, to further validate the benefit of SpuFix, we collected 10 images each for the classes hummingbird, gondola and flagpole containing \textbf{only} the spurious feature (bird feeder, building/canal, US flag) \textbf{without} any automated filtering, i.e. neither model predictions nor NPCA components were used. Fig.~\ref{fig:testset} shows the predictions as well as the class probability for the spuriously correlated class of the original model and the SpuFix version for these images. The harmful predictions, i.e. predictions of the spuriously correlated class, are reduced from 6 to 3 for hummingbird, 8 to 4 for gondola and 7 to 0 for flagpole. SpuFix also decreases the mean class probability over the 10 images: from 0.57 to 0.13 (hummingbird), 0.61 to 0.13 (gondola), and 0.70 to 0.05 (flagpole). The actual improvements for the individual images are even larger due to the fact that the original model already does not predict the corresponding class or shows a low class probability for some of them. This shows that SpuFix indeed mitigates the reliance on these spurious features independent of the image collection procedure.
\input{figure_testset.tex}

\textbf{SpuFix on other ImageNet classifiers:} In addition to the values for the original models, Tab.~\ref{tab:app_extended} also contains the mAUC and accuracies for the SpuFix versions of all evaluated models. Furthermore, the improvements are depicted as arrows in Fig.~\ref{fig:app_acc_vs_auc}. One can see that SpuFix consistently improves the mAUC consistently for \textbf{all} models that were trained or fine-tuned on ImageNet1k. Even the top performing models still benefit significantly, e.g. 0.919 to 0.925 (+0.6\%) for the EVA-Giant\textbackslash14 CLIP 560 (MIM) or 0.921 to 0.927 (+0.6\%) for the BEiT-L\textbackslash16 pre-trained on ImageNet21k. On the other hand, the effect of SpuFix on the validation accuracy is negligible. Only the different variants of the ResNet50 architecture show a decrease of more than $0.1\%$. However, these models also achieve the largest improvements in spurious mAUC, e.g. the ResNet50 SSL (1B Instagram) loses $0.4\%$ accuracy but also has a significant gain of +1.5\% in spurious mAUC.
We want to stress again that SpuFix can be applied to any ImageNet classifier with minimal effort and the code for doing so is part of the github repo (no retraining, labels etc required).

\textbf{Models with high mAUC still rely on harmful spurious features:} While the general trend of accuracy versus mAUC validates the progress of recent vision models, it does not mean that spurious features are no longer a problem for those models, especially since the worst-performing classes are still severe modes of failure. To better understand this behavior on individual classes, we plot the class-wise AUC for all 100 classes and a selection of models in Fig.~\ref{fig:app_bar1} to \ref{fig:app_bar3}. While both robust ResNet50 models are overall worse than the much larger comparison models ViT-B AugReg, we highlight that our SpuFix method (see Section \ref{sec:exp-mitigation}) does significantly improve the mean spurious AUC of both ResNets50 models without requiring retraining. The ViT-AugReg also benefits from SpuFix but due to the transfer to smaller extent - nevertheless one can observe improvements by more than 5\% in AUC for the classes 
flagpole, pole, puck, bookshop, lighter. On average, it is again observable that ImageNet-21k pre-training does improve spurious AUC. However, in terms of the final ImageNet-1k classifier after fine-tuning, classes like bakery, flagpole, wing, or pole remain challenging and can have a class-wise AUC as low as 0.5. Thus the improvements seem to depend heavily on the structure of the dataset used for pretraining and whether or not this dataset contains the spurious feature as an individually labeled class that allows the model to distinguish the class object from the spurious feature. For example, ImageNet-21k contains both flagpole and flag as separate classes, thus in21k ViT-B model achieves much better spurious AUC for these class (Spurious ImageNet contains flag images without flagpoles) than the ViT-B with a ImageNet-1k classification head. Nevertheless, even the in21k models still show a low AUC for flag pole and thus have problems distinguishing between flags (spurious feature) and flag pole. It also has to be noticed that the seemingly high AUC values are sometimes misleading. First of all, we stress again that the images of Spurious ImageNet do not contain the class object and thus an AUC of one should be easily obtainable for a classifier. Second, even if the AUC is one, it only means that the predicted probability for the validation set images (containing the class object) is always higher than the predicted probability for images from Spurious ImageNet (not containing the class object). However, still, a large fraction of the Spurious ImageNet images can be classified as the corresponding class, e.g. the ConvNext-L-1kFT21k has a class-wise AUC of 0.93 for ``quill'', but still 71\% of all Images from Spurious ImageNet are classified as ``quill'', see Fig.~\ref{fig:quill} where we show for each image from ``Spurious ImageNet'' the top-3 predictions with their predicted probabilities. Thus the class extension can still be significant even for such a strong model. There are also classes which are completely broken like puck with an AUC of 0.69, where 100\% of all images in ``Spurious ImageNet'' are classified as puck. The reason is that the puck is simply too small in the image (or sometimes even not visible at all), whereas the ice hockey players and also part of the playing field boundary are the main objects in the image. Thus the classification is only based on spurious features and the object ``puck'' has never been learned at all.

\input{appendix_bar_plots.tex}

\input{app_749_quill.tex}

\begin{figure*}
    \setlength{\tabcolsep}{0.1em}
    \centering
    \normalsize
    \includegraphics[width=\linewidth]{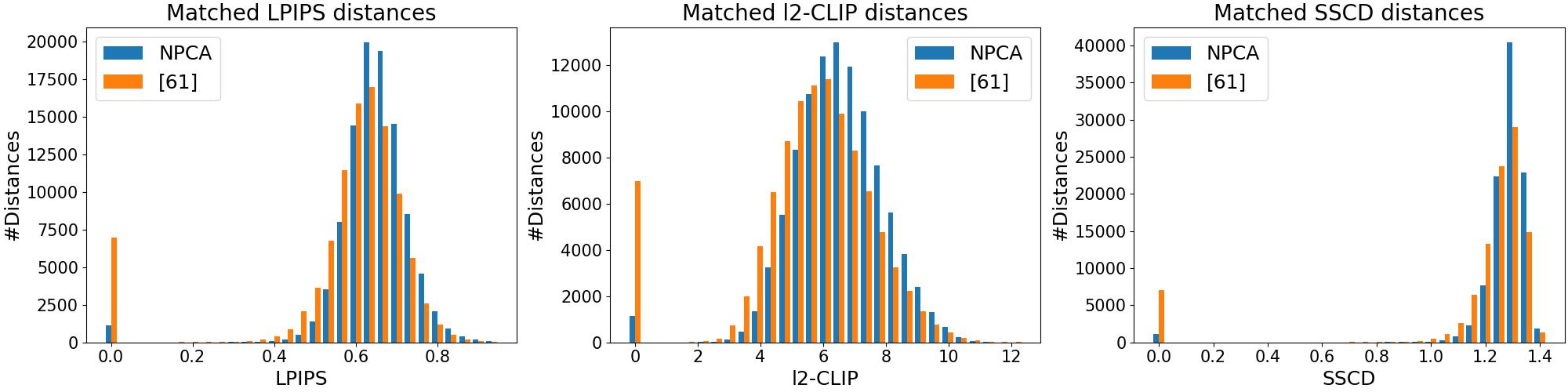}
    \caption{\label{fig:hist-matched}\textbf{Histograms of matched distances:} We consider the 5 maximally activating training images for each of the top-5 NPCA components resp. top-5 neurons of \cite{Singla2022salient}. For each of these images (and each of the components/neurons) we find the best matching maximally activating training image of a different component/neuron. %
    We call the distances of the corresponding images ``matched distance'' and plot these for three different metrics: the neural perceptual metric \cite{zhang2018perceptual}, the l2-distance of clip embeddings \cite{eyuboglu2022domino}, the SSCD distance used for image copy detection \cite{pizzi2022self}.}
\end{figure*}

\section{Comparison to Neural Features of \cite{Singla2022salient}}\label{app:comparison}
In this section, we quantitatively and qualitatively evaluate the diversity of the subpopulations detected by our top-5 NPCA components and the top-5 neurons of \cite{Singla2022salient}, respectively, on all 1000 classes. To enable a direct comparison, we consider the NPCA components computed on their robust ResNet-50 \cite{Singla2022salient} and compare them to the top-5 neurons detected in \cite{Singla2022salient} for the same model.

A larger variety of subpopulations increases annotation efficiency as duplicates of the same semantic feature do not add more information. Moreover, the probability of missing a (harmful) spurious feature is higher when several of the top components/neurons correspond to the same feature because it might drop out of the set selected for human supervision. 
\begin{figure}[ht]
    \setlength{\tabcolsep}{0.1em}
    \centering
    \normalsize
    \includegraphics[width=\linewidth]{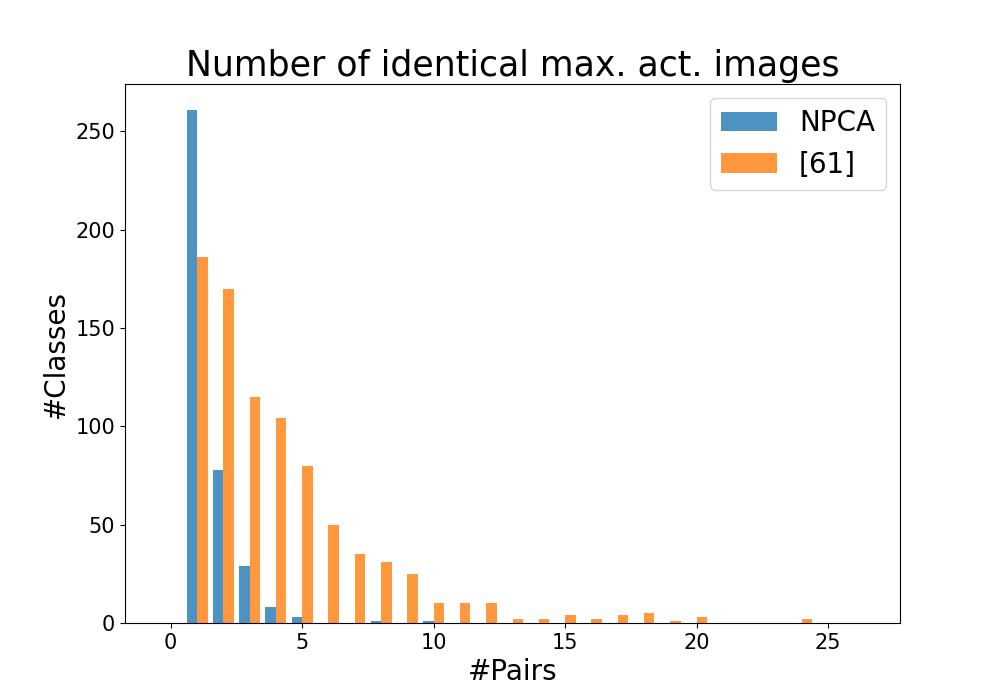}
    \caption{\label{fig:hist-zeros}\textbf{Histogram of identical maximal activating training images for the top-5 NPCA components resp. the top-5 neurons of \cite{Singla2022salient} for the robust model used by \cite{Singla2022salient}} We observe that for NPCA  only a few components have identical maximally activating training images and if it happens in the majority of cases only a single maximal activating training image of two components is identical. In contrast, \cite{Singla2022salient} has a long tail, meaning that several maximal activating training images of top-5 neurons are identical. This confirms that maximally activated neurons do not necessarily capture different semantic concepts. This is different for NPCA as the orthogonality constraint of PCA enforces to explore different directions/regions in feature space.}
\end{figure}

 For the quantitative evaluation, we measure the perceptual similarity of the subpopulations based on the \emph{matched distances} of the maximally activating training images:\\
Let $I_i^{(k)}$ be the set of the maximally activating images of component $i$ for class $k$ and let $d$ be a perceptual metric. We define the \emph{matched distance} $d_{m}\left(x, I_j^{(k)}\right)$ of an image $x \in I_i^{(k)}$ to a component $I_j^{(k)}, j\neq i,$ as the minimal (perceptual) distance between $x$ and the five images in $I_j^{(k)}$:
\[d_m\left(x, I_j^{(k)}\right) := \min_{x'\in I_j^{(k)}} d(x, x').\]
\indent For every class $k$, we consider the top-5 components/neurons, each represented by the 5 maximally activating training images of the corresponding class $k$. We compute the matched distances of every image to the other four components, resulting in a total of 100000 matched distances. Fig.~\ref{fig:hist-matched} shows histograms for the perceptual metrics LPIPS \cite{zhang2018perceptual}, l2-distance of the CLIP embeddings \cite{eyuboglu2022domino}, and the SSCD distance \cite{pizzi2022self}. In all three of them, the distribution corresponding to the NPCA components is shifted to the right compared to the one of the neural features which supports the hypothesis that NPCA detects more diverse subpopulations. However, as the purpose of the available perceptual metrics is to measure perturbations of the same image \cite{zhang2018perceptual} or to detect (close) copies \cite{pizzi2022self}, there is no guarantee that these distances are still meaningful for larger values. Nevertheless, the maximally activating training images of the top-5 neurons of \cite{Singla2022salient} also have a larger amount of distances equal to zero which corresponds to identical images. The histogram in Fig.~\ref{fig:hist-zeros} shows how many of these identical pairs occur per class. For the NPCA, identical maximally training images occur less often and for most of the cases there is only one per class. The method of \cite{Singla2022salient} produces several identical images per class much more frequently which confirms that single neurons do not necessarily capture different concepts. Due to the orthogonality constraint of PCA, the NPCA components explore different directions in feature space and detect subpopulations with less overlap.

\input{worst12_compare_figure_1}
\input{worst12_compare_figure_2}

To visualize these results, Fig.~\ref{fig:worst12_compare_figure_1} and Fig.~\ref{fig:worst12_compare_figure_2} show the three worst classes with respect to identical maximally activating training images for \cite{Singla2022salient} and NPCA, respectively. In both figures, we show the 5 maximally training images together with their GradCam images per top-5 neuron of \cite{Singla2022salient} on the left resp. top-5 NPCA component on the right.

\textbf{Worst three classes -- \cite{Singla2022salient} -- (28/24/24 identical pairs):}\\
Considering the classes "badger" and "king snake", the five neurons capture almost equivalent semantic features which are all labeled as "core". For  "badger", three neurons have exactly the same 5 maximally activating images. A similar pattern holds for the class "groom". Here, four of the subpopulations found by \cite{Singla2022salient} are very similar. Note that three of those were labeled as "spurious feature" and one as "core feature". This is a consequence of the use of a majority vote of the human labelers as a selection criterium. Our stricter criterium (unanimous vote) prevents such inconsistencies. %
While, in all three cases, one of the NPCA components ("badger": first comp., "king snake": first comp., "groom": third comp.) resembles the features detected by the neurons, the remaining components capture a much more diverse set of features. In the case of "king snake", the fourth NPCA components detects, with respect to the criteria of \cite{Singla2022salient}, a spurious feature (hands). For "groom", we even have three NPCA components corresponding to spurious features (bride, ceiling/lights, trees/bushes). This illustrates how a lack of diversity in the subpopulations of \cite{Singla2022salient} for some classes can hinder the detection of spurious features.

\textbf{Worst three classes -- NPCA -- (10/8/5 identical pairs):}\\
First, we note that regarding the number of identical pairs for the ``worst cases'' of NPCA, there exist many classes for the top-5 neurons which have similar number of identical pairs.

The class "lumbermill" is the worst class for NPCA. One can see two pairs of duplicate subpopulations (trunks and trunks/planks) that overlap to large extent with each other (which is an absolute outlier for NPCA). However, interestingly this subpopulation which is clearly spurious for ``lumbermill'' as there are no particular features for "lumbermill" is not present in the top-5 neurons of \cite{Singla2022salient}. The second worst class for NPCA is "barbershop" with two largely overlapping components showing store/house fronts. However, there are also three neurons that capture this semantic feature. In fact the number of identical pairs, NPCA 8, \cite{Singla2022salient} 6, is not so different. In the case of "English foxhound", two NPCA components correspond to a white fence which is a spurious feature. While the neurons' subpopulations are unique for this class, they do not detect the spurious fence as GradCam for the neuron mainly activates on the dog. 

Overall, these examples demonstrate that the problem of identical maximally activating images is a lot less severe for the NPCA components.

\section{Extended Qualitative Evaluation}\label{app:qualitative}
Here, we extend our qualitative evaluation of the found harmful spurious components from Figure \ref{fig:spurious-exp}. Concretely, for each such pair (class $k$, component $l$) we show in Fig.~\ref{fig:spurious-exp2} and \ref{fig:spurious-exp3}: i) random training images from class $k$, 
ii) NPFV of the component $l$ together with the most activating images of $\alpha^{(k)}_l$, and iii) examples of images that display only the spurious feature but no class features and are \textbf{incorrectly}  classified by four ImageNet classifiers as class $k$.
\input{spurious_examples_2.tex}

\input{spurious_examples_3.tex}
\section{Random samples from our ``Spurious ImageNet'' dataset}
\label{app:dataset}

To visualize our ``Spurious ImageNet'' dataset, for each of the 100 classes in our dataset, we show 4 randomly drawn images (out of the 75 in total) in Fig.~\ref{fig:app_spurious_random1} and \ref{fig:app_spurious_random2}. We also provide a label for the spurious feature shown in brackets. We again highlight that none of the images contains the actual class object. 

\input{appendix_spurious_imagenet.tex}

\section{Generating the spurious feature to change predictions}\label{app:dvces_flips}
In this section we show how one can adapt the recent method ``Diffusion Visual Counterfactual Explanations'' \cite{Augustin2022Diffusion} to generate the spurious feature on a given image without changing the overall structure of the image.
We first introduce the necessary notation. We denote by $n(x) = \frac{x}{\norm{x}_2}$ for $x \neq 0$,  the normalization of a vector by its $l_2$ norm and the confidence of the robust ResNet50 classifier in a target class $k$ as
\[p_{\text{robust},\psi} : [0, 1]^d \rightarrow (0, 1), \;\;\; x \mapsto \frac{e^{f_{\text{robust},\psi,k}(x)}}{\sum_{i=1}^K e^{f_{\text{robust},\psi,i}(x)}}.\]
Here, $f_{\text{robust},\psi}: [0, 1]^d \rightarrow \mathbb{R}^K$ are the logits of the robust classifier, and $f_{\text{robust},\psi,k}(x)$ denotes the logit of class $k$.

To automatically add spurious features to any given image, we adapt a recently proposed method Diffusion Visual Counterfactual Explanations (DVCEs) \cite{Augustin2022Diffusion}, where at a step $t$ the shifted mean $\mu_t$ is of the form
\begin{align*}
\label{eq:guided_regularized}
     g_\mathrm{update} &=  C_c g_c - C_d g_d + C_a g_a, \\
     \mu_{t} &= \mu_{\theta}(x_{t},t) \\& + \Sigma_{\theta}(x_{t},t) \norm{\mu_{\theta}(x_{t},t)}_2 g_\mathrm{update}, \\
     p(x_{t-1}|x_{t},\hat{x},k) &= 
     \mathcal{N}(\mu_{t} ,\Sigma_{\theta}(x_{t},t)),
 \end{align*}
 where $g_c:= n(\nabla_{x_t} \log p_{\text{robust},\psi}\big(k|f_\mathrm{dn}(x_t, t)\big))$ is the normalized gradient of the adversarially robust classifier, $g_d := n(\nabla_{x_t} d(\hat{x}, f_\mathrm{dn}(x_t, t)))$ - normalized gradient of the distance term. We add as additional guidance $g_a :=  n(\nabla_{x_t}  \alpha^{(k)}_j(f_\mathrm{dn}(x_t, t)))$ - the normalized gradient of the contribution $\alpha^{(k)}_j$ of the $j$-th neural PCA component to the logit of class $k$.
 As the derivative of the diffusion models, relies on noisy updates, and the classifier has not been trained on such inputs, \cite{Augustin2022Diffusion} propose to use the denoised sample $\hat{x}_0=f_\mathrm{dn}(x, t)$ of the noisy input $x_t$ as an input to the classifier. 
  Intuitively,  at every step $t$ of the generative denoising process, the method of \cite{Augustin2022Diffusion} follows i) the direction $g_a$ that increases the contribution of neural PCA component $j$ (corresponding to a  desired spurious feature) of class $k$ to the logit $f_k(x)$ of this class, ii) the direction $g_c$ that increases the confidence of the classifier in the class $k$, and iii) the direction $g_d$ that decreases the distance to the original image $\hat{x}$.
 
 In our experiments, we set $d(x, y):=\norm{x - y}_1$ following \cite{Augustin2022Diffusion} and coefficients as follows: $C_c = 0.1, C_d = 0.35, C_a = 0.05.$ With these parameters, we generate the desired DVCEs in Fig.~\ref{fig:dvces_flips}. There, using minimal realistic perturbations to the original image we can change the prediction of the classifier in the target class $k$ with high confidence. Moreover, these perturbations introduce only \textit{harmful} spurious features to the image and not class-specific features e.g. for freight car the DVCE generates graffiti but no features of a freight car.
 
  This happens, because, as has been shown qualitatively in Fig.~\ref{fig:spurious-exp} and quantitatively in Fig.~\ref{fig:barplot}, this classifier has learned to associate class ``fireboat'' with the spurious feature ``water jet'', ``freight car'' - with ``graffiti'', ``flagpole'' with a flag without the pole and mostly with ``US flag'', and ``hard disc'' - with ``label'', and therefore introducing only these \textit{harmful} spurious features is enough to increase the confidence in the target class $k$ significantly.

\input{panels_tables/dvces_flips}

\end{document}

%% file: teaser_figure.tex
\begin{figure}[ht!]
    \setlength{\tabcolsep}{.1em}
    \centering
    \begin{tabular}{cccc}
    \multicolumn{4}{c}{{\red  Spurious Features} in {\blue Training Data}}\\
    {\red bird feeder} & {\red graffiti} & {\red eucalyptus} & {\red label} %
    \\
    {\includegraphics[width=0.105\textwidth]{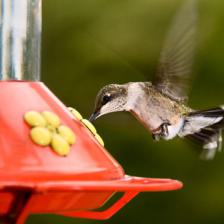}}
    &{\includegraphics[width=0.105\textwidth]{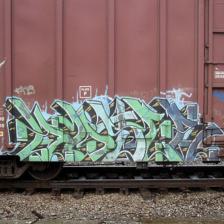}}
    &{\includegraphics[width=0.105\textwidth]{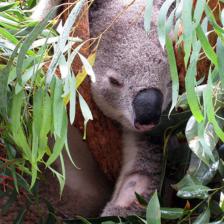}}
    &{\includegraphics[width=0.105\textwidth]{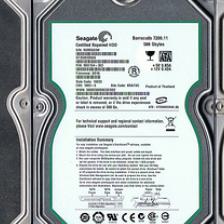}}\\[-1pt]
     {\blue Hummingbird} &  {\blue Freight Car} &  {\blue Koala} &  {\blue Hard Disc} %
    \\[4pt]
    \multicolumn{4}{c}{Images from the web with {\bf {\red spurious feature}} }\\
    \multicolumn{4}{c}{but {\blue \textbf{no class features}} classified as {\blue class} below}\\
    {\includegraphics[width=0.105\textwidth]{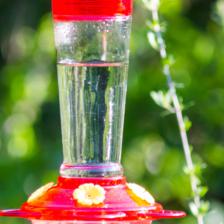}}
    &{\includegraphics[width=0.105\textwidth]{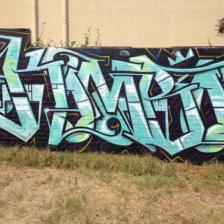}}
    &{\includegraphics[width=0.105\textwidth]{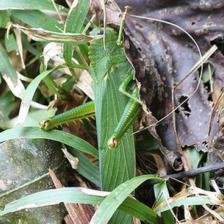}}
    &\includegraphics[width=0.105\textwidth]{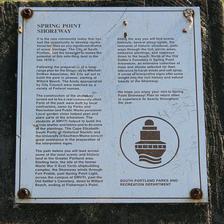}\\
   {\includegraphics[width=0.105\textwidth]{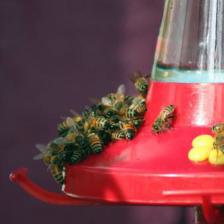}}
    &{\includegraphics[width=0.105\textwidth]{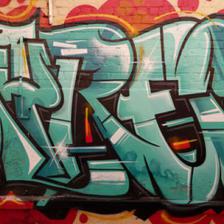}}
    &{\includegraphics[width=0.105\textwidth]{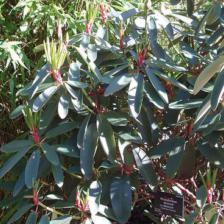}}
    &%
    {\includegraphics[width=0.105\textwidth]{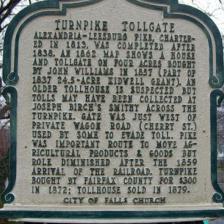}}
    \\[-1pt]
   {\blue  Hummingbird}  & {\blue Freight Car} & {\blue Koala} &  {\blue Hard Disc} %
    \end{tabular}
    \caption{\label{fig:teaser}\textbf{Top:} Examples of spurious features found via our neural PCA components but not in previous study \cite{Singla2022salient}. \textbf{Bottom:} We validate our spurious features by mining images from the web showing \textbf{only the spurious feature but not the class}. They are classified by four ImageNet models as the corresponding class. Some of them even contain ImageNet classes (bees on feeder, grasshopper in leaves).}
    \vspace{-4mm}
\end{figure}

%% file: top5_compare_hummingbird_figure.tex
\begin{figure}
    \setlength{\tabcolsep}{.01em}

    \centering
    \begin{tabular}{p{0.02\textwidth} ccccc }%
        &\multicolumn{5}{c}{Top 5 neural PCA components (ours)} \\%&&& \multicolumn{5}{c}{Top 5 mean feature components \cite{Singla2022salient}}\\
        \raisebox{1.5\normalbaselineskip}[0pt][0pt]{\rotatebox[origin=c]{90}{NPFV}}
        &{\includegraphics[width=0.09\textwidth]{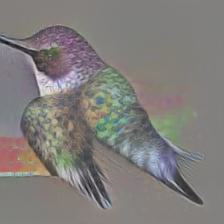}}
        &{\includegraphics[width=0.09\textwidth]{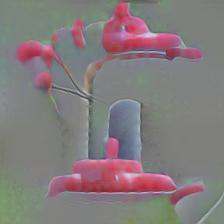}}
        &{\includegraphics[width=0.09\textwidth]{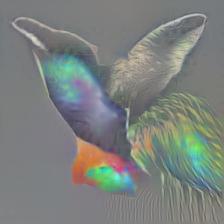}}
        &{\includegraphics[width=0.09\textwidth]{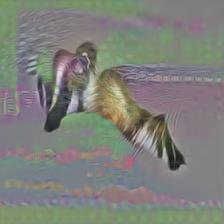}}
        &{\includegraphics[width=0.09\textwidth]{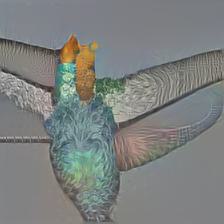}}
        \\
       \raisebox{-.25\normalbaselineskip}[0pt][0pt]{\rotatebox[origin=tl]{90}{Max. act.}}
        &{\includegraphics[width=0.09\textwidth]{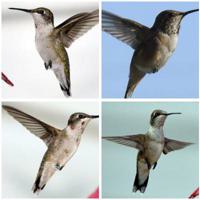}}
        &{\includegraphics[width=0.09\textwidth]{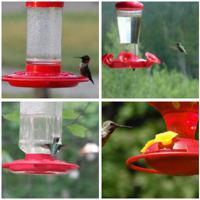}}
        &{\includegraphics[width=0.09\textwidth]{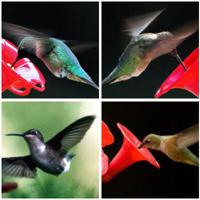}}
        &{\includegraphics[width=0.09\textwidth]{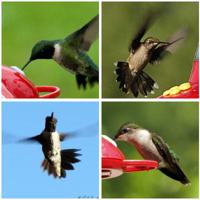}}
        &{\includegraphics[width=0.09\textwidth]{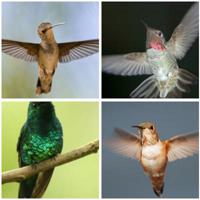}} 
        \\
                \raisebox{-.2\normalbaselineskip}[0pt][0pt]{\rotatebox[origin=tl]{90}{Heatmap}}
       &{\includegraphics[width=0.09\textwidth]{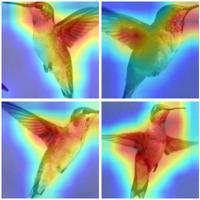}}
        &{\includegraphics[width=0.09\textwidth]{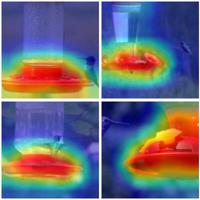}}
        &{\includegraphics[width=0.09\textwidth]{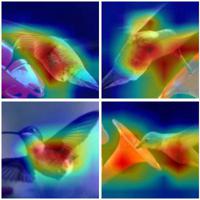}}
        &{\includegraphics[width=0.09\textwidth]{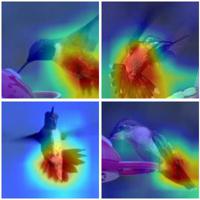}}
        &{\includegraphics[width=0.09\textwidth]{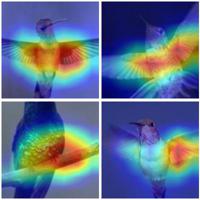}} 
\\
    \end{tabular}
    \caption{\label{fig:PCA-vs-TopFeature}\textbf{Top 5 Neural PCA components %
    for class hummingbird:} first row shows our neural PCA feature visualization (NPFV), second row shows four most activating training images of each NPCA component and the last row GradCAM for the NPCA component. Our NPCA components capture different subpopulations in the training data. Comp. 2 is identified as spurious feature ``bird feeder'', see NPFV and most activating training images (see also Fig. \ref{fig:spurious-exp}). }
    \vspace{-3mm}
\end{figure}

%% file: spurious_examples.tex
\begin{figure*}
    \setlength{\tabcolsep}{0.1em}
    \centering
    \normalsize
    \begin{tabular}{c c c c c p{0.02\textwidth} c c c c c}
        \multicolumn{5}{l}{{\blue\textbf{Hummingbird} }-  Random train. images (\textbf{confidence}\,\,/\,$\alpha_k$)} 
        &&\multicolumn{5}{c}{Images with spurious \textbf{bird feeder} but \textbf{no hummingbird}}\\ 
        {\includegraphics[width=0.095\textwidth]{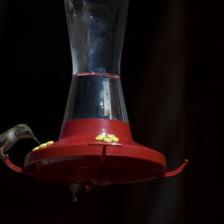}} 
        &{\includegraphics[width=0.095\textwidth]{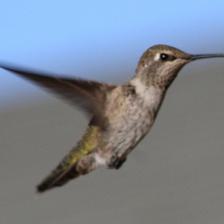}}
        &{\includegraphics[width=0.095\textwidth]{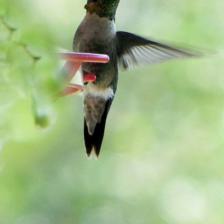}}
        &{\includegraphics[width=0.095\textwidth]{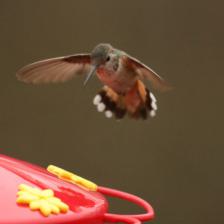}} 
        &{\includegraphics[width=0.095\textwidth]{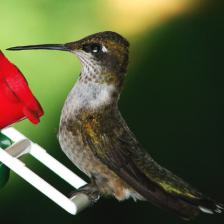}}
        &&{\includegraphics[width=0.095\textwidth]{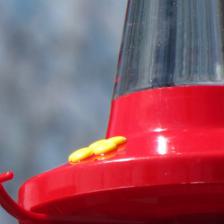}}%
        &{\includegraphics[width=0.095\textwidth]{figures/img/hummingbird/pc_idx_1_img_5_-hummingbird_feeder-_bee.jpg}}
        &{\includegraphics[width=0.095\textwidth]{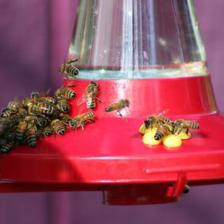}}
        &{\includegraphics[width=0.095\textwidth]{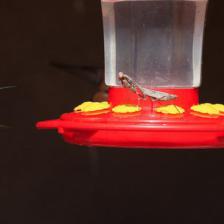}}
        &{\includegraphics[width=0.095\textwidth]{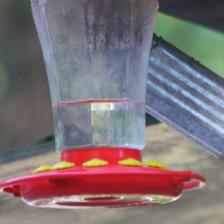}}\\
        $\mathbf{0.93}$\,/\,$1.7$
        &$\mathbf{1.00}$\,/\,$-0.9$
        &$\mathbf{0.96}$\,/\,$-1.0$
        &$\mathbf{0.99}$\,/\,$2.2$
        &$\mathbf{1.00}$\,/\,$1.54$
        &&$\mathbf{0.94}$\,/\,$5.7$
        &$\mathbf{0.94}$\,/\,$3.4$
        &$\mathbf{0.82}$\,/\,$2.9$ 
        &$\mathbf{0.91}$\,/\,$5.6$
        & $\mathbf{0.91}$\,/\,$4.7$\\
        \multicolumn{1}{c}{\NPFV-2} & \multicolumn{4}{c}{Max. activating train. images - NPCA Comp. 2} 
        && \multicolumn{5}{c}{all classified as \textbf{humming bird} by four ImageNet models}\\
        {\includegraphics[width=0.095\textwidth]{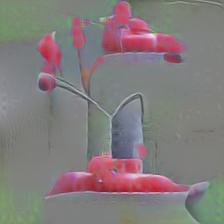}}& 
        {\includegraphics[width=0.095\textwidth]{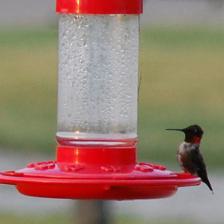}}&  {\includegraphics[width=0.095\textwidth]{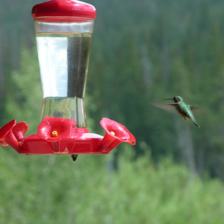}}&  {\includegraphics[width=0.095\textwidth]{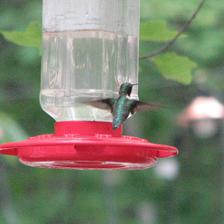}} &
        {\includegraphics[width=0.095\textwidth]{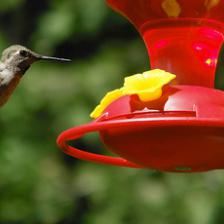}}&&
        {\includegraphics[width=0.095\textwidth]{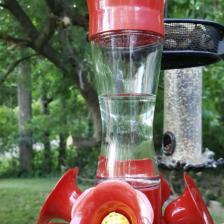}}&  {\includegraphics[width=0.095\textwidth]{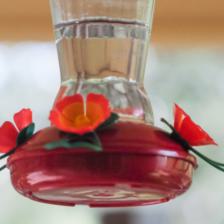}}&  {\includegraphics[width=0.095\textwidth]{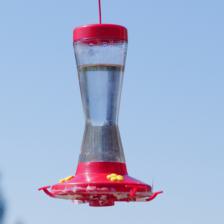}}&  {\includegraphics[width=0.095\textwidth]{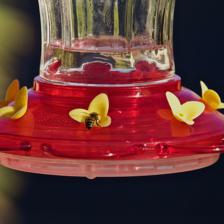}}&{\includegraphics[width=0.095\textwidth]{figures/img/hummingbird/wiki_3.jpg}}\\
        $\mathbf{1.00}$\,/\,$9.7$
        &$\mathbf{1.00}$\,/\,$7.5$
        &$\mathbf{1.00}$\,/\,$5.9$
        &$\mathbf{1.00}$\,/\,$5.6$
        &$\mathbf{1.00}$\,/\,$5.6$
        &&$\mathbf{0.86}$\,/\,$4.3$
        &$\mathbf{0.81}$\,/\,$3.5$
        &$\mathbf{0.89}$\,/\,$3.3$
        &$\mathbf{0.89}$\,/\,$3.8$
        &$\mathbf{0.78}$\,/\,$5.91$\\
        \hline
         \multicolumn{5}{l}{{\blue\textbf{Freight car}} - Random train. images (\textbf{confidence}\,\,/\,$\alpha_k$)}
         &&\multicolumn{5}{c}{Images with spurious \textbf{grafitti} but \textbf{no freight car}}\\ 
         {\includegraphics[width=0.095\textwidth]{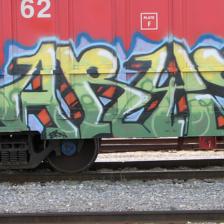}} 
         &{\includegraphics[width=0.095\textwidth]{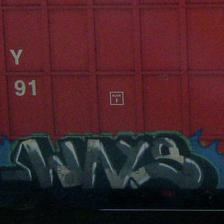}}
         &{\includegraphics[width=0.095\textwidth]{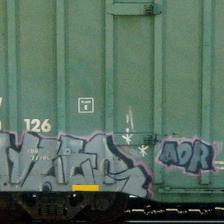}}
         &{\includegraphics[width=0.095\textwidth]{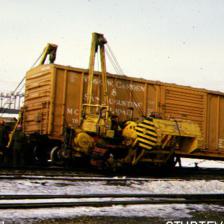}} 
         &{\includegraphics[width=0.095\textwidth]{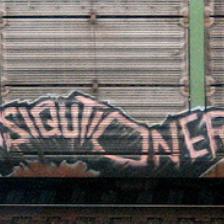}}
         &&{\includegraphics[width=0.095\textwidth]{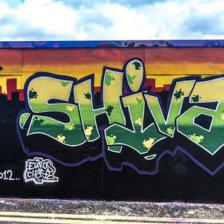}}
         &{\includegraphics[width=0.095\textwidth]{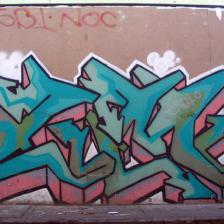}}
         &{\includegraphics[width=0.095\textwidth]{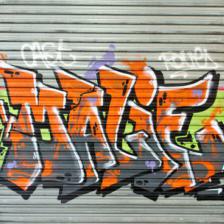}}
         &{\includegraphics[width=0.095\textwidth]{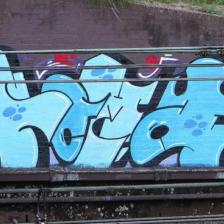}}
         &{\includegraphics[width=0.095\textwidth]{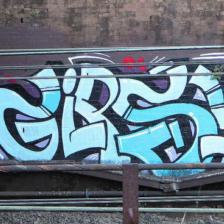}}\\
         $\mathbf{1.00}$\,/\,$7.7$
         &$\mathbf{1.00}$\,/\,$3.4$
         &$\mathbf{1.00}$\,/\,$6.3$
         &$\mathbf{0.98}$\,/\,$-0.8$
         &$\mathbf{1.00}$\,/\,$4.2$
         &&$\mathbf{0.97}$\,/\,$3.5$ 
         &$\mathbf{0.79}$\,/\,$4.0$
         &$\mathbf{0.81}$\,/\,$2.4$
         &$\mathbf{0.85}$\,/\,$3.2$
         &$\mathbf{0.88}$\,/\,$2.6$\\
         \multicolumn{1}{c}{\NPFV-1} & \multicolumn{4}{c}{Max. activating train. images - NPCA Comp. 1} 
         && \multicolumn{5}{c}{all classified as \textbf{freight car} by four ImageNet models}\\
         {\includegraphics[width=0.095\textwidth]{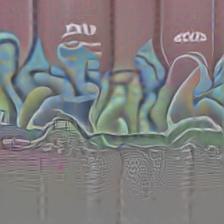}}& 
         {\includegraphics[width=0.095\textwidth]{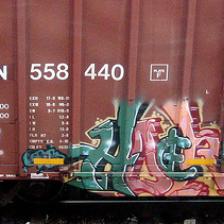}}&  {\includegraphics[width=0.095\textwidth]{figures/img/freight_car/act_pc_0_max_1.jpg}}&  {\includegraphics[width=0.095\textwidth]{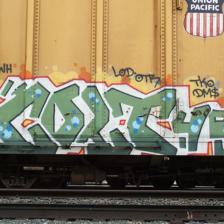}} &
         {\includegraphics[width=0.095\textwidth]{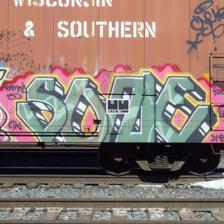}}& &
         {\includegraphics[width=0.095\textwidth]{figures/img/freight_car/151.jpg}}&  {\includegraphics[width=0.095\textwidth]{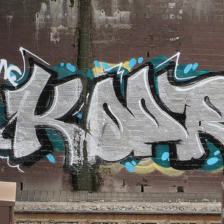}}&  {\includegraphics[width=0.095\textwidth]{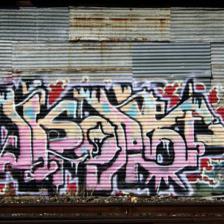}}&  {\includegraphics[width=0.095\textwidth]{figures/img/freight_car/182.jpg}}&{\includegraphics[width=0.095\textwidth]{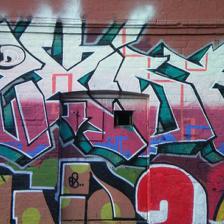}}\\
         $\mathbf{1.00}$\,/\,$12.1$
         &$\mathbf{1.00}$\,/\,$10.9$
         &$\mathbf{1.00}$\,/\,$10.4$
         &$\mathbf{1.00}$\,/\,$10.2$
         &$\mathbf{1.00}$\,/\,$10.2$
         && $\mathbf{0.85}$\,/\,$2.5$
         &$\mathbf{0.86}$\,/\,$2.5$
         &$\mathbf{0.90}$\,/\,$2.3$
         &$\mathbf{0.82}$\,/\,$2.2$
         &$\mathbf{0.87}$\,/\,$2.2$\\
         \hline
          \multicolumn{5}{l}{{\blue\textbf{Koala}} -  Random train. images (\textbf{confidence}\,\,/\,$\alpha_k$)}  
         &&\multicolumn{5}{c}{Images with spurious \textbf{eucalyptus/plants} but \textbf{no koala}}\\ 
         {\includegraphics[width=0.095\textwidth]{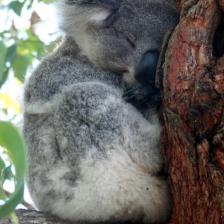}} 
         &{\includegraphics[width=0.095\textwidth]{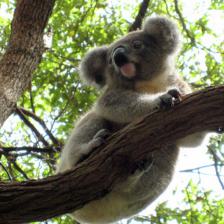}}
         &{\includegraphics[width=0.095\textwidth]{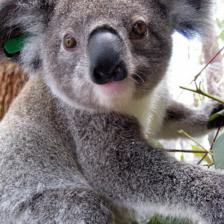}}
         &{\includegraphics[width=0.095\textwidth]{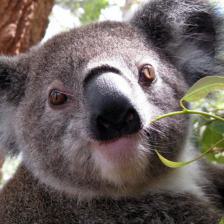}} 
         &{\includegraphics[width=0.095\textwidth]{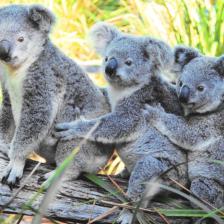}}
         &&{\includegraphics[width=0.095\textwidth]{figures/img/koala/1.jpg}}
         &{\includegraphics[width=0.095\textwidth]{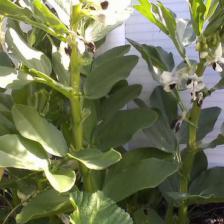}}
         &{\includegraphics[width=0.095\textwidth]{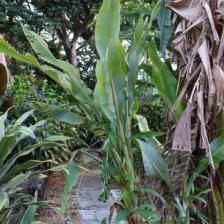}}
         &{\includegraphics[width=0.095\textwidth]{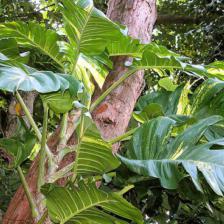}}
         &{\includegraphics[width=0.095\textwidth]{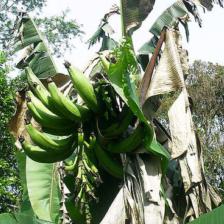}}\\
         $\mathbf{1.00}$\,/\,$0.77$
         &$\mathbf{0.87}$\,/\,$2.4$
         &$\mathbf{1.00}$\,/\,$0.4$
         &$\mathbf{1.00}$\,/\,$0.0$
         &$\mathbf{0.95}$\,/\,$0.5$
         &&$\mathbf{0.49}$\,/\,$3.5$
         &$\mathbf{0.61}$\,/\,$3.1$
         &$\mathbf{0.36}$\,/\,$3.1$
         &$\mathbf{0.36}$\,/\,$3.1$
         &$\mathbf{0.69}$\,/\,$2.7$\\
         \multicolumn{1}{c}{\NPFV-3} & \multicolumn{4}{c}{Max. activating train. images - NPCA Comp. 3} && \multicolumn{5}{c}{all classified as \textbf{koala} by four ImageNet models}\\
         {\includegraphics[width=0.095\textwidth]{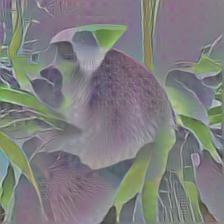}}& 
         {\includegraphics[width=0.095\textwidth]{figures/img/koala/act_pc_1_max_0.jpg}}&  {\includegraphics[width=0.095\textwidth]{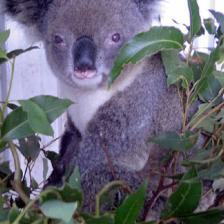}}&  {\includegraphics[width=0.095\textwidth]{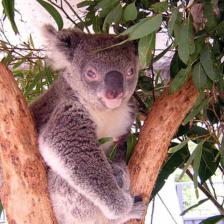}} &
         {\includegraphics[width=0.095\textwidth]{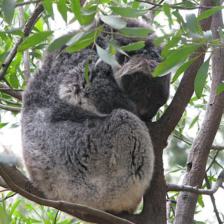}}&&
         {\includegraphics[width=0.095\textwidth]{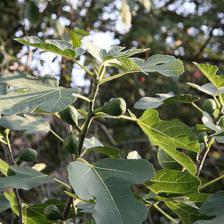}}&
         {\includegraphics[width=0.095\textwidth]{figures/img/koala/71.jpg}}&  {\includegraphics[width=0.095\textwidth]{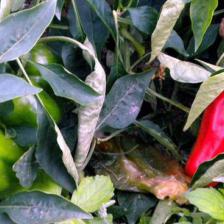}}&  %
         {\includegraphics[width=0.095\textwidth]{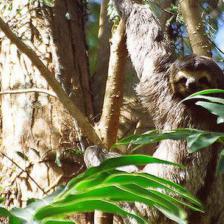}}&{\includegraphics[width=0.095\textwidth]{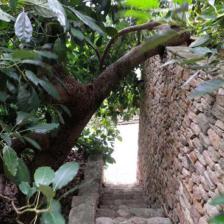}}\\
         $\mathbf{1.00}$\,/\,$5.5$
         &$\mathbf{1.00}$\,/\,$4.6$
         &$\mathbf{1.00}$\,/\,$4.5$
         &$\mathbf{1.00}$\,/\,$4.4$
         &$\mathbf{0.86}$\,/\,$4.3$
         &&$\mathbf{0.56}$\,/\,$2.6$
         &$\mathbf{0.62}$\,/\,$2.5$
         &$\mathbf{0.74}$\,/\,$2.5$
         &$\mathbf{0.72}$\,/\,$2.4$
         &$\mathbf{0.66}$\,/\,$2.2$\\
         \hline
         \multicolumn{5}{l}{{\blue\textbf{Fireboat}} - Random train. images (\textbf{confidence}\,\,/\,$\alpha_k$)} 
         &&\multicolumn{5}{c}{Images with spurious \textbf{water jet} but \textbf{no fireboat}}\\ 
         {\includegraphics[width=0.095\textwidth]{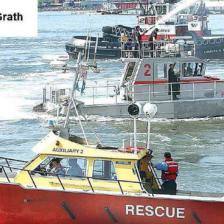}} 
         &{\includegraphics[width=0.095\textwidth]{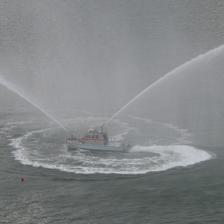}}
         &{\includegraphics[width=0.095\textwidth]{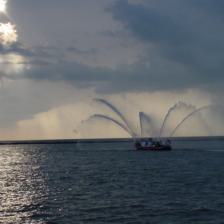}}
         &{\includegraphics[width=0.095\textwidth]{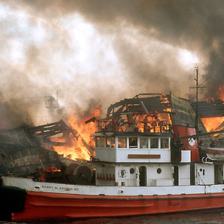}} 
         &{\includegraphics[width=0.095\textwidth]{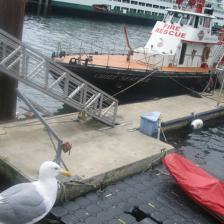}}
         &&{\includegraphics[width=0.095\textwidth]{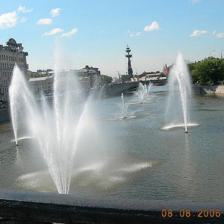}}
         &{\includegraphics[width=0.095\textwidth]{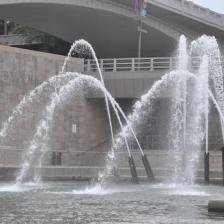}}
         &{\includegraphics[width=0.095\textwidth]{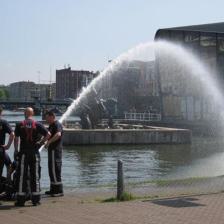}}
         &{\includegraphics[width=0.095\textwidth]{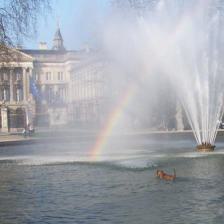}}
         &{\includegraphics[width=0.095\textwidth]{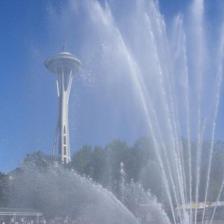}}\\
         $\mathbf{0.88}$\,/\,$-1.1$
         &$\mathbf{0.95}$\,/\,$0.5$
         &$\mathbf{0.12}$\,/\,$-0.6$
         &$\mathbf{0.84}$\,/\,$-0.2$
         &$\mathbf{0.02}$\,/\,$-1.1$
         &&$\mathbf{0.63}$\,/\,$2.3$
         &$\mathbf{0.71}$\,/\,$2.2$
         &$\mathbf{0.84}$\,/\,$2.2$
         &$\mathbf{0.65}$\,/\,$2.1$
         &$\mathbf{0.63}$\,/\,$1.9$\\
         \multicolumn{1}{c}{\NPFV-2} & \multicolumn{4}{c}{Max. activating train. images - NPCA Comp. 2} && \multicolumn{5}{c}{all classified as \textbf{fireboat} by four ImageNet models}\\
         {\includegraphics[width=0.095\textwidth]{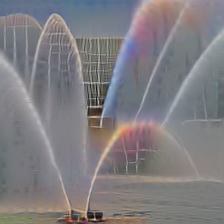}}& 
         {\includegraphics[width=0.095\textwidth]{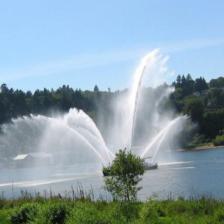}}&  {\includegraphics[width=0.095\textwidth]{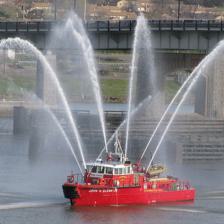}}&  {\includegraphics[width=0.095\textwidth]{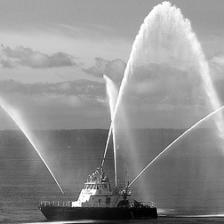}} &
         {\includegraphics[width=0.095\textwidth]{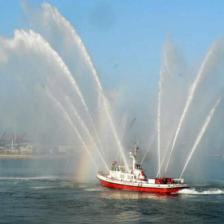}}&&
         {\includegraphics[width=0.095\textwidth]{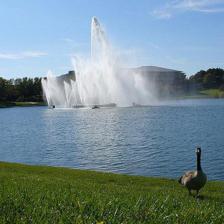}}&  {\includegraphics[width=0.095\textwidth]{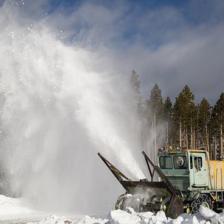}}&  {\includegraphics[width=0.095\textwidth]{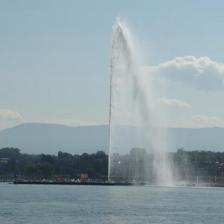}}&{\includegraphics[width=0.095\textwidth]{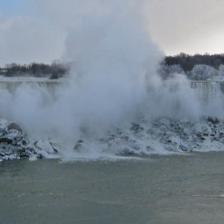}}&
         {\includegraphics[width=0.095\textwidth]{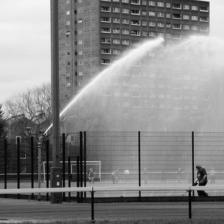}}\\
         $\mathbf{1.00}$\,/\,$5.5$
         &$\mathbf{0.42}$\,/\,$4.1$
         &$\mathbf{1.00}$\,/\,$4.0$
         &$\mathbf{1.00}$\,/\,$3.9$
         &$\mathbf{1.00}$\,/\,$3.9 $
         &&$\mathbf{0.53}$\,/\,$1.9$
         &$\mathbf{0.79}$\,/\,$1.9$
         &$\mathbf{0.78}$\,/\,$1.8$
         &$\mathbf{0.79}$\,/\,$1.8$
         &$\mathbf{0.76}$\,/\,$1.8$
    \end{tabular}%
    \caption{\label{fig:spurious-exp} \textbf{Spurious features (ImageNet):} found by human labeling of our neural PCA components. For each class we show 5 random train. images (top left), the neural PCA Feature Visual. (NPFV) and 4 most activating train. images for the spurious feature component (bottom left). Right: four ImageNet models classify images \textbf{showing only the spurious feature but no class object} as this class.} %
\end{figure*}

%% file: top5_compare_their_resnet50_hummingbird_figure.tex
\begin{figure}
    \setlength{\tabcolsep}{.01em}

    \centering
    \begin{tabular}{ p{0.02\textwidth} ccccc }%
        &\multicolumn{5}{c}{Top 5 neural PCA components (ours)} \\
       \raisebox{-.25\normalbaselineskip}[0pt][0pt]{\rotatebox[origin=tl]{90}{Max. act.}}
        & {\includegraphics[width=0.09\textwidth]{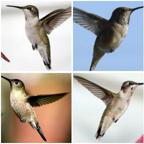}}
        &\,{\includegraphics[width=0.09\textwidth]{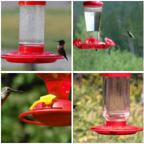}}
        & \,{\includegraphics[width=0.09\textwidth]{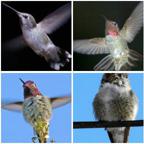}}
        & \,{\includegraphics[width=0.09\textwidth]{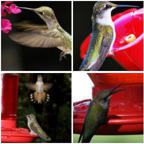}}
        & \,{\includegraphics[width=0.09\textwidth]{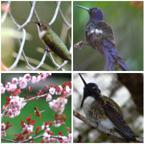}} 
        \\
        \raisebox{-.2\normalbaselineskip}[0pt][0pt]{\rotatebox[origin=tl]{90}{Heatmap}}
       &\,{\includegraphics[width=0.09\textwidth]{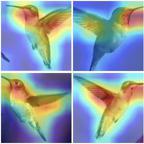}}
        &\,{\includegraphics[width=0.09\textwidth]{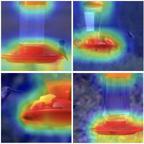}}
        &\,{\includegraphics[width=0.09\textwidth]{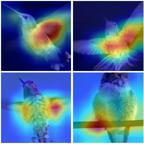}}
        &\,{\includegraphics[width=0.09\textwidth]{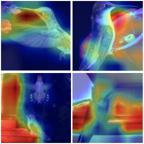}}
        &\,{\includegraphics[width=0.09\textwidth]{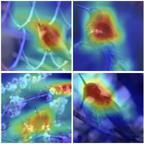}} 

\\
&\multicolumn{5}{c}{Top 5 neurons (theirs)} \\
       \raisebox{-.25\normalbaselineskip}[0pt][0pt]{\rotatebox[origin=tl]{90}{Max. act.}}
        &{\includegraphics[width=0.09\textwidth]{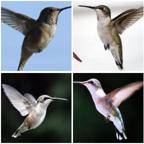}}
        &{\includegraphics[width=0.09\textwidth]{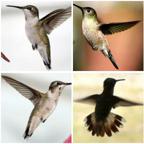}}
        &{\includegraphics[width=0.09\textwidth]{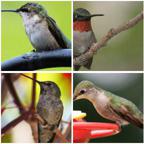}}
        &{\includegraphics[width=0.09\textwidth]{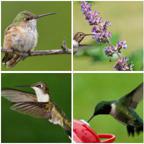}}
        &{\includegraphics[width=0.09\textwidth]{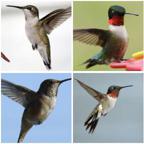}} 

        \\
        \raisebox{-.2\normalbaselineskip}[0pt][0pt]{\rotatebox[origin=tl]{90}{Heatmap}}
       &{\includegraphics[width=0.09\textwidth]{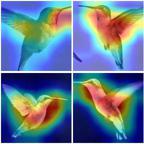}}
        &{\includegraphics[width=0.09\textwidth]{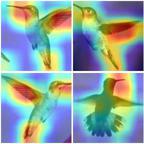}}
        &{\includegraphics[width=0.09\textwidth]{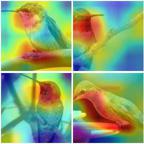}}
        &{\includegraphics[width=0.09\textwidth]{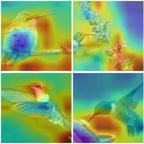}}
        &{\includegraphics[width=0.09\textwidth]{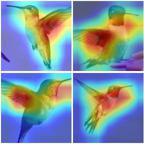}} 
    \end{tabular} \vspace{-2mm}
    \caption{\label{fig:comparison}\textbf{Comparison of top-5 NPCA and top-5 neurons %
    for class hummingbird for the robust model of \cite{Singla2022salient}:} Our NPCA components identify diverse subpopulations in the training set whereas the top neurons show similar ones and the spurious bird feeder is not detected, see also App. \ref{app:koala-PCA}.} 
    \vspace{-3mm}
\end{figure}

%% file: top5_compare_figure.tex
\begin{figure*}
    \setlength{\tabcolsep}{.01em}

    \centering
    \begin{tabular}{ p{0.02\textwidth} ccccc p{0.02\textwidth} p{0.02\textwidth} ccccc}
       \multicolumn{13}{c}{{\blue \textbf{Koala}}}\\
        &\multicolumn{5}{c}{Top-5 neural PCA components (ours)} &&& \multicolumn{5}{c}{Top-5 neurons \cite{Singla2022salient}}\\
        \raisebox{1.5\normalbaselineskip}[0pt][0pt]{\rotatebox[origin=c]{90}{NPFV}}
        &{\includegraphics[width=0.09\textwidth]{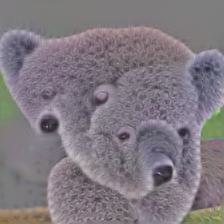}}
        &{\includegraphics[width=0.09\textwidth]{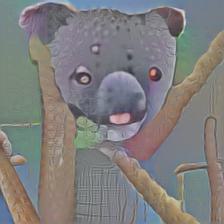}}
        &{\includegraphics[width=0.09\textwidth]{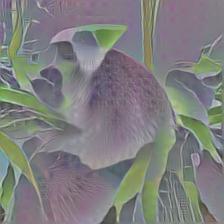}}
        &{\includegraphics[width=0.09\textwidth]{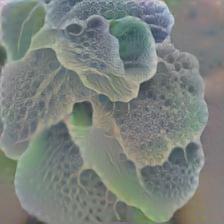}}
        &{\includegraphics[width=0.09\textwidth]{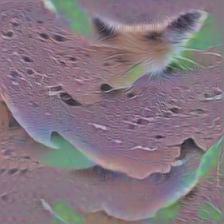}}
        &&
         \raisebox{0\normalbaselineskip}[0pt][0pt]{\rotatebox[origin=tl]{90}{FeatAtt}}
        &{\includegraphics[width=0.09\textwidth]{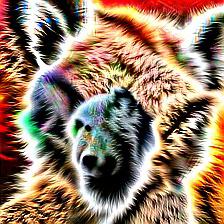}}
        &{\includegraphics[width=0.09\textwidth]{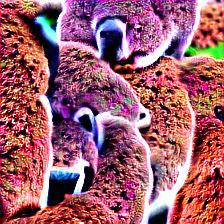}}
        &{\includegraphics[width=0.09\textwidth]{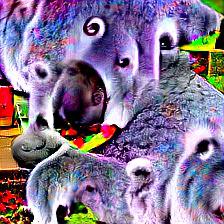}}
        &{\includegraphics[width=0.09\textwidth]{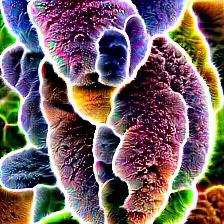}}
        &{\includegraphics[width=0.09\textwidth]{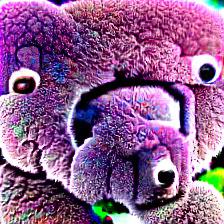}}\\
       \raisebox{-.25\normalbaselineskip}[0pt][0pt]{\rotatebox[origin=tl]{90}{Max. act.}}
        &{\includegraphics[width=0.09\textwidth]{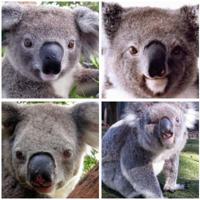}}
        &{\includegraphics[width=0.09\textwidth]{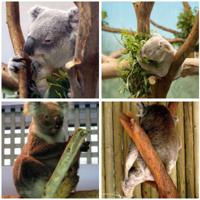}}
        &{\includegraphics[width=0.09\textwidth]{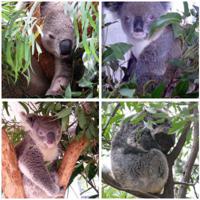}}
        &{\includegraphics[width=0.09\textwidth]{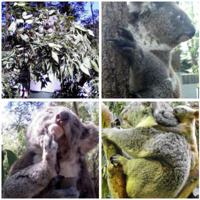}}
        &{\includegraphics[width=0.09\textwidth]{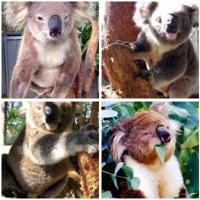}} &&
       \raisebox{-.25\normalbaselineskip}[0pt][0pt]{\rotatebox[origin=tl]{90}{Max. act.}}
        &{\includegraphics[width=0.09\textwidth]{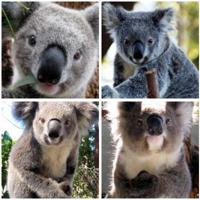}}
        &{\includegraphics[width=0.09\textwidth]{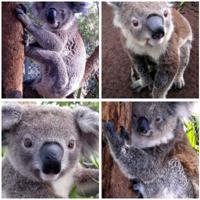}}
        &{\includegraphics[width=0.09\textwidth]{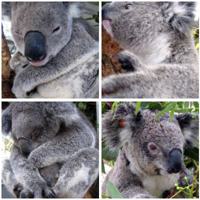}}
        &{\includegraphics[width=0.09\textwidth]{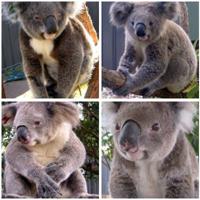}}
        &{\includegraphics[width=0.09\textwidth]{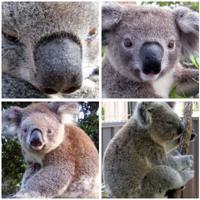}}\\
                \raisebox{-.2\normalbaselineskip}[0pt][0pt]{\rotatebox[origin=tl]{90}{Heatmap}}
       &{\includegraphics[width=0.09\textwidth]{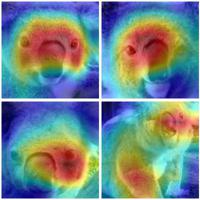}}
        &{\includegraphics[width=0.09\textwidth]{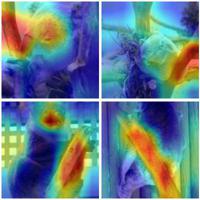}}
        &{\includegraphics[width=0.09\textwidth]{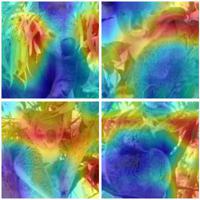}}
        &{\includegraphics[width=0.09\textwidth]{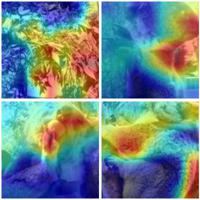}}
        &{\includegraphics[width=0.09\textwidth]{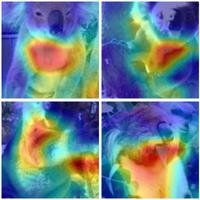}} &&
       \raisebox{-.2\normalbaselineskip}[0pt][0pt]{\rotatebox[origin=tl]{90}{Heatmap}}
        &{\includegraphics[width=0.09\textwidth]{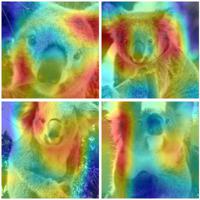}}
        &{\includegraphics[width=0.09\textwidth]{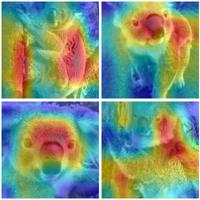}}
        &{\includegraphics[width=0.09\textwidth]{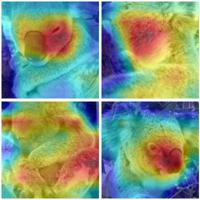}}
        &{\includegraphics[width=0.09\textwidth]{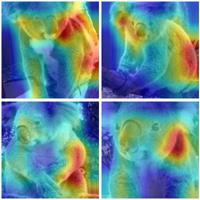}}
        &{\includegraphics[width=0.09\textwidth]{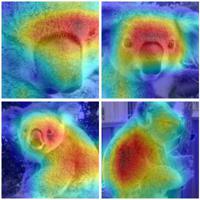}}\\
          \hline\\
        \multicolumn{13}{c}{{\blue \textbf{Indigo Bunting}}}\\
        &\multicolumn{5}{c}{Top-5 neural PCA components (ours)} &&& \multicolumn{5}{c}{Top-5 neurons \cite{Singla2022salient}}\\
        \raisebox{1.5\normalbaselineskip}[0pt][0pt]{\rotatebox[origin=c]{90}{NPFV}}
        &{\includegraphics[width=0.09\textwidth]{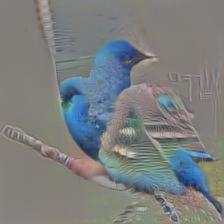}}
        &{\includegraphics[width=0.09\textwidth]{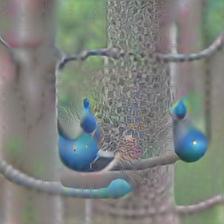}}
        &{\includegraphics[width=0.09\textwidth]{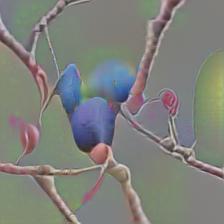}}
        &{\includegraphics[width=0.09\textwidth]{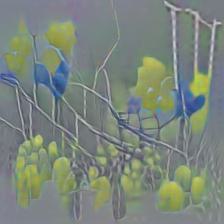}}
        &{\includegraphics[width=0.09\textwidth]{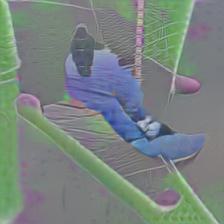}}
        &&
         \raisebox{0\normalbaselineskip}[0pt][0pt]{\rotatebox[origin=tl]{90}{FeatAtt}}
        &{\includegraphics[width=0.09\textwidth]{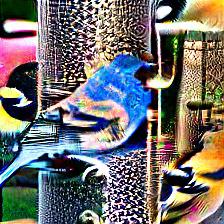}}
        &{\includegraphics[width=0.09\textwidth]{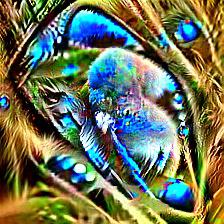}}
        &{\includegraphics[width=0.09\textwidth]{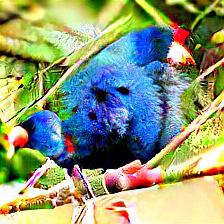}}
        &{\includegraphics[width=0.09\textwidth]{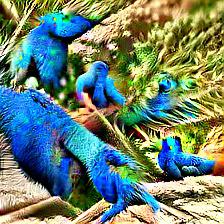}}
        &{\includegraphics[width=0.09\textwidth]{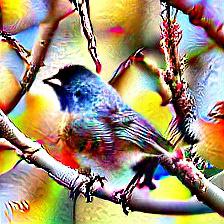}}\\
       \raisebox{-.25\normalbaselineskip}[0pt][0pt]{\rotatebox[origin=tl]{90}{Max. act.}}
        &{\includegraphics[width=0.09\textwidth]{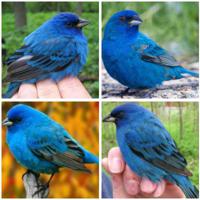}}
        &{\includegraphics[width=0.09\textwidth]{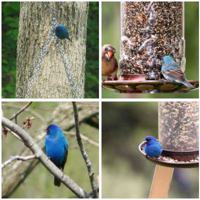}}
        &{\includegraphics[width=0.09\textwidth]{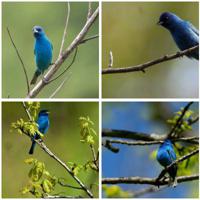}}
        &{\includegraphics[width=0.09\textwidth]{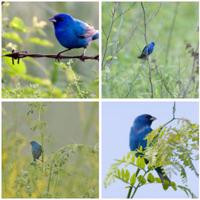}}
        &{\includegraphics[width=0.09\textwidth]{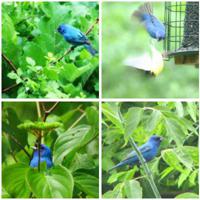}} &&
       \raisebox{-.25\normalbaselineskip}[0pt][0pt]{\rotatebox[origin=tl]{90}{Max. act.}}
        &{\includegraphics[width=0.09\textwidth]{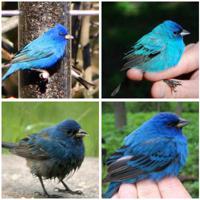}}
        &{\includegraphics[width=0.09\textwidth]{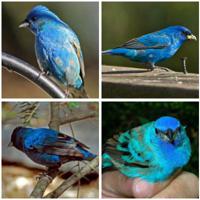}}
        &{\includegraphics[width=0.09\textwidth]{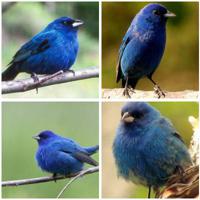}}
        &{\includegraphics[width=0.09\textwidth]{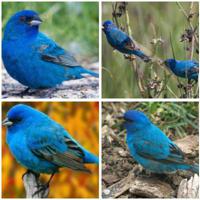}}
        &{\includegraphics[width=0.09\textwidth]{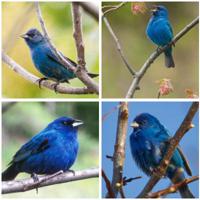}}\\
        \raisebox{-.2\normalbaselineskip}[0pt][0pt]{\rotatebox[origin=tl]{90}{Heatmap}}
       &{\includegraphics[width=0.09\textwidth]{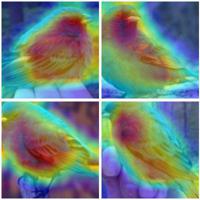}}
        &{\includegraphics[width=0.09\textwidth]{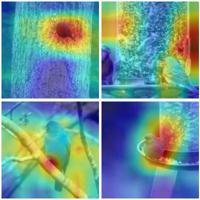}}
        &{\includegraphics[width=0.09\textwidth]{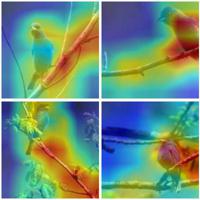}}
        &{\includegraphics[width=0.09\textwidth]{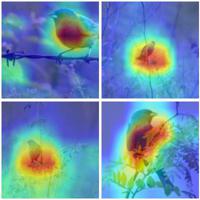}}
        &{\includegraphics[width=0.09\textwidth]{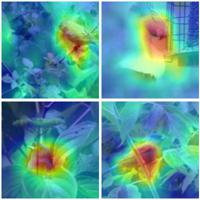}} &&
       \raisebox{-.2\normalbaselineskip}[0pt][0pt]{\rotatebox[origin=tl]{90}{Heatmap}}
        &{\includegraphics[width=0.09\textwidth]{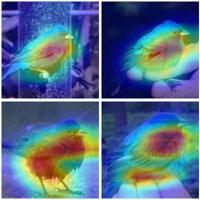}}
        &{\includegraphics[width=0.09\textwidth]{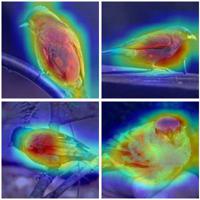}}
        &{\includegraphics[width=0.09\textwidth]{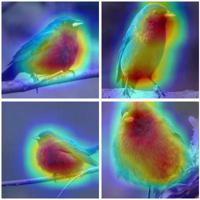}}
        &{\includegraphics[width=0.09\textwidth]{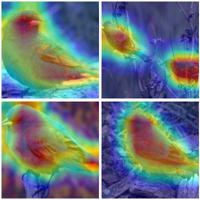}}
        &{\includegraphics[width=0.09\textwidth]{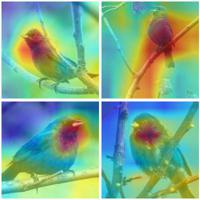}}\\
        \hline\\
         \multicolumn{13}{c}{{\blue \textbf{Mountain Bike}}}\\
        &\multicolumn{5}{c}{Top-5 neural PCA components (ours)} &&& \multicolumn{5}{c}{Top-5 neurons \cite{Singla2022salient}}\\
        \raisebox{1.5\normalbaselineskip}[0pt][0pt]{\rotatebox[origin=c]{90}{NPFV}}
        &{\includegraphics[width=0.09\textwidth]{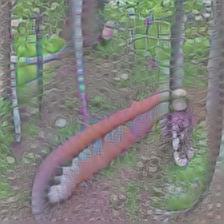}}
        &{\includegraphics[width=0.09\textwidth]{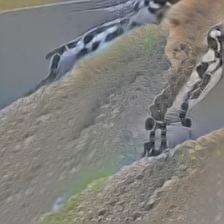}}
        &{\includegraphics[width=0.09\textwidth]{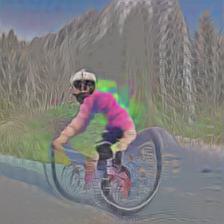}}
        &{\includegraphics[width=0.09\textwidth]{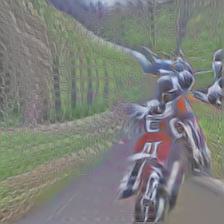}}
        &{\includegraphics[width=0.09\textwidth]{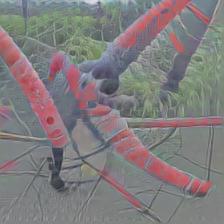}}
        &&
         \raisebox{0\normalbaselineskip}[0pt][0pt]{\rotatebox[origin=tl]{90}{FeatAtt}}
        &{\includegraphics[width=0.09\textwidth]{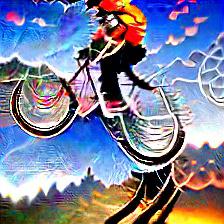}}
        &{\includegraphics[width=0.09\textwidth]{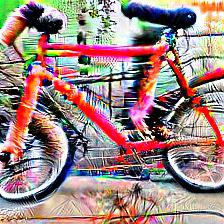}}
        &{\includegraphics[width=0.09\textwidth]{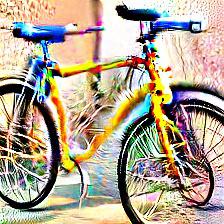}}
        &{\includegraphics[width=0.09\textwidth]{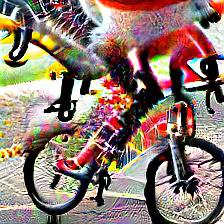}}
        &{\includegraphics[width=0.09\textwidth]{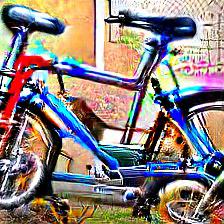}}\\
       \raisebox{-.25\normalbaselineskip}[0pt][0pt]{\rotatebox[origin=tl]{90}{Max. act.}}
        &{\includegraphics[width=0.09\textwidth]{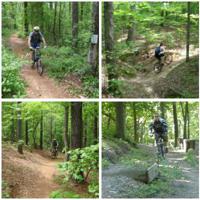}}
        &{\includegraphics[width=0.09\textwidth]{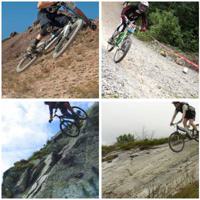}}
        &{\includegraphics[width=0.09\textwidth]{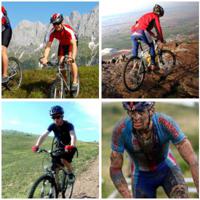}}
        &{\includegraphics[width=0.09\textwidth]{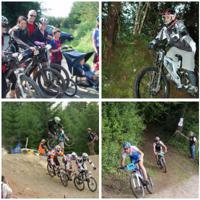}}
        &{\includegraphics[width=0.09\textwidth]{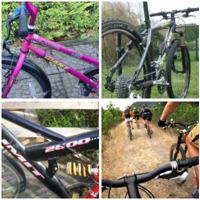}} &&
       \raisebox{-.25\normalbaselineskip}[0pt][0pt]{\rotatebox[origin=tl]{90}{Max. act.}}
        &{\includegraphics[width=0.09\textwidth]{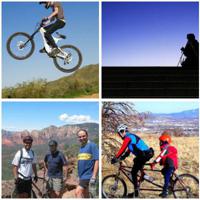}}
        &{\includegraphics[width=0.09\textwidth]{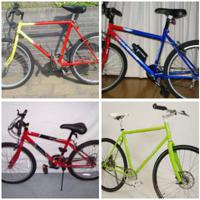}}
        &{\includegraphics[width=0.09\textwidth]{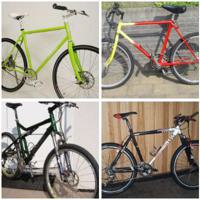}}
        &{\includegraphics[width=0.09\textwidth]{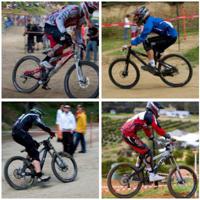}}
        &{\includegraphics[width=0.09\textwidth]{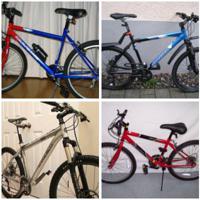}}\\
        \raisebox{-.2\normalbaselineskip}[0pt][0pt]{\rotatebox[origin=tl]{90}{Heatmap}}
       &{\includegraphics[width=0.09\textwidth]{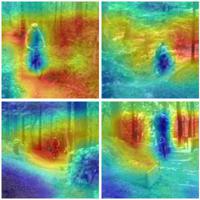}}
        &{\includegraphics[width=0.09\textwidth]{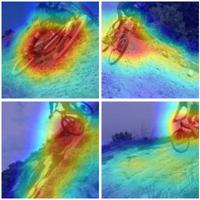}}
        &{\includegraphics[width=0.09\textwidth]{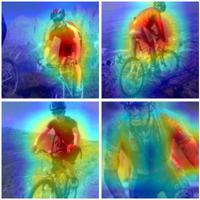}}
        &{\includegraphics[width=0.09\textwidth]{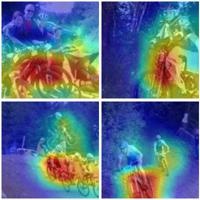}}
        &{\includegraphics[width=0.09\textwidth]{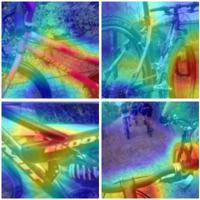}} &&
       \raisebox{-.2\normalbaselineskip}[0pt][0pt]{\rotatebox[origin=tl]{90}{Heatmap}}
        &{\includegraphics[width=0.09\textwidth]{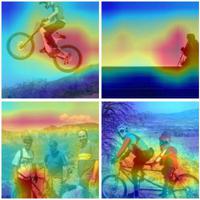}}
        &{\includegraphics[width=0.09\textwidth]{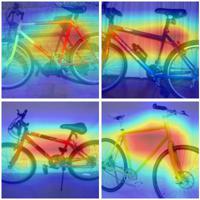}}
        &{\includegraphics[width=0.09\textwidth]{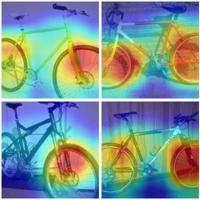}}
        &{\includegraphics[width=0.09\textwidth]{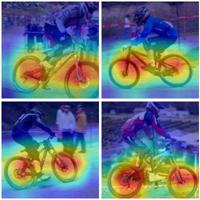}}
        &{\includegraphics[width=0.09\textwidth]{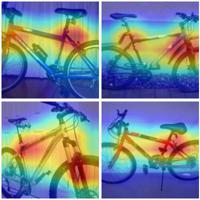}}\\

    \end{tabular}
    \caption{\label{fig:PCA-vs-TopFeature-extra}\textbf{Neural PCA feature components vs neural features of \cite{Singla2022salient} for different classes (computed on our own multiple-norm robust model vs. their $l_2$-robust model):} First row: NPCA feature visualization (NPFV) of our top-5 NPCA components (left), and the feature attacks of the top-5 neurons of \cite{Singla2022salient} (right). Second row: four most activating training images of the components/neurons. Last row: GradCAM for the NPCA components (left) and the neural activation map of \cite{Singla2022salient} (right). For these three classes \cite{Singla2022salient} report no spurious feature. As in Figure \ref{fig:PCA-vs-TopFeature} our NPCA components are capturing different subpopulations in the training data. Our NPCA Component 3 of Koala shows prominent leaves in the NPFV and neural PCA GradCAM heatmaps and is identified as spurious, similar for our component 3 of Indigo Bunting showing twigs in the NPFV and in the heatmaps, and component 1 for mountain bike where the forest appears in the NPFV and is active in the heatmap. The feature attack of \cite{Singla2022salient} generates an image similar to the most activating training image which adds less new information. In contrast, our NPFV allows to identify which features the component has picked up. }
\end{figure*}

%% file: appendix_mauc_table_new.tex
\begin{table*}
\centering
\small
\begin{tabular}{|c|c|c|c|c| c |c|c|c|c|c|} 
\cline{1-5}\cline{7-11}
 & \multicolumn{2}{c|}{Original} & \multicolumn{2}{c|}{SpuFix} && &\multicolumn{2}{c|}{Original} & \multicolumn{2}{c|}{SpuFix} \\
\cline{1-5}\cline{7-11}
Name & Acc. & Spu. & Acc. & Spu. && Name & Acc. & Spu. & Acc. & Spu. \\
\cline{1-5}\cline{7-11}
\multicolumn{5}{|c|}{ImageNet1k} && \multicolumn{5}{|c|}{1B Instagram\cite{yalniz2019billion}}\\
\cline{1-5}
\cline{7-11}
Rob. ResNet50 & 57.4\% & 0.630 & 56.8\% &\textbf{0.763} && ResNeXt101 SSL \cite{yalniz2019billion} & 83.3\% & 0.872 & 83.3\% &\textbf{0.875}\\
Rob. ResNet50\cite{Singla2022salient} & 57.9\% & 0.651 & 57.2\% &\textbf{0.764} && ResNeXt50 SSL \cite{yalniz2019billion} & 82.2\% & 0.857 & 82.1\% &\textbf{0.862}\\
ResNet50\cite{HeZhaRen2015} & 81.2\% & 0.851 & 81.2\% &\textbf{0.860} && ResNet50 SSL \cite{yalniz2019billion} & 81.2\% & 0.850 & 80.8\% &\textbf{0.865}\\
\cline{7-11}
ResNet101\cite{HeZhaRen2015} & 82.8\% & 0.748 & 82.8\% &\textbf{0.795} && \multicolumn{5}{|c|}{ImageNet21kFT1k}\\
\cline{7-11}
ResNeXt50 32x4d\cite{xie2017aggregated} & 82.0\% & 0.783 & 82.0\% &\textbf{0.808} && ResNetV2-152 BiT\cite{KolesnikovEtAl2019} & 84.9\% & 0.895 & 84.9\% &\textbf{0.900}\\
ResNeXt101 32x8d\cite{xie2017aggregated} & 79.3\% & 0.797 & 79.2\% &\textbf{0.811} && ResNetV2-50 BiT\cite{KolesnikovEtAl2019} & 84.0\% & 0.887 & 84.0\% &\textbf{0.895}\\
ResNeXt101 64x4d\cite{xie2017aggregated} & 83.2\% & 0.779 & 83.2\% &\textbf{0.786} && EfficientNetV2-M\cite{tan2021efficientnetv2} & 86.0\% & 0.892 & 86.0\% &\textbf{0.897}\\
EfficientNet B5 RA\cite{cubuk2020randaugment} & 83.8\% & 0.829 & 83.8\% &\textbf{0.833} && EfficientNetV2-L\cite{tan2021efficientnetv2} & 86.8\% & 0.893 & 86.8\% &\textbf{0.898}\\
EfficientNet B5 AP\cite{xu2021adversarial} & 84.3\% & 0.828 & 84.2\% &\textbf{0.832} && ConvNeXt-B\cite{liu2022convnet} & 86.3\% & 0.892 & 86.3\% &\textbf{0.895}\\
EfficientNet B6 AA\cite{tan2019efficientnet} & 84.1\% & 0.830 & 84.1\% &\textbf{0.836} && ConvNeXt-L\cite{liu2022convnet} & 87.0\% & 0.910 & 87.0\% &\textbf{0.913}\\
EfficientNet B6 AP\cite{xu2021adversarial} & 84.8\% & 0.831 & 84.8\% &\textbf{0.838} && ConvNeXt-XL\cite{liu2022convnet} & 87.3\% & 0.908 & 87.3\% &\textbf{0.913}\\
EfficientNet B7 RA\cite{cubuk2020randaugment} & 84.9\% & 0.834 & 84.9\% &\textbf{0.839} && ConvNeXtV2-B\cite{woo2023convnext} & 87.6\% & 0.907 & 87.6\% &\textbf{0.911}\\
EfficientNet B7 AP\cite{xu2021adversarial} & 85.1\% & 0.826 & 85.1\% &\textbf{0.831} && ConvNeXtV2-L\cite{woo2023convnext} & 88.2\% & 0.905 & 88.2\% &\textbf{0.907}\\
EfficientNetV2-M\cite{tan2021efficientnetv2} & 85.2\% & 0.846 & 85.2\% &\textbf{0.856} && ConvNeXtV2-H\cite{woo2023convnext} & 88.7\% & 0.919 & 88.7\% &\textbf{0.923}\\
EfficientNetV2-L\cite{tan2021efficientnetv2} & 85.7\% & 0.851 & 85.7\% &\textbf{0.860} && DeiT3-S\textbackslash16\cite{touvron2022deit3} & 83.1\% & 0.845 & 83.1\% &\textbf{0.860}\\
ConvNeXt-B\cite{liu2022convnet} & 84.4\% & 0.802 & 84.4\% &\textbf{0.816} && DeiT3-L\textbackslash16\cite{touvron2022deit3} & 87.7\% & 0.895 & 87.7\% &\textbf{0.901}\\
ConvNeXt-L\cite{liu2022convnet} & 84.8\% & 0.803 & 84.8\% &\textbf{0.819} && Swin-B 224\cite{liu2021swin} & 85.3\% & 0.877 & 85.3\% &\textbf{0.883}\\
ConvNeXtV2-B\cite{woo2023convnext} & 85.5\% & 0.848 & 85.5\% &\textbf{0.856} && Swin-L 384\cite{liu2021swin} & 87.1\% & 0.898 & 87.1\% &\textbf{0.901}\\
ConvNeXtV2-L\cite{woo2023convnext} & 86.1\% & 0.845 & 86.1\% &\textbf{0.858} && SwinV2-L\cite{liu2022swinv2} & 87.5\% & 0.889 & 87.5\% &\textbf{0.891}\\
ConvNeXtV2-H\cite{woo2023convnext} & 86.6\% & 0.867 & 86.6\% &\textbf{0.879} && ViT-B\textbackslash16 224 & 81.8\% & 0.881 & 81.7\% &\textbf{0.889}\\
DeiT3-S\textbackslash16 224\cite{touvron2022deit3} & 81.4\% & 0.851 & 81.4\% &\textbf{0.859} && ViT-B\textbackslash16 384\cite{dosovitskiy2020image} & 84.2\% & 0.905 & 84.2\% &\textbf{0.912}\\
DeiT3-L\textbackslash16 384\cite{touvron2022deit3} & 85.8\% & 0.863 & 85.8\% &\textbf{0.877} && ViT-L\textbackslash16 $\dagger$ & 85.8\% & 0.914 & 85.8\% &\textbf{0.923}\\
ViT-B\textbackslash16 $\dagger$ & 81.1\% & 0.850 & 81.1\% &\textbf{0.859} && ViT-B\textbackslash16 $\dagger$ & 86.0\% & 0.917 & 85.9\% &\textbf{0.925}\\
VOLO-D5 512\cite{yuan2022volo} & 87.1\% & 0.882 & 87.1\% &\textbf{0.907} && BEiT-B\textbackslash16 224\cite{bao2021beit} & 85.2\% & 0.890 & 85.2\% &\textbf{0.897}\\
VOLO-D5 224\cite{yuan2022volo} & 85.4\% & 0.863 & 85.3\% &\textbf{0.890} && BEiT-L\textbackslash16\cite{bao2021beit} & 88.6\% & 0.921 & 88.6\% &\textbf{0.927}\\
\cline{1-5}
\multicolumn{5}{|c|}{JFT-300M\cite{gupta2017revisiting}} && BEiTV2-L\textbackslash16 224\cite{peng2022beitv2} & 88.4\% & 0.921 & 88.4\% &\textbf{0.925}\\
\cline{1-5}
\cline{7-11}
EfficientNet B5 NS \cite{xie2020selftraining} & 86.1\% & 0.924 & 86.1\% &\textbf{0.924} && \multicolumn{5}{|c|}{ImageNet21k}\\
\cline{7-11}
EfficientNet B6 NS \cite{xie2020selftraining} & 86.5\% & 0.875 & 86.5\% &\textbf{0.880} && ResNetV2-152 BiT\cite{KolesnikovEtAl2019} & - & 0.908 & - &\textbf{0.908}\\
EfficientNet B7 NS \cite{xie2020selftraining} & 86.8\% & 0.907 & 86.9\% &\textbf{0.912} && ResNetV2-50 BiT\cite{KolesnikovEtAl2019} & - &\textbf{0.910} & -& 0.910\\
EfficientNet L2 NS \cite{xie2020selftraining} & 88.4\% & 0.914 & 88.3\% &\textbf{0.917} && EfficientNetV2-M\cite{tan2021efficientnetv2} & - &\textbf{0.919} & -& 0.919\\
\cline{1-5}
\multicolumn{5}{|c|}{YFFC-100M} && EfficientNetV2-L\cite{tan2021efficientnetv2} & - &\textbf{0.929} & -& 0.929\\
\cline{1-5}
ResNeXt101 SSL \cite{yalniz2019billion} & 81.8\% & 0.833 & 81.8\% &\textbf{0.841} && ConvNeXt-B\cite{liu2022convnet} & - & 0.939 & - &\textbf{0.939}\\
ResNeXt50 SSL \cite{yalniz2019billion} & 80.3\% & 0.821 & 80.2\% &\textbf{0.831} && ConvNeXt-L\cite{liu2022convnet} & - &\textbf{0.943} & -& 0.943\\
ResNet50 SSL \cite{yalniz2019billion} & 79.2\% & 0.804 & 78.8\% &\textbf{0.828} && ConvNeXt-XL\cite{liu2022convnet} & - & 0.945 & - &\textbf{0.945}\\
\cline{1-5}
\multicolumn{5}{|c|}{LAION-2B\cite{schuhmann2022laion}} && Swin-B 224\cite{liu2021swin} & - & 0.808 & - &\textbf{0.808}\\
\cline{1-5}
CNeXt-B CLIP\cite{radford2021learning} $\dagger$  & 86.2\% & 0.859 & 86.2\% &\textbf{0.865} && Swin-L 384\cite{liu2021swin} & - &\textbf{0.820} & -& 0.820\\
CNeXt-L CLIP\cite{radford2021learning} $\dagger$ 224 & 87.3\% & 0.858 & 87.3\% &\textbf{0.865} && ViT-L\textbackslash16 $\dagger$ & - &\textbf{0.931} & -& 0.931\\
CNeXt-L CLIP\cite{radford2021learning} $\dagger$ 384 & 87.8\% & 0.879 & 87.9\% &\textbf{0.884} && ViT-B\textbackslash8 $\dagger$ & - &\textbf{0.931} & -& 0.931\\
ViT-L\textbackslash14 CLIP\cite{radford2021learning} 336 & 88.2\% & 0.912 & 88.2\% &\textbf{0.914} && BEiT-B\textbackslash16 224\cite{bao2021beit}  & - &\textbf{0.935} & -& 0.935\\
\cline{1-5}
\multicolumn{5}{|c|}{LAION-400M\cite{schuhmann2021laion}} && BEiT-L\textbackslash16 224\cite{bao2021beit} & - &\textbf{0.940} & -& 0.940\\
\cline{1-5}
EVA-G\textbackslash14 CLIP 336\cite{EVA} & 89.5\% & 0.911 & 89.4\% &\textbf{0.915} && BEiTV2-L\textbackslash16 224\cite{peng2022beitv2} & - & 0.951 & - &\textbf{0.951}\\
\cline{1-5}
\cline{7-11}
\multicolumn{5}{|c|}{MIM\cite{EVA}}\\
\cline{1-5}
EVA-G\textbackslash14 CLIP 560\cite{EVA} & 89.8\% & 0.919 & 89.8\% &\textbf{0.925}\\
\cline{1-5}
\end{tabular}

\caption{\label{tab:app_extended}Extended version of Tab.~\ref{tab:quantitative} from the main paper. We show ImageNet1k Accuracy (Acc.) and mean spurious AUC (Spu.) for the original model and the SpuFix version for a wide selection of state-of-the-art ImageNet classifiers, trained on: either ImageNet1k only (ImageNet1k), pre-trained on ImageNet21k and then fine-tuned on ImageNet1k (ImageNet21kFT1k), full ImageNet21k classifiers (ImageNet21k) or pre-training on a range of other datasets (JFT-300M, YFFC-100M, LAION-2B, LAION-400M, MIM, 1B Instagram). For models commonly used with different input resolutions, we state the used one at the end of the name. Models using AugReg\cite{steiner2021train} are marked with $\dagger$. The ResNext50 and ResNext101 trained with SSL \cite{yalniz2019billion} have cardinality 32 and group width 4 and 16, respectively.}
\vspace{2mm}
\end{table*}

\begin{figure*}
\centering
\includegraphics[width=\textwidth]{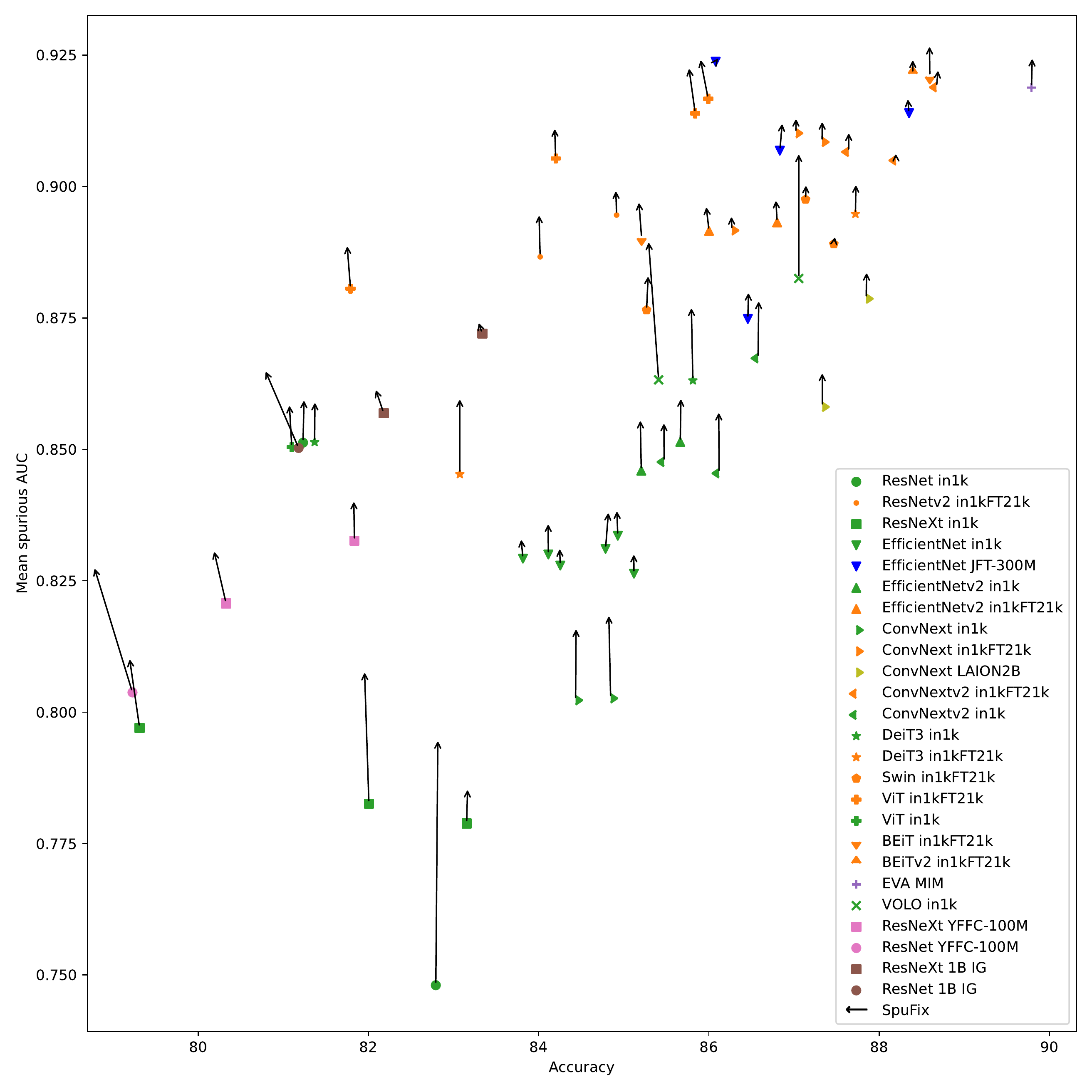}

\caption{\label{fig:app_acc_vs_auc}We plot Accuracy versus mean spurious AUC for a wide variety of SOTA ImageNet classifiers. Models that use the same architecture family use the same marker and we use color coding for the (pre-taining) datasets. For example, all models that are pre-trained on ImageNet21k and then fine-tuned are marked in yellow whereas standard ImageNet1k models are marked green. As can be observed, the addition of larger datasets like ImageNet21k, JFT-300M or LAION does decrease vulnerability to spurious features over ImageNet1k models with comparable accuracy. The arrows show the consistent improvement of mean spurious AUC after applying SpuFix while the change of accuracy is negligible for most of the models.}
\vspace{10mm}
\end{figure*}

%% file: figure_testset.tex
\begin{figure*}
    \centering
    \small
    \begin{tabular}{c c c c c}
        \hline
         \multicolumn{5}{c}{\textbf{\normalsize{Hummingbird}}}\\
         \multicolumn{5}{c}{Predicted class original/\textbf{SpuFix} (class prob. hummingbird original/\textbf{SpuFix})}\\
         \hline

         hummingbird/&
         bee/ &
         lycaenid/ &
         pop bottle/&
         hummingbird/\\

         \textbf{hummingbird} &
         \textbf{bee}&
         \textbf{lycaenid}&
         \textbf{pop bottle} &
         \textbf{hummingbird}\\
         
         \includegraphics[width=0.1\textwidth]{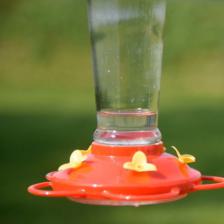}& 
         \includegraphics[width=0.1\textwidth]{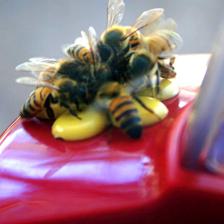}&
         \includegraphics[width=0.1\textwidth]{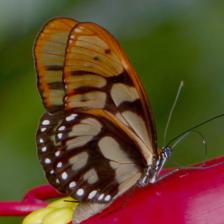}& 
         \includegraphics[width=0.1\textwidth]{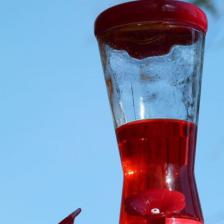}& 
         \includegraphics[width=0.1\textwidth]{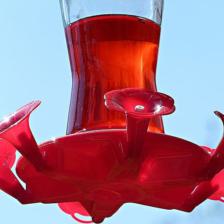} \\
        
         1.00/\textbf{0.84}&
         0.00/\textbf{0.00}&
         0.00/\textbf{0.00}&
         0.02/\textbf{0.01}&
         0.99/\textbf{0.10}\\

         \hline

         hummingbird/&
         hummingbird/&
         hummingbird/&
         hummingbird/&
         fox squirrel\\
        
         \textbf{red wine}&
         \textbf{lipstick}&
         \textbf{hummingbird}&
         \textbf{vase}&
         \textbf{fox squirrel}\\
         
         \includegraphics[width=0.1\textwidth]{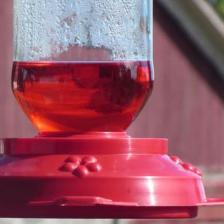}& 
         \includegraphics[width=0.1\textwidth]{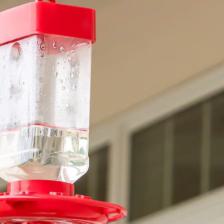}&
         \includegraphics[width=0.1\textwidth]{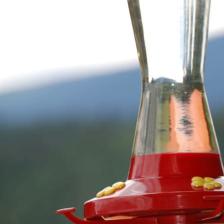}& 
         \includegraphics[width=0.1\textwidth]{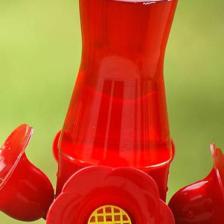}& 
         \includegraphics[width=0.1\textwidth]{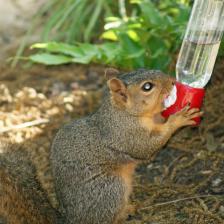} \\

        0.91/\textbf{0.07}&
        0.90/\textbf{0.03}&
        0.99/\textbf{0.18}&
        0.89/\textbf{0.02}&
        0.00/\textbf{0.00} \\

        \hline
         \multicolumn{5}{c}{\normalsize{\textbf{Gondola}}}\\
         \multicolumn{5}{c}{Predicted class original/\textbf{SpuFix} (class prob. gondola original/\textbf{SpuFix})}\\
         \hline

        gondola/&
        prison/&
        gondola/&
        gondola/&
        gondola/ \\

        \textbf{gondola}&
        \textbf{prison}&
        \textbf{prison}&
        \textbf{gondola}&
        \textbf{palace} \\
        
         \includegraphics[width=0.1\textwidth]{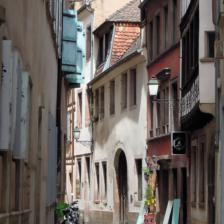}& 
         \includegraphics[width=0.1\textwidth]{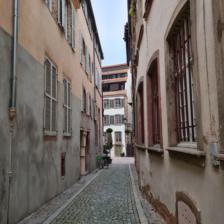}&
         \includegraphics[width=0.1\textwidth]{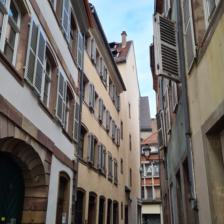}& 
         \includegraphics[width=0.1\textwidth]{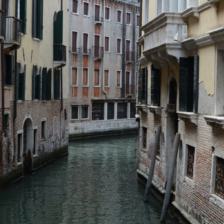}& 
         \includegraphics[width=0.1\textwidth]{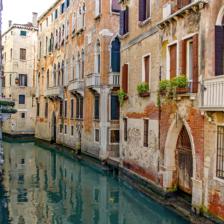} \\

        0.85/\textbf{0.16} &
        0.04/\textbf{0.04} &
        0.32/\textbf{0.06} &
        0.99/\textbf{0.10} &
        0.99/\textbf{0.12} \\
        
        \hline
        gondola/&
        gondola/&
        gondola/&
        gondola/&
        palace/\\

        \textbf{streetcar}&
        \textbf{gondola}&
        \textbf{gondola}&
        \textbf{palace}&
        \textbf{palace}\\

         \includegraphics[width=0.1\textwidth]{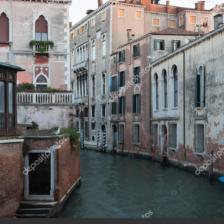}& 
         \includegraphics[width=0.1\textwidth]{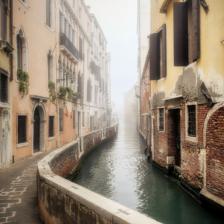}&
         \includegraphics[width=0.1\textwidth]{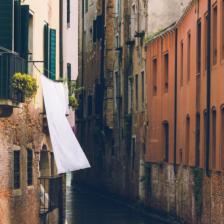}& 
         \includegraphics[width=0.1\textwidth]{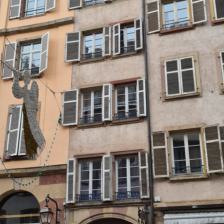}& 
         \includegraphics[width=0.1\textwidth]{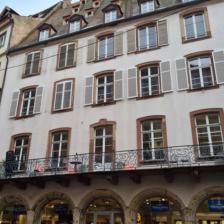} \\

        0.93/\textbf{0.01} &
        0.96/\textbf{0.16}&
        0.65/\textbf{0.16}&
        0.34/\textbf{0.01}&
        0.01/\textbf{0.01} \\

        \hline
         \multicolumn{5}{c}{\textbf{\normalsize{Flagpole}}}\\
         \multicolumn{5}{c}{Predicted class original/\textbf{SpuFix} (class prob. flagpole original/\textbf{SpuFix})}\\
        \hline
         flagpole/&
         flagpole/&
         flagpole/&
         flagpole/&
         flagpole/ \\
         \textbf{Windsor tie}&
         \textbf{parachute}&
         \textbf{parachute}&
         \textbf{parachute}&
         \textbf{parachute} \\
         
         \includegraphics[width=0.1\textwidth]{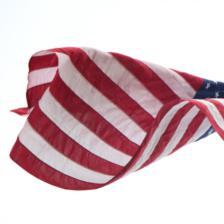}& 
         \includegraphics[width=0.1\textwidth]{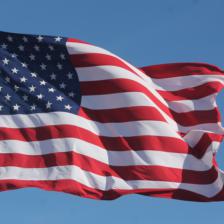}&
         \includegraphics[width=0.1\textwidth]{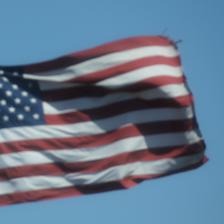}& 
         \includegraphics[width=0.1\textwidth]{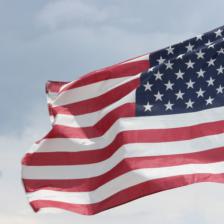}& 
         \includegraphics[width=0.1\textwidth]{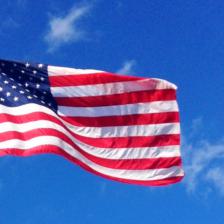} \\

         0.97/\textbf{0.01}&
         0.99/\textbf{0.02}&
         1.00/\textbf{0.12}&
         1.00/\textbf{0.04}&
         1.00/\textbf{0.01} \\

         \hline

         park bench/&
         flagpole/&
         flagpole/&
         Windsor tie/&
         Christmas st./\\

         \textbf{park bench}&
         \textbf{bow tie}&
         \textbf{parachute}&
         \textbf{Windsor tie}&
         \textbf{Christmas st.}\\
         
         \includegraphics[width=0.1\textwidth]{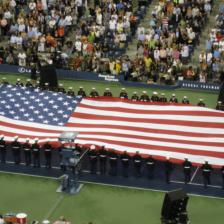}& 
         \includegraphics[width=0.1\textwidth]{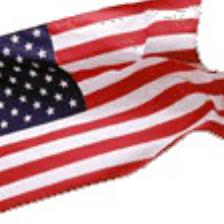}&
         \includegraphics[width=0.1\textwidth]{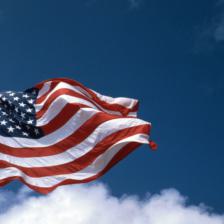}& 
         \includegraphics[width=0.1\textwidth]{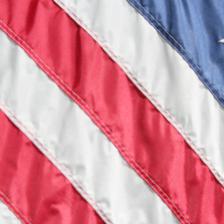}& 
         \includegraphics[width=0.1\textwidth]{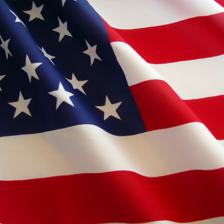} \\

         0.03/\textbf{0.03}&
         0.81/\textbf{0.00}&
         1.00/\textbf{0.17}&
         0.01/\textbf{0.01}&
         0.14/\textbf{0.00}
    \end{tabular}

    \caption{\textbf{Validation of SpuFix independent of Spurious ImageNet:} We collected 10 images each for the classes hummingbird/gondola/flagpole containing \textbf{only} the spurious feature (bird feeder/building/US flag) without filtering by model predictions or $\alpha_l^{(k)}$. Our robust ResNet50 classifies 6/8/7 as the corresponding class (mean class probability 0.57/0.61/0.70), the SpuFix version only 3/4/0 (mean class probability 0.13/0.13/0.05). Therefore, SpuFix reduces the reliance on these harmful spurious features independent of the image collection procedure.}
    \label{fig:testset}
\end{figure*}

%% file: appendix_bar_plots.tex
\begin{figure*}

\centering
\includegraphics[width=\textwidth]{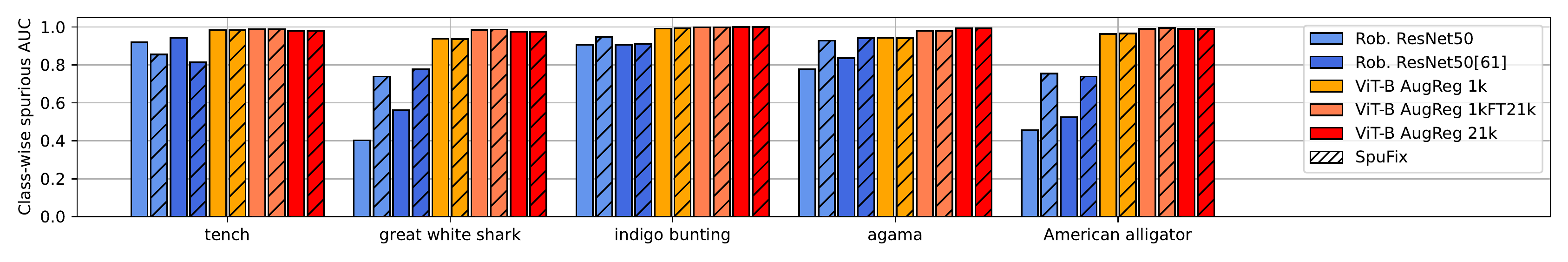}
\includegraphics[width=\textwidth]{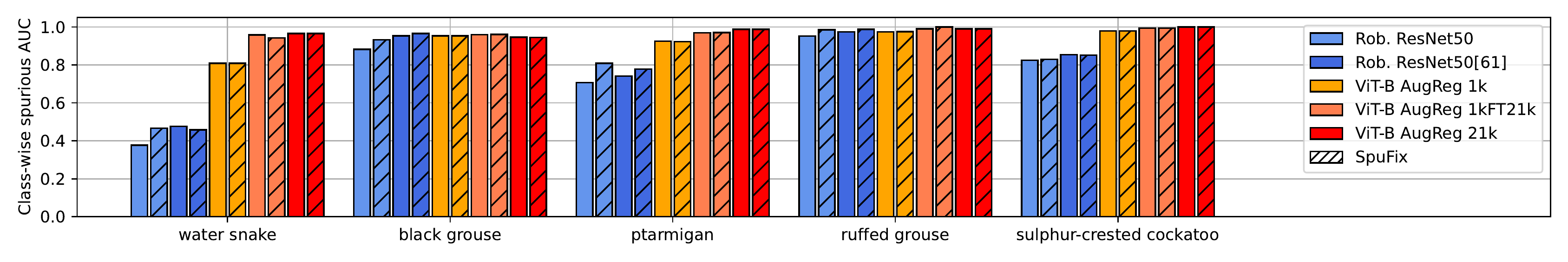}
\includegraphics[width=\textwidth]{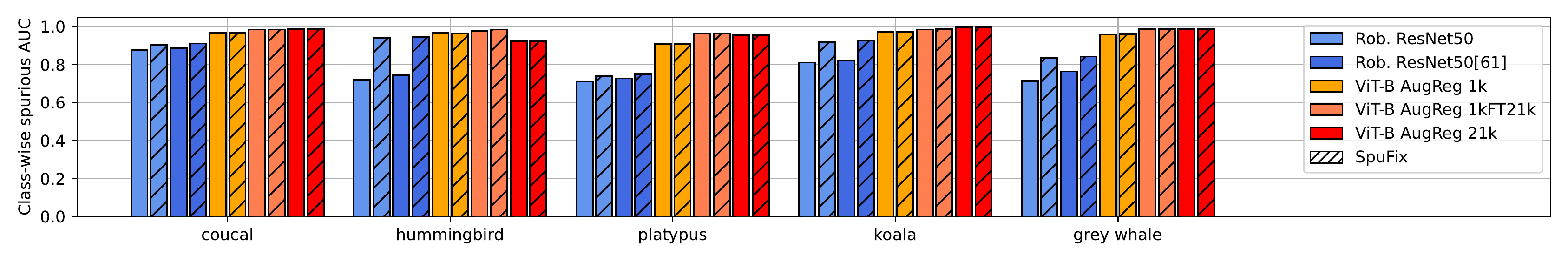}
\includegraphics[width=\textwidth]{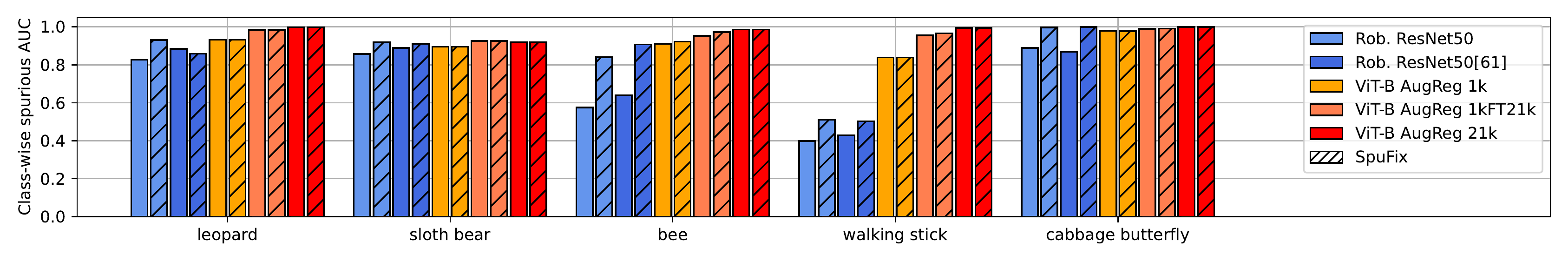}
\includegraphics[width=\textwidth]{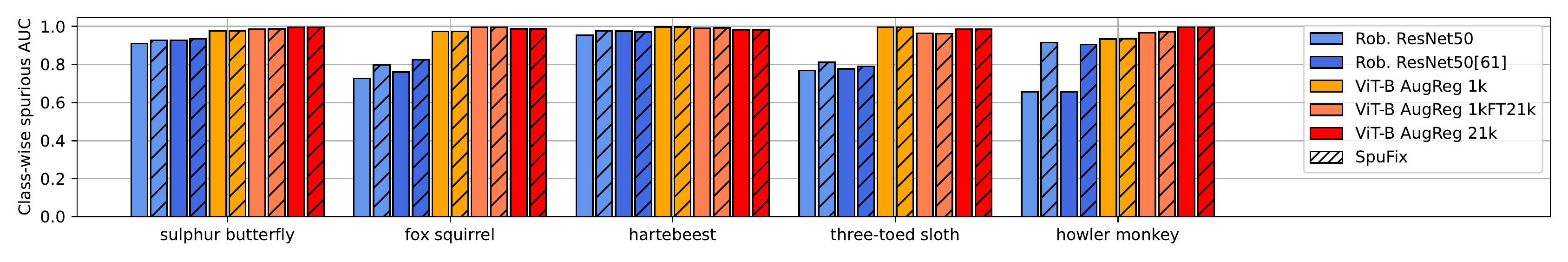}
\includegraphics[width=\textwidth]{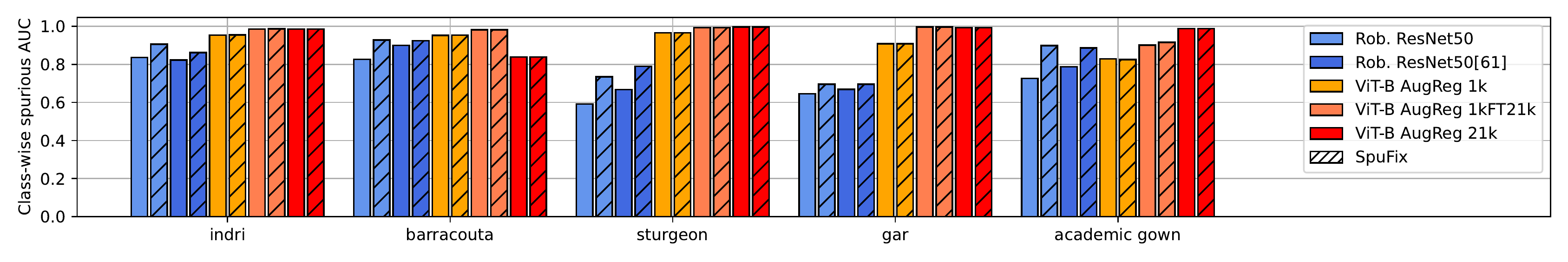}
\includegraphics[width=\textwidth]{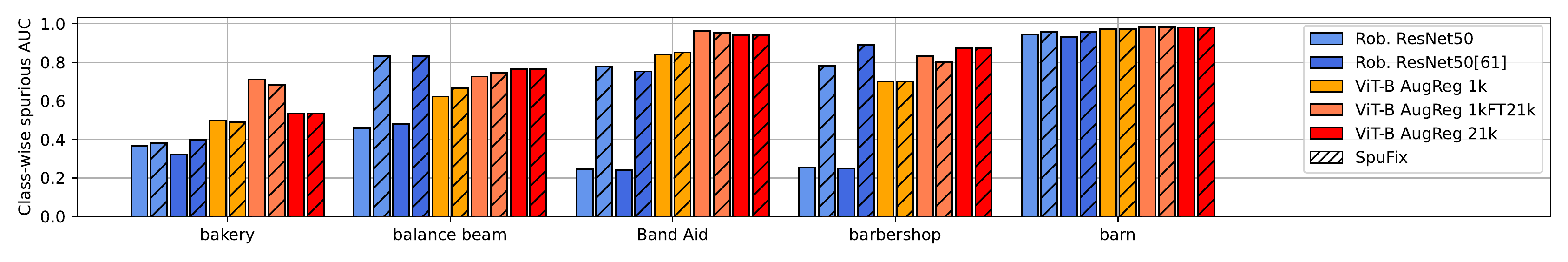}
\caption{\label{fig:app_bar1}Extended version of Fig.~\ref{fig:barplot} from the main paper for the first 35 classes in our dataset. We plot class-wise spurious AUC for our robust ResNet50, the robust ResNet50 from \cite{Singla2022salient}, and a ViT-B, both trained on ImageNet1k with and without pre-training on ImageNet21k as well as pure ImageNet21k training. Additionally, we show the corresponding SpuFix versions of the five models.}
\end{figure*}

\begin{figure*}
\centering
\includegraphics[width=\textwidth]{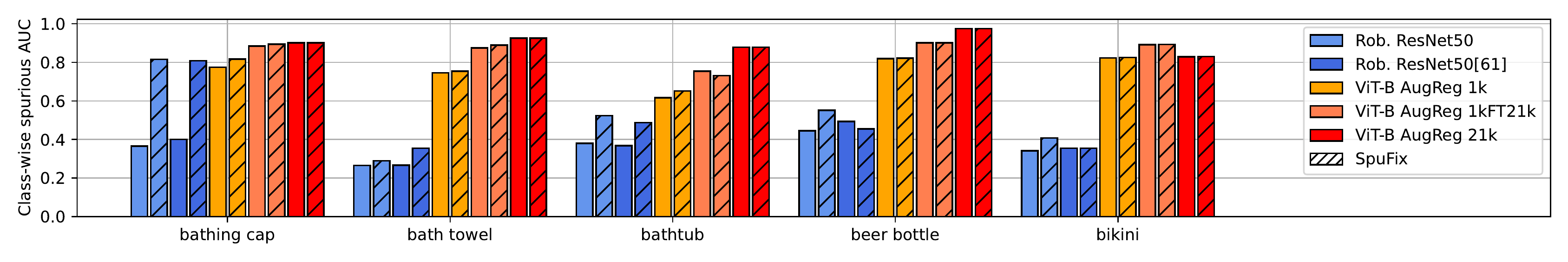}
\includegraphics[width=\textwidth]{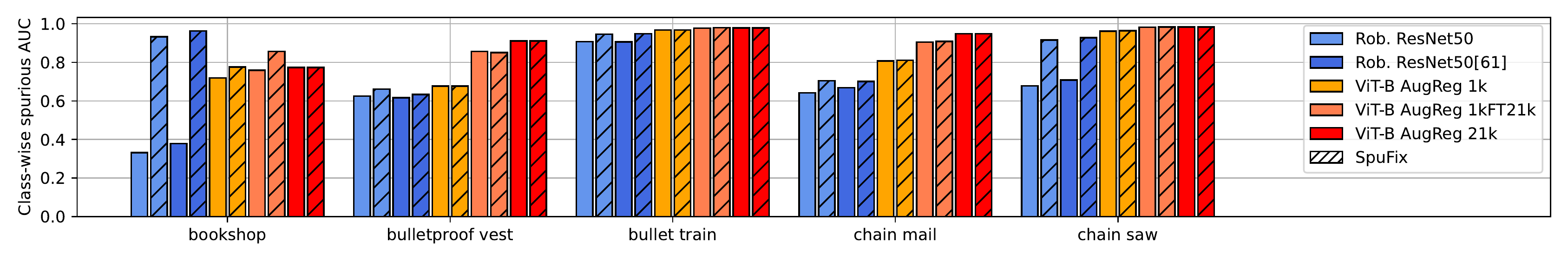}
\includegraphics[width=\textwidth]{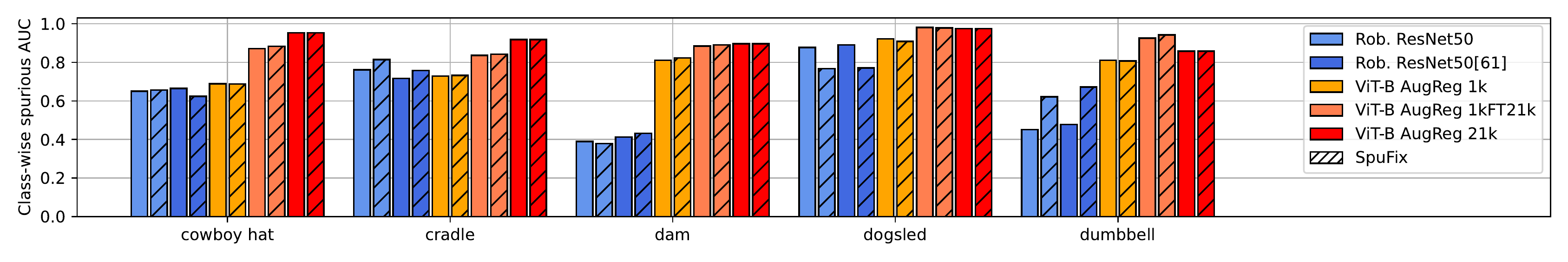}
\includegraphics[width=\textwidth]{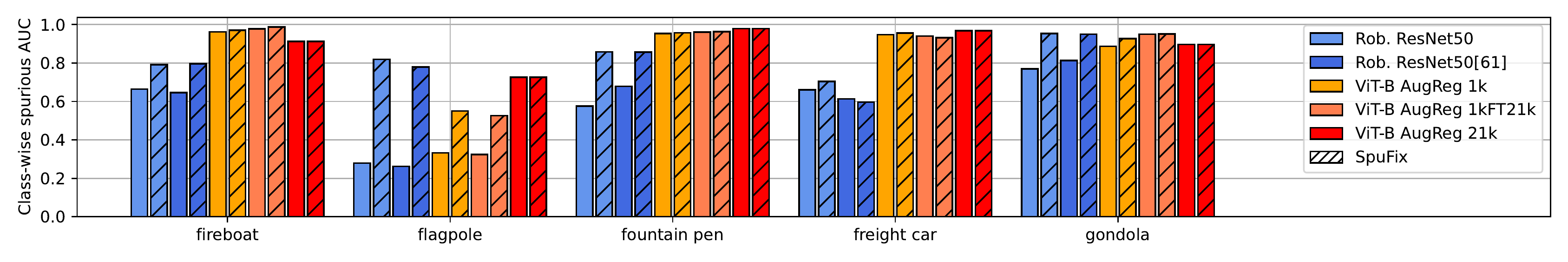}
\includegraphics[width=\textwidth]{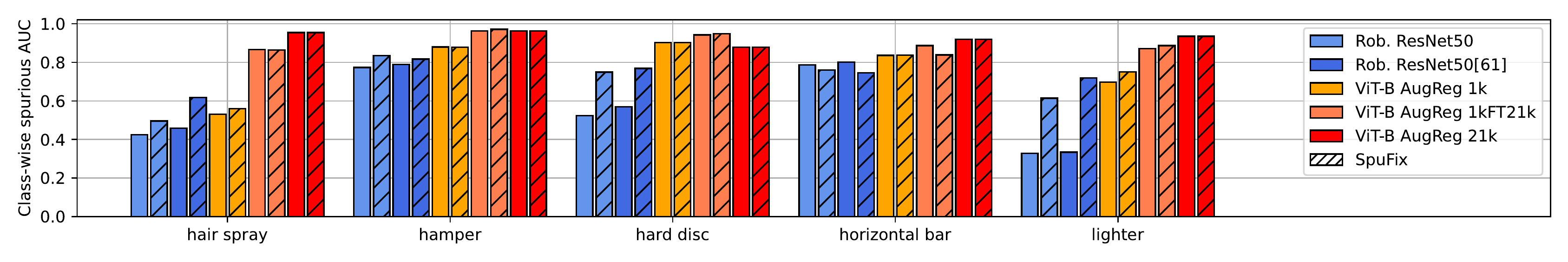}
\includegraphics[width=\textwidth]{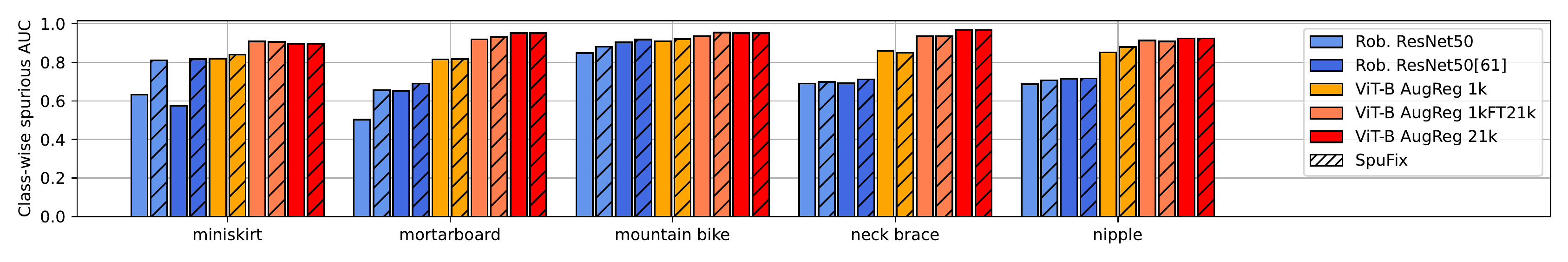}
\includegraphics[width=\textwidth]{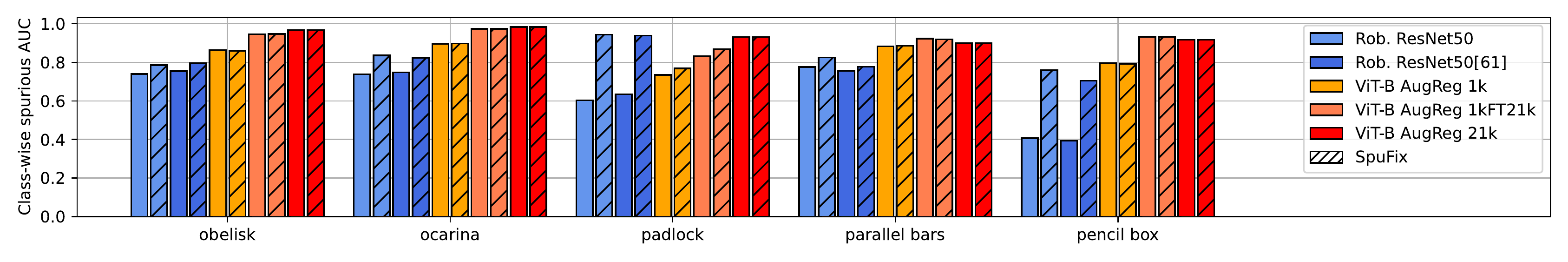}
\caption{\label{fig:app_bar2} Continued from Fig.~\ref{fig:app_bar1} for classes 36-70.}
\end{figure*}

\begin{figure*}
\centering
\includegraphics[width=\textwidth]{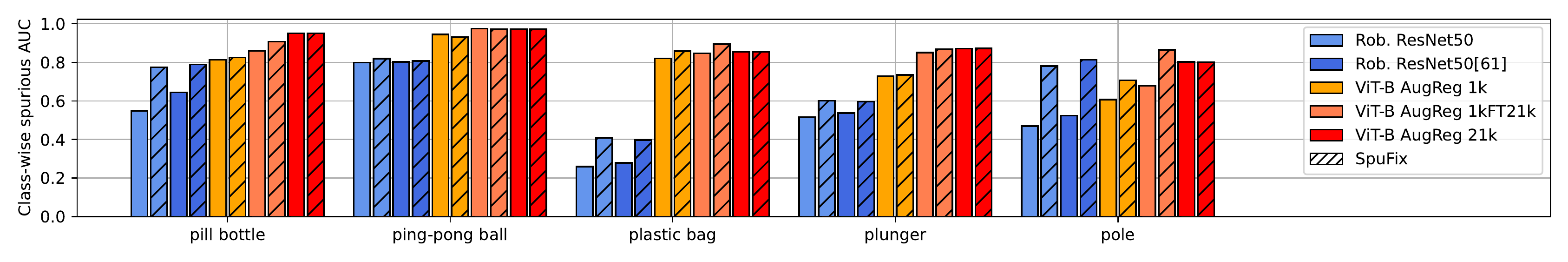}
\includegraphics[width=\textwidth]{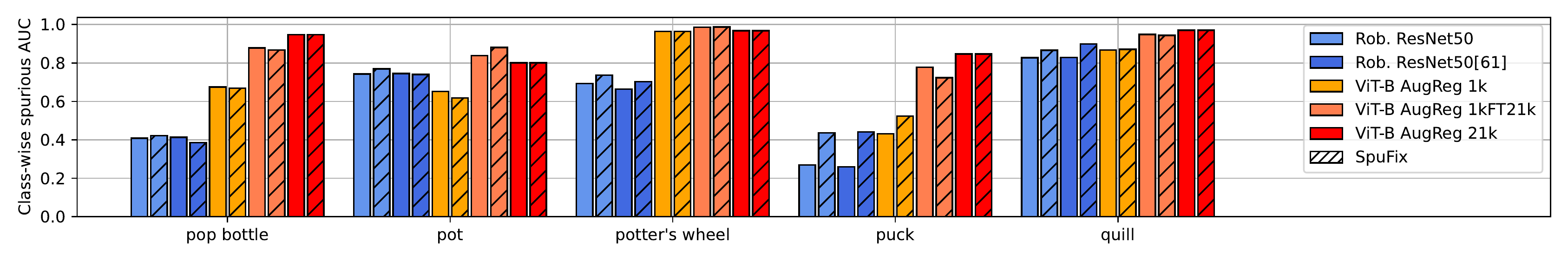}
\includegraphics[width=\textwidth]{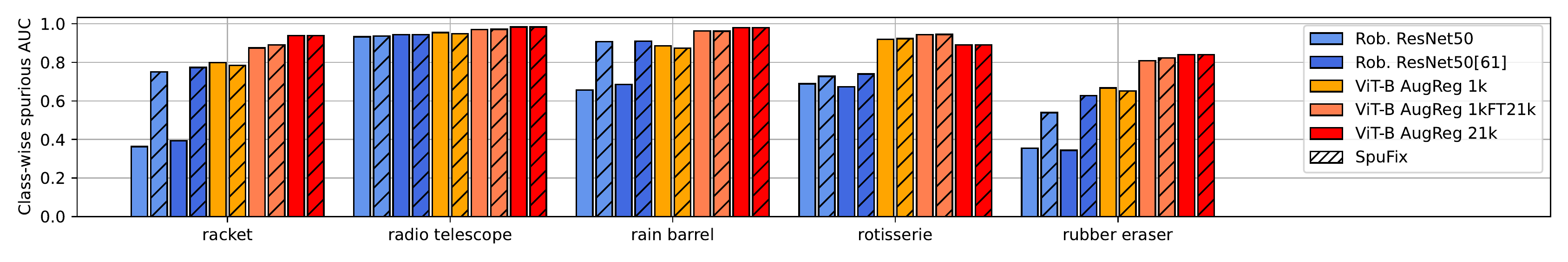}
\includegraphics[width=\textwidth]{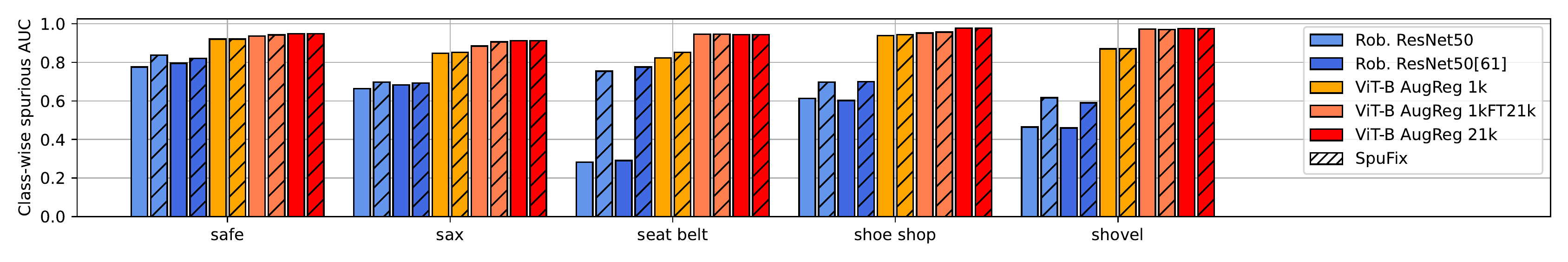}
\includegraphics[width=\textwidth]{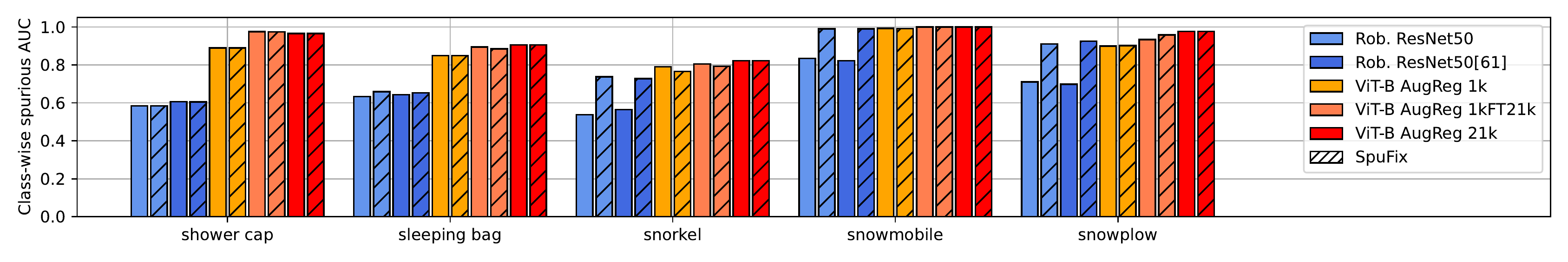}
\includegraphics[width=\textwidth]{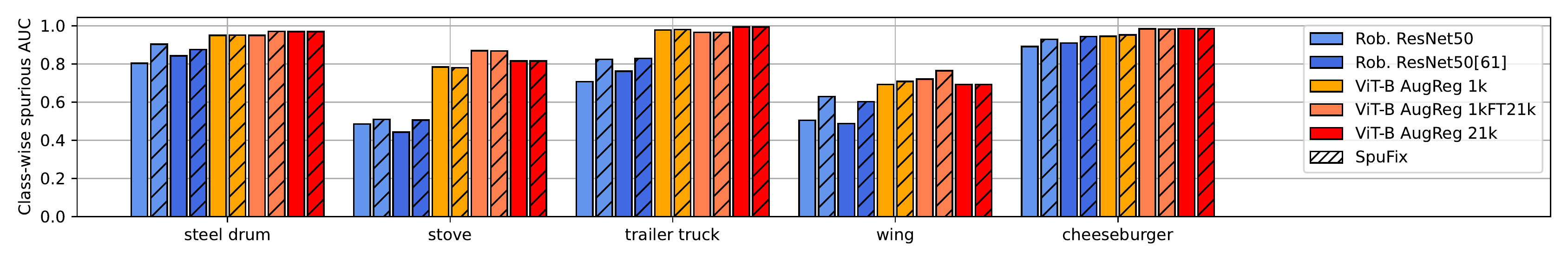}
\caption{\label{fig:app_bar3} Continued from Fig.~\ref{fig:app_bar1} for classes 71-100.}
\vspace{50mm}
\end{figure*}

%% file: app_749_quill.tex
\begin{table*}
\centering
\begin{tabular}{c|c|c|c|c|c|c|c|c|c}
	\begin{subfigure}{0.08\textwidth}\centering\caption*{\tiny\textcolor{red}{\makecell{quill: 0.97\\book jacket: 0.01\\fountain pen: 0.00}}}\includegraphics[width=1\textwidth]{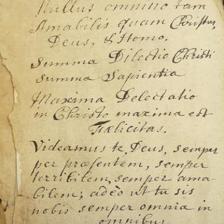}\end{subfigure} & 
	\begin{subfigure}{0.08\textwidth}\centering\caption*{\tiny\textcolor{red}{\makecell{quill: 0.92\\envelope: 0.04\\fountain pen: 0.01}}}\includegraphics[width=1\textwidth]{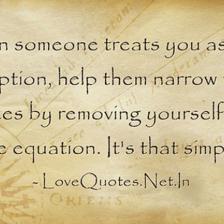}\end{subfigure} & 
	\begin{subfigure}{0.08\textwidth}\centering\caption*{\tiny\textcolor{red}{\makecell{quill: 0.91\\book jacket: 0.02\\fountain pen: 0.01}}}\includegraphics[width=1\textwidth]{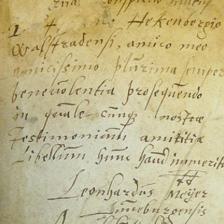}\end{subfigure} & 
	\begin{subfigure}{0.08\textwidth}\centering\caption*{\tiny\textcolor{red}{\makecell{quill: 0.89\\fountain pen: 0.01\\packet: 0.01}}}\includegraphics[width=1\textwidth]{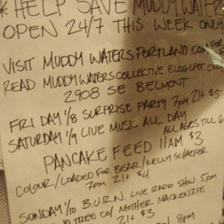}\end{subfigure} & 
	\begin{subfigure}{0.08\textwidth}\centering\caption*{\tiny\textcolor{red}{\makecell{quill: 0.88\\fountain pen: 0.08\\envelope: 0.01}}}\includegraphics[width=1\textwidth]{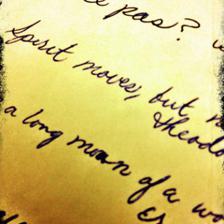}\end{subfigure} & 
	\begin{subfigure}{0.08\textwidth}\centering\caption*{\tiny\textcolor{red}{\makecell{quill: 0.85\\envelope: 0.04\\fountain pen: 0.03}}}\includegraphics[width=1\textwidth]{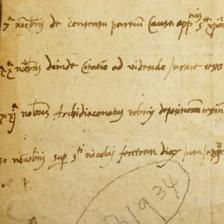}\end{subfigure} & 
	\begin{subfigure}{0.08\textwidth}\centering\caption*{\tiny\textcolor{red}{\makecell{quill: 0.83\\fountain pen: 0.05\\book jacket: 0.01}}}\includegraphics[width=1\textwidth]{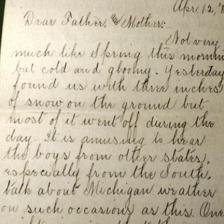}\end{subfigure} & 
	\begin{subfigure}{0.08\textwidth}\centering\caption*{\tiny\textcolor{red}{\makecell{quill: 0.81\\envelope: 0.08\\fountain pen: 0.01}}}\includegraphics[width=1\textwidth]{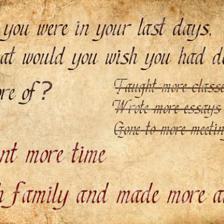}\end{subfigure} & 
	\begin{subfigure}{0.08\textwidth}\centering\caption*{\tiny\textcolor{red}{\makecell{quill: 0.79\\book jacket: 0.07\\fountain pen: 0.03}}}\includegraphics[width=1\textwidth]{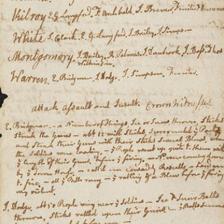}\end{subfigure} & 
	\begin{subfigure}{0.08\textwidth}\centering\caption*{\tiny\textcolor{red}{\makecell{quill: 0.79\\envelope: 0.07\\fountain pen: 0.03}}}\includegraphics[width=1\textwidth]{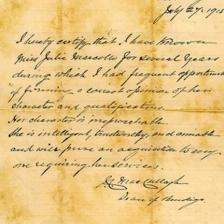}\end{subfigure}\\
	\begin{subfigure}{0.08\textwidth}\centering\caption*{\tiny\textcolor{red}{\makecell{quill: 0.78\\book jacket: 0.07\\fountain pen: 0.03}}}\includegraphics[width=1\textwidth]{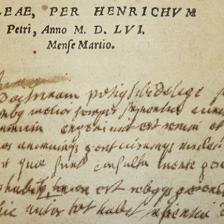}\end{subfigure} & 
	\begin{subfigure}{0.08\textwidth}\centering\caption*{\tiny\textcolor{red}{\makecell{quill: 0.78\\book jacket: 0.04\\envelope: 0.03}}}\includegraphics[width=1\textwidth]{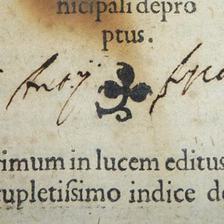}\end{subfigure} & 
	\begin{subfigure}{0.08\textwidth}\centering\caption*{\tiny\textcolor{red}{\makecell{quill: 0.78\\envelope: 0.05\\book jacket: 0.03}}}\includegraphics[width=1\textwidth]{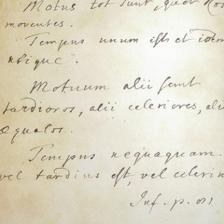}\end{subfigure} & 
	\begin{subfigure}{0.08\textwidth}\centering\caption*{\tiny\textcolor{red}{\makecell{quill: 0.77\\book jacket: 0.08\\envelope: 0.05}}}\includegraphics[width=1\textwidth]{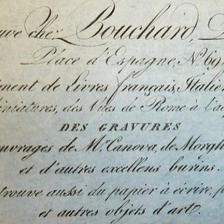}\end{subfigure} & 
	\begin{subfigure}{0.08\textwidth}\centering\caption*{\tiny\textcolor{red}{\makecell{quill: 0.72\\envelope: 0.09\\menu: 0.02}}}\includegraphics[width=1\textwidth]{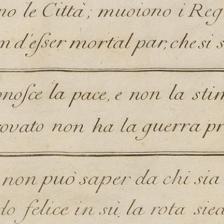}\end{subfigure} & 
	\begin{subfigure}{0.08\textwidth}\centering\caption*{\tiny\textcolor{red}{\makecell{quill: 0.71\\book jacket: 0.04\\prayer rug: 0.02}}}\includegraphics[width=1\textwidth]{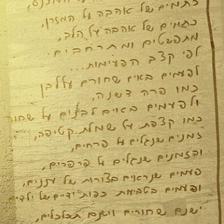}\end{subfigure} & 
	\begin{subfigure}{0.08\textwidth}\centering\caption*{\tiny\textcolor{red}{\makecell{quill: 0.69\\fountain pen: 0.06\\book jacket: 0.02}}}\includegraphics[width=1\textwidth]{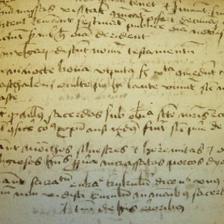}\end{subfigure} & 
	\begin{subfigure}{0.08\textwidth}\centering\caption*{\tiny\textcolor{red}{\makecell{quill: 0.67\\book jacket: 0.09\\envelope: 0.02}}}\includegraphics[width=1\textwidth]{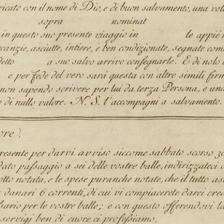}\end{subfigure} & 
	\begin{subfigure}{0.08\textwidth}\centering\caption*{\tiny\textcolor{red}{\makecell{quill: 0.65\\fountain pen: 0.10\\book jacket: 0.06}}}\includegraphics[width=1\textwidth]{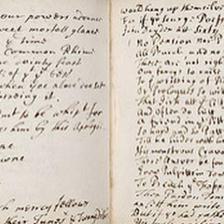}\end{subfigure} & 
	\begin{subfigure}{0.08\textwidth}\centering\caption*{\tiny\textcolor{red}{\makecell{quill: 0.65\\book jacket: 0.13\\fountain pen: 0.04}}}\includegraphics[width=1\textwidth]{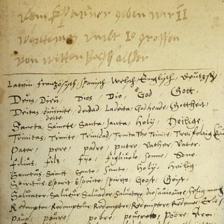}\end{subfigure}\\
	\begin{subfigure}{0.08\textwidth}\centering\caption*{\tiny\textcolor{red}{\makecell{quill: 0.64\\envelope: 0.08\\book jacket: 0.04}}}\includegraphics[width=1\textwidth]{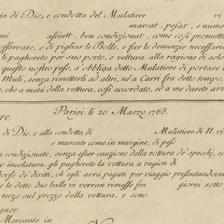}\end{subfigure} & 
	\begin{subfigure}{0.08\textwidth}\centering\caption*{\tiny\textcolor{red}{\makecell{quill: 0.64\\book jacket: 0.06\\fountain pen: 0.03}}}\includegraphics[width=1\textwidth]{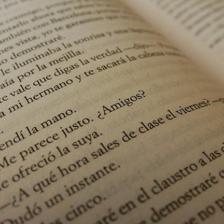}\end{subfigure} & 
	\begin{subfigure}{0.08\textwidth}\centering\caption*{\tiny\textcolor{red}{\makecell{quill: 0.61\\envelope: 0.04\\fountain pen: 0.03}}}\includegraphics[width=1\textwidth]{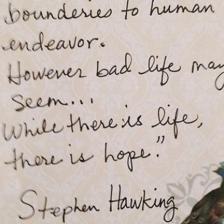}\end{subfigure} & 
	\begin{subfigure}{0.08\textwidth}\centering\caption*{\tiny\textcolor{red}{\makecell{quill: 0.61\\book jacket: 0.08\\fountain pen: 0.08}}}\includegraphics[width=1\textwidth]{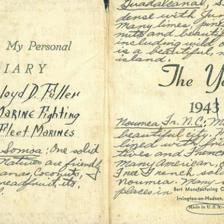}\end{subfigure} & 
	\begin{subfigure}{0.08\textwidth}\centering\caption*{\tiny\textcolor{red}{\makecell{quill: 0.59\\fountain pen: 0.23\\book jacket: 0.02}}}\includegraphics[width=1\textwidth]{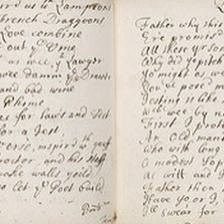}\end{subfigure} & 
	\begin{subfigure}{0.08\textwidth}\centering\caption*{\tiny\textcolor{red}{\makecell{quill: 0.56\\fountain pen: 0.09\\binder: 0.07}}}\includegraphics[width=1\textwidth]{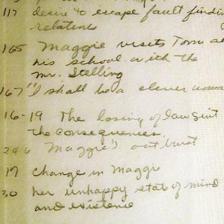}\end{subfigure} & 
	\begin{subfigure}{0.08\textwidth}\centering\caption*{\tiny\textcolor{red}{\makecell{quill: 0.55\\envelope: 0.21\\book jacket: 0.04}}}\includegraphics[width=1\textwidth]{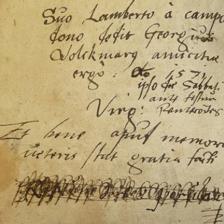}\end{subfigure} & 
	\begin{subfigure}{0.08\textwidth}\centering\caption*{\tiny\textcolor{red}{\makecell{quill: 0.54\\fountain pen: 0.15\\binder: 0.02}}}\includegraphics[width=1\textwidth]{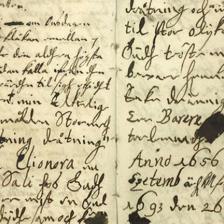}\end{subfigure} & 
	\begin{subfigure}{0.08\textwidth}\centering\caption*{\tiny\textcolor{red}{\makecell{quill: 0.52\\fountain pen: 0.25\\binder: 0.04}}}\includegraphics[width=1\textwidth]{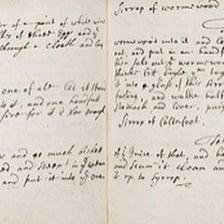}\end{subfigure} & 
	\begin{subfigure}{0.08\textwidth}\centering\caption*{\tiny\textcolor{red}{\makecell{quill: 0.51\\fountain pen: 0.17\\book jacket: 0.08}}}\includegraphics[width=1\textwidth]{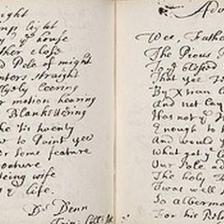}\end{subfigure}\\
	\begin{subfigure}{0.08\textwidth}\centering\caption*{\tiny\textcolor{red}{\makecell{quill: 0.50\\book jacket: 0.15\\fountain pen: 0.12}}}\includegraphics[width=1\textwidth]{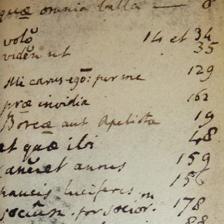}\end{subfigure} & 
	\begin{subfigure}{0.08\textwidth}\centering\caption*{\tiny\textcolor{red}{\makecell{quill: 0.48\\fountain pen: 0.31\\binder: 0.03}}}\includegraphics[width=1\textwidth]{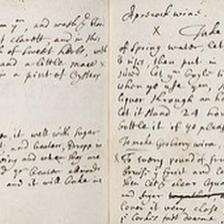}\end{subfigure} & 
	\begin{subfigure}{0.08\textwidth}\centering\caption*{\tiny\textcolor{red}{\makecell{quill: 0.48\\book jacket: 0.14\\fountain pen: 0.10}}}\includegraphics[width=1\textwidth]{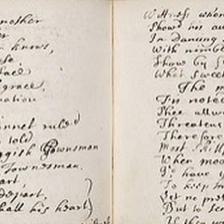}\end{subfigure} & 
	\begin{subfigure}{0.08\textwidth}\centering\caption*{\tiny\textcolor{red}{\makecell{quill: 0.47\\envelope: 0.27\\book jacket: 0.08}}}\includegraphics[width=1\textwidth]{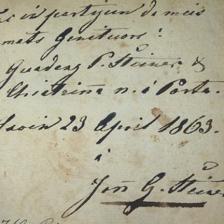}\end{subfigure} & 
	\begin{subfigure}{0.08\textwidth}\centering\caption*{\tiny\textcolor{red}{\makecell{quill: 0.47\\fountain pen: 0.10\\book jacket: 0.08}}}\includegraphics[width=1\textwidth]{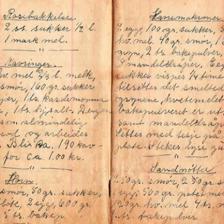}\end{subfigure} & 
	\begin{subfigure}{0.08\textwidth}\centering\caption*{\tiny\textcolor{red}{\makecell{quill: 0.44\\book jacket: 0.37\\fountain pen: 0.01}}}\includegraphics[width=1\textwidth]{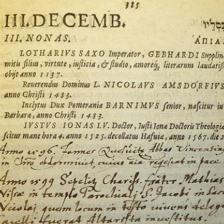}\end{subfigure} & 
	\begin{subfigure}{0.08\textwidth}\centering\caption*{\tiny\textcolor{red}{\makecell{quill: 0.44\\tray: 0.34\\book jacket: 0.03}}}\includegraphics[width=1\textwidth]{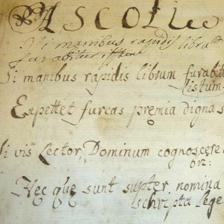}\end{subfigure} & 
	\begin{subfigure}{0.08\textwidth}\centering\caption*{\tiny\textcolor{red}{\makecell{quill: 0.44\\fountain pen: 0.24\\book jacket: 0.05}}}\includegraphics[width=1\textwidth]{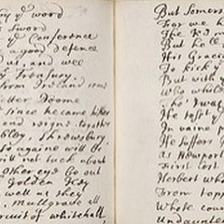}\end{subfigure} & 
	\begin{subfigure}{0.08\textwidth}\centering\caption*{\tiny\textcolor{red}{\makecell{quill: 0.44\\envelope: 0.12\\fountain pen: 0.10}}}\includegraphics[width=1\textwidth]{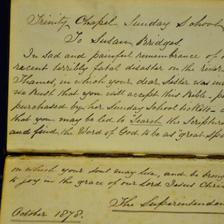}\end{subfigure} & 
	\begin{subfigure}{0.08\textwidth}\centering\caption*{\tiny\textcolor{red}{\makecell{quill: 0.43\\book jacket: 0.26\\fountain pen: 0.04}}}\includegraphics[width=1\textwidth]{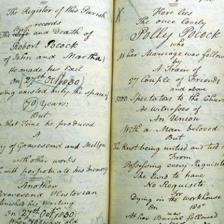}\end{subfigure}\\
	\begin{subfigure}{0.08\textwidth}\centering\caption*{\tiny\textcolor{red}{\makecell{quill: 0.42\\envelope: 0.09\\book jacket: 0.06}}}\includegraphics[width=1\textwidth]{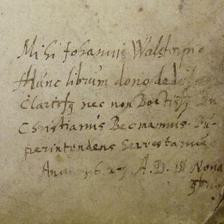}\end{subfigure} & 
	\begin{subfigure}{0.08\textwidth}\centering\caption*{\tiny\textcolor{red}{\makecell{quill: 0.42\\book jacket: 0.32\\fountain pen: 0.04}}}\includegraphics[width=1\textwidth]{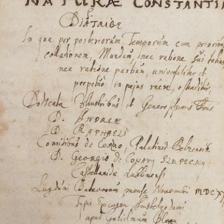}\end{subfigure} & 
	\begin{subfigure}{0.08\textwidth}\centering\caption*{\tiny\textcolor{red}{\makecell{quill: 0.42\\fountain pen: 0.09\\candle: 0.08}}}\includegraphics[width=1\textwidth]{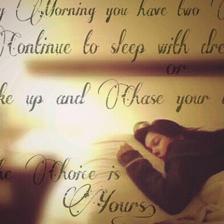}\end{subfigure} & 
	\begin{subfigure}{0.08\textwidth}\centering\caption*{\tiny\textcolor{red}{\makecell{quill: 0.41\\fountain pen: 0.07\\book jacket: 0.04}}}\includegraphics[width=1\textwidth]{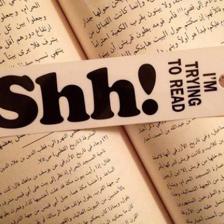}\end{subfigure} & 
	\begin{subfigure}{0.08\textwidth}\centering\caption*{\tiny\textcolor{red}{\makecell{quill: 0.39\\envelope: 0.16\\slide rule: 0.07}}}\includegraphics[width=1\textwidth]{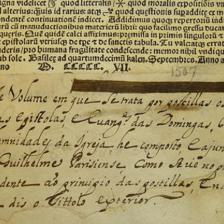}\end{subfigure} & 
	\begin{subfigure}{0.08\textwidth}\centering\caption*{\tiny\textcolor{red}{\makecell{quill: 0.39\\book jacket: 0.36\\fountain pen: 0.02}}}\includegraphics[width=1\textwidth]{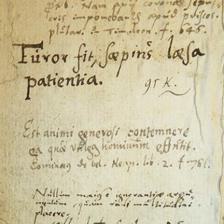}\end{subfigure} & 
	\begin{subfigure}{0.08\textwidth}\centering\caption*{\tiny\textcolor{red}{\makecell{quill: 0.38\\fountain pen: 0.16\\binder: 0.08}}}\includegraphics[width=1\textwidth]{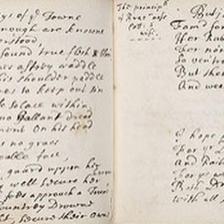}\end{subfigure} & 
	\begin{subfigure}{0.08\textwidth}\centering\caption*{\tiny\textcolor{red}{\makecell{quill: 0.37\\fountain pen: 0.21\\binder: 0.04}}}\includegraphics[width=1\textwidth]{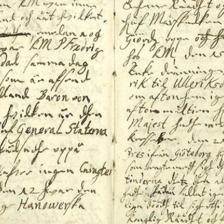}\end{subfigure} & 
	\begin{subfigure}{0.08\textwidth}\centering\caption*{\tiny\textcolor{red}{\makecell{quill: 0.37\\fountain pen: 0.11\\book jacket: 0.09}}}\includegraphics[width=1\textwidth]{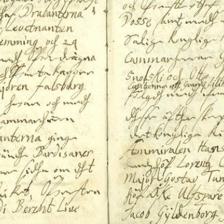}\end{subfigure} & 
	\begin{subfigure}{0.08\textwidth}\centering\caption*{\tiny\textcolor{red}{\makecell{quill: 0.36\\book jacket: 0.20\\fountain pen: 0.12}}}\includegraphics[width=1\textwidth]{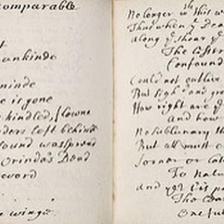}\end{subfigure}\\
	\begin{subfigure}{0.08\textwidth}\centering\caption*{\tiny\textcolor{red}{\makecell{quill: 0.34\\fountain pen: 0.20\\book jacket: 0.10}}}\includegraphics[width=1\textwidth]{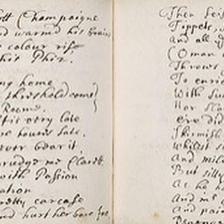}\end{subfigure} & 
	\begin{subfigure}{0.08\textwidth}\centering\caption*{\tiny\textcolor{red}{\makecell{quill: 0.33\\fountain pen: 0.31\\binder: 0.05}}}\includegraphics[width=1\textwidth]{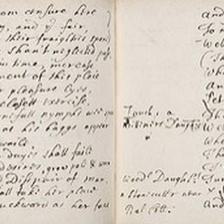}\end{subfigure} & 
	\begin{subfigure}{0.08\textwidth}\centering\caption*{\tiny\textcolor{red}{\makecell{quill: 0.31\\book jacket: 0.24\\fountain pen: 0.12}}}\includegraphics[width=1\textwidth]{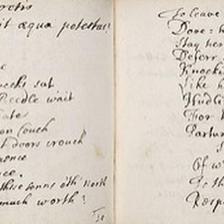}\end{subfigure} & 
	\begin{subfigure}{0.08\textwidth}\centering\caption*{\tiny\textcolor{red}{\makecell{quill: 0.31\\fountain pen: 0.16\\book jacket: 0.15}}}\includegraphics[width=1\textwidth]{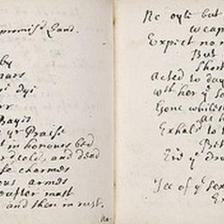}\end{subfigure} & 
	\begin{subfigure}{0.08\textwidth}\centering\caption*{\tiny\textcolor{red}{\makecell{quill: 0.30\\binder: 0.13\\fountain pen: 0.11}}}\includegraphics[width=1\textwidth]{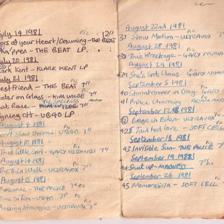}\end{subfigure} & 
	\begin{subfigure}{0.08\textwidth}\centering\caption*{\tiny\textcolor{red}{\makecell{quill: 0.30\\book jacket: 0.29\\fountain pen: 0.05}}}\includegraphics[width=1\textwidth]{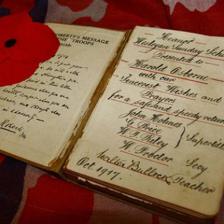}\end{subfigure} & 
	\begin{subfigure}{0.08\textwidth}\centering\caption*{\tiny\textcolor{red}{\makecell{quill: 0.28\\fountain pen: 0.12\\book jacket: 0.11}}}\includegraphics[width=1\textwidth]{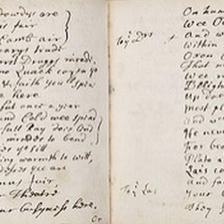}\end{subfigure} & 
	\begin{subfigure}{0.08\textwidth}\centering\caption*{\tiny\textcolor{red}{\makecell{quill: 0.24\\fountain pen: 0.17\\book jacket: 0.14}}}\includegraphics[width=1\textwidth]{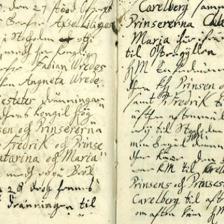}\end{subfigure} & 
	\begin{subfigure}{0.08\textwidth}\centering\caption*{\tiny\textcolor{red}{\makecell{quill: 0.24\\envelope: 0.13\\book jacket: 0.11}}}\includegraphics[width=1\textwidth]{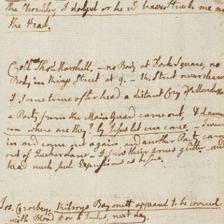}\end{subfigure} & 
	\begin{subfigure}{0.08\textwidth}\centering\caption*{\tiny\textcolor{red}{\makecell{quill: 0.23\\fountain pen: 0.19\\book jacket: 0.12}}}\includegraphics[width=1\textwidth]{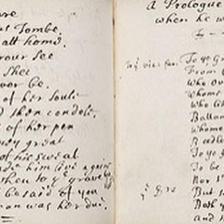}\end{subfigure}\\
	\begin{subfigure}{0.08\textwidth}\centering\caption*{\tiny\textcolor{red}{\makecell{quill: 0.21\\book jacket: 0.18\\fountain pen: 0.11}}}\includegraphics[width=1\textwidth]{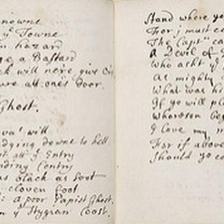}\end{subfigure} & 
	\begin{subfigure}{0.08\textwidth}\centering\caption*{\tiny\textcolor{red}{\makecell{quill: 0.15\\paper towel: 0.09\\menu: 0.05}}}\includegraphics[width=1\textwidth]{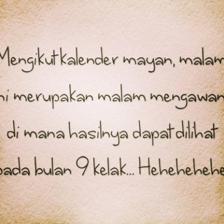}\end{subfigure} & 
	\begin{subfigure}{0.08\textwidth}\centering\caption*{\tiny\textcolor{black}{\makecell{book jacket: 0.85\\quill: 0.03\\menu: 0.01}}}\includegraphics[width=1\textwidth]{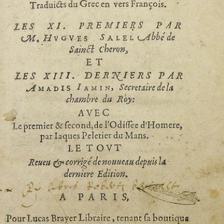}\end{subfigure} & 
	\begin{subfigure}{0.08\textwidth}\centering\caption*{\tiny\textcolor{black}{\makecell{envelope: 0.68\\quill: 0.16\\web site: 0.01}}}\includegraphics[width=1\textwidth]{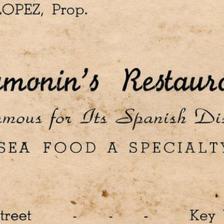}\end{subfigure} & 
	\begin{subfigure}{0.08\textwidth}\centering\caption*{\tiny\textcolor{black}{\makecell{book jacket: 0.66\\quill: 0.16\\binder: 0.01}}}\includegraphics[width=1\textwidth]{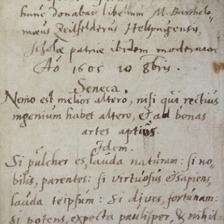}\end{subfigure} & 
	\begin{subfigure}{0.08\textwidth}\centering\caption*{\tiny\textcolor{black}{\makecell{book jacket: 0.59\\quill: 0.28\\menu: 0.01}}}\includegraphics[width=1\textwidth]{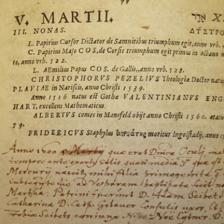}\end{subfigure} & 
	\begin{subfigure}{0.08\textwidth}\centering\caption*{\tiny\textcolor{black}{\makecell{book jacket: 0.58\\quill: 0.27\\fountain pen: 0.02}}}\includegraphics[width=1\textwidth]{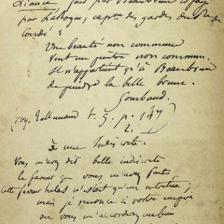}\end{subfigure} & 
	\begin{subfigure}{0.08\textwidth}\centering\caption*{\tiny\textcolor{black}{\makecell{book jacket: 0.50\\quill: 0.09\\fountain pen: 0.08}}}\includegraphics[width=1\textwidth]{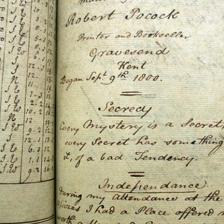}\end{subfigure} & 
	\begin{subfigure}{0.08\textwidth}\centering\caption*{\tiny\textcolor{black}{\makecell{walking stick: 0.49\\quill: 0.23\\fountain pen: 0.07}}}\includegraphics[width=1\textwidth]{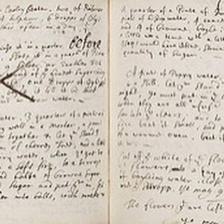}\end{subfigure} & 
	\begin{subfigure}{0.08\textwidth}\centering\caption*{\tiny\textcolor{black}{\makecell{fountain pen: 0.42\\quill: 0.33\\book jacket: 0.07}}}\includegraphics[width=1\textwidth]{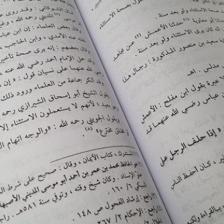}\end{subfigure}\\
	\begin{subfigure}{0.08\textwidth}\centering\caption*{\tiny\textcolor{black}{\makecell{book jacket: 0.37\\quill: 0.27\\whiskey jug: 0.05}}}\includegraphics[width=1\textwidth]{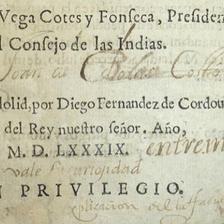}\end{subfigure} & 
	\begin{subfigure}{0.08\textwidth}\centering\caption*{\tiny\textcolor{black}{\makecell{book jacket: 0.36\\quill: 0.20\\fountain pen: 0.09}}}\includegraphics[width=1\textwidth]{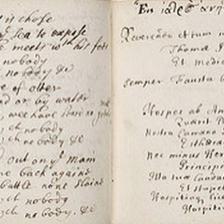}\end{subfigure} & 
	\begin{subfigure}{0.08\textwidth}\centering\caption*{\tiny\textcolor{black}{\makecell{book jacket: 0.34\\whiskey jug: 0.19\\quill: 0.06}}}\includegraphics[width=1\textwidth]{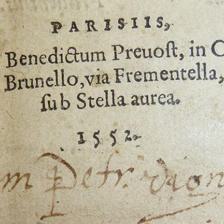}\end{subfigure} & 
	\begin{subfigure}{0.08\textwidth}\centering\caption*{\tiny\textcolor{black}{\makecell{book jacket: 0.33\\quill: 0.23\\fountain pen: 0.08}}}\includegraphics[width=1\textwidth]{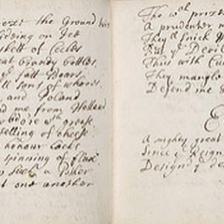}\end{subfigure} & 
	\begin{subfigure}{0.08\textwidth}\centering\caption*{\tiny\textcolor{black}{\makecell{book jacket: 0.30\\envelope: 0.13\\quill: 0.11}}}\includegraphics[width=1\textwidth]{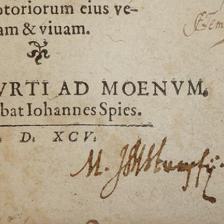}\end{subfigure}
\end{tabular}
    \captionof{figure}{\label{fig:quill}We show all images of class ``quill'' (spurious feature: handwritten text) in Spurious ImageNet together with the top-3 predicted probablities of the ConvNext-L-1kFT21k. Despite a class-wise AUC of 0.93 which seemingly suggests that the spurious feature is not playing a big role anymore, we observe that 76\% of the images are classified as ``quill'' despite no ``quill'' being present. Thus the spurious class extension is still present but the classifier produces slightly less confident predictions on these images.}
\end{table*}

%% file: worst12_compare_figure_1.tex
\begin{figure*}
    \setlength{\tabcolsep}{.01em}
    \centering
    \begin{tabular}{cc}
       \multicolumn{2}{c}{{\blue \textbf{Badger}}}\\
        \multicolumn{1}{c}{Top-5 max. act. images of \cite{Singla2022salient}(\textbf{28 identical pairs})} &\multicolumn{1}{c}{Top-5 max. act. images of NPCA (ours) (\textbf{0 identical pairs})}\\{\includegraphics[width=0.5\textwidth]{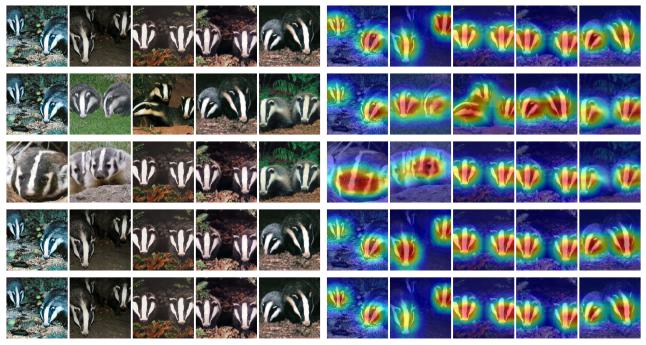}}
        &{\includegraphics[width=0.5\textwidth]{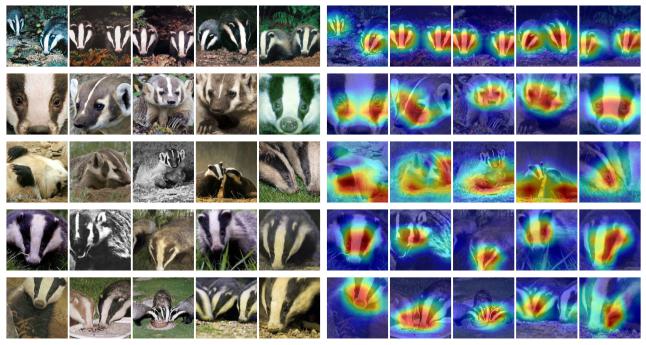}}
          \\ \hline
        \multicolumn{2}{c}{{\blue \textbf{King snake}}}\\
        \multicolumn{1}{c}{Top-5 max. act. images of \cite{Singla2022salient} (\textbf{24 identical pairs})} &\multicolumn{1}{c}{Top-5 max. act. images of NPCA (ours) (\textbf{1 identical pair})}\\
        {\includegraphics[width=0.5\textwidth]{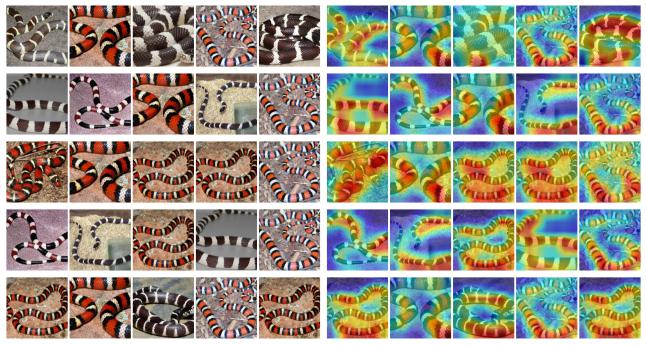}}
        &{\includegraphics[width=0.5\textwidth]{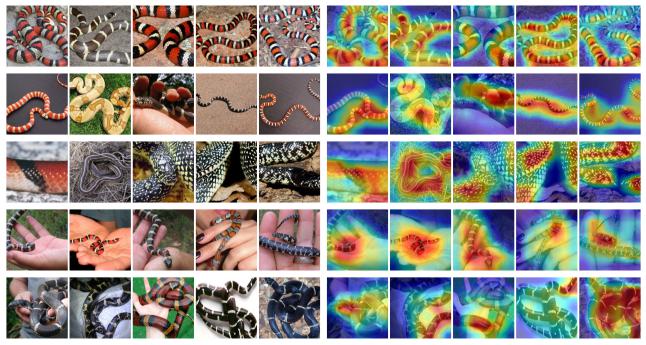}}
          \\ \hline
          \multicolumn{2}{c}{{\blue \textbf{Groom}}}\\
       \multicolumn{1}{c}{Top-5 max. act. images of \cite{Singla2022salient} (\textbf{24 identical pairs})} &\multicolumn{1}{c}{Top-5 max. act. images of NPCA (ours) (\textbf{1 identical pair})}\\
        {\includegraphics[width=0.5\textwidth]{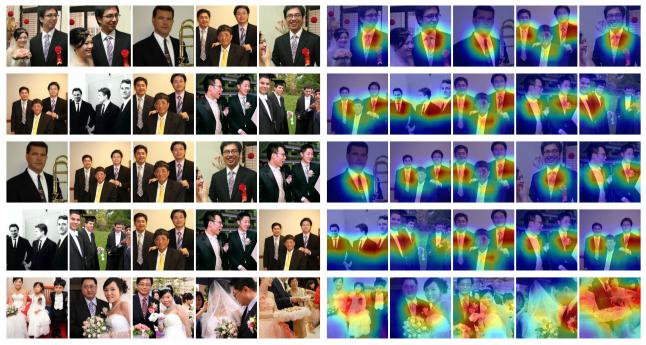}}
        &{\includegraphics[width=0.5\textwidth]{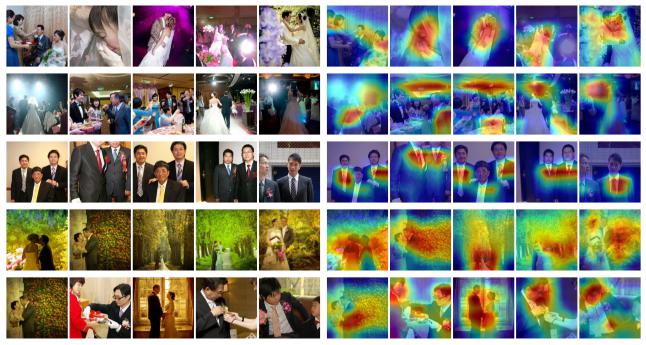}} 
        
    \end{tabular}
    \caption{\label{fig:worst12_compare_figure_1}\textbf{%
    Classes, where \cite{Singla2022salient} has the most identical pairs among the top-5 maximally activating images across different neurons (NPCA and neurons computed on the same model, Rob. ResNet50\cite{Singla2022salient}).} For each method we provide the number of identical pairs of images across different neurons for \cite{Singla2022salient} resp. different components for NPCA as described in App.~\ref{app:comparison}. Each row shows the 5 maximally activating images together with the corresponding GradCAM heatmaps for the top-5 neurons of \cite{Singla2022salient} (left) and the top-5 NPCA components (right), respectively. As in Fig.~\ref{fig:comparison}, our NPCA components are capturing different subpopulations in the training data for these classes, while the different neurons of \cite{Singla2022salient} are finding many identical pairs, see App.~\ref{app:comparison} for more details. \newline
    \textbf{Note:} for these components, where \cite{Singla2022salient} fails to find different subpopulations, our NPCA components find more diverse and even several spurious features: ``hands'' for the class ``king snake'' and  ``bride'', ``ceiling/lamps'', and ``trees/bushes'' for the class ``groom''. The top-5 neurons of \cite{Singla2022salient} for class ``groom'' identify three of the first four neurons as spurious and one as core even though the images are semantically the same and GradCAM activations are also similar. Semantically similar neurons are labeled differently due to their majority vote (for three of the neurons we have a 3:2 decision among the human labelers, for one a 4:1 decision), whereas we require that the two human labelers need to agree. %
    }
\end{figure*}

%% file: worst12_compare_figure_2.tex
\begin{figure*}
    \setlength{\tabcolsep}{.01em}
    \centering
    \begin{tabular}{cc}
       \multicolumn{2}{c}{{\blue \textbf{Lumbermill}}}\\
        \multicolumn{1}{c}{Top-5 max. act. images of \cite{Singla2022salient} (\textbf{0 identical pairs})} &\multicolumn{1}{c}{Top-5 max. act. images of NPCA (ours) (\textbf{10 identical pairs})}\\{\includegraphics[width=0.5\textwidth]{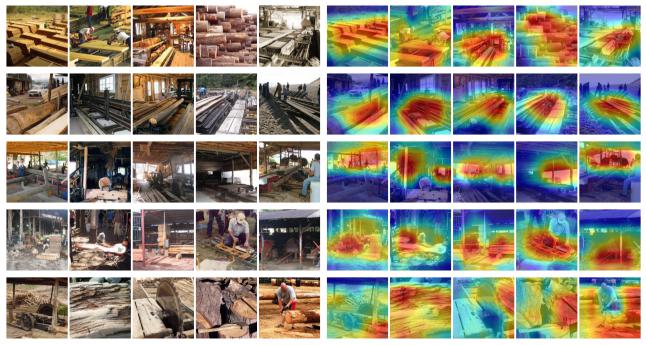}}
        &{\includegraphics[width=0.5\textwidth]{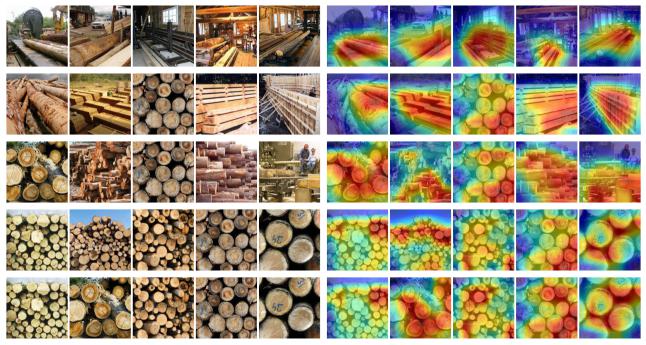}}
          \\ \hline
        \multicolumn{2}{c}{{\blue \textbf{Barbershop}}}\\
        \multicolumn{1}{c}{Top-5 max. act. images of \cite{Singla2022salient} (\textbf{6 identical pairs})} &\multicolumn{1}{c}{Top-5 max. act. images of NPCA (ours) (\textbf{8 identical pairs})}\\
        {\includegraphics[width=0.5\textwidth]{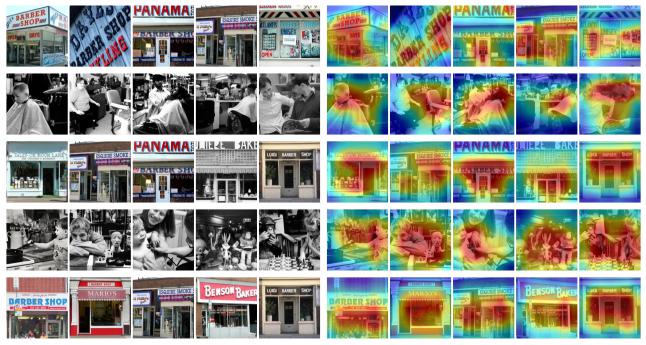}}
        &{\includegraphics[width=0.5\textwidth]{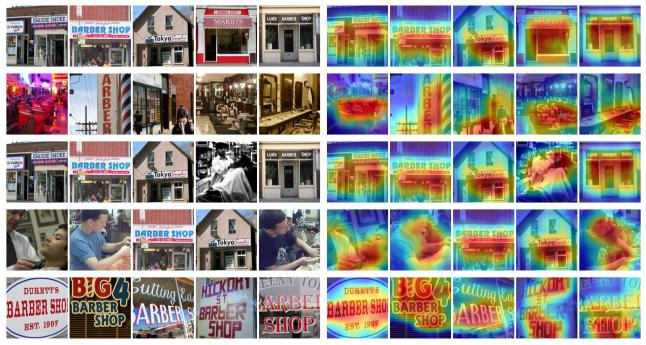}}
          \\ \hline
          \multicolumn{2}{c}{{\blue \textbf{English foxhound}}}\\
       \multicolumn{1}{c}{Top-5 max. act. images of \cite{Singla2022salient} (\textbf{1 identical pair})} &\multicolumn{1}{c}{Top-5 max. act. images of NPCA (ours) (\textbf{5 identical pairs})}\\
        {\includegraphics[width=0.5\textwidth]{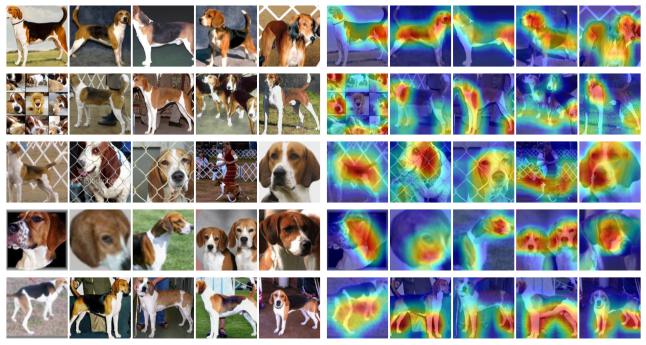}}
        &{\includegraphics[width=0.5\textwidth]{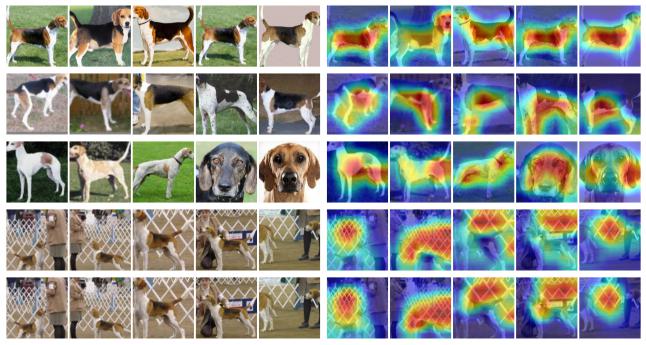}} 
        
    \end{tabular}
    \caption{\label{fig:worst12_compare_figure_2}\textbf{%
    Classes where NPCA has the most identical pairs among the top-5 maximally activating images across different NPCA components (NPCA and neurons computed on the same model, Rob. ResNet50\cite{Singla2022salient}).} For each method we provide the number of identical pairs of images across different neurons for \cite{Singla2022salient} resp. different components for NPCA as described in App.~\ref{app:comparison}. Each row shows the 5 maximally activating images together with the corresponding GradCAM heatmaps for the top-5 neurons of \cite{Singla2022salient} (left) and the top-5 NPCA components (right), respectively. While some of our NPCA components have identical pairs, the highest number of them (\textbf{10}) is almost three times smaller than the largest number of identical pairs of \cite{Singla2022salient} (\textbf{28}). These examples show that even the worst classes for the NPCA components only contain a few overlapping subpopulations. This aligns with the observations in Fig.~\ref{fig:hist-zeros}. Therefore, the problem of a lack of diversity in the detected features is much less severe for the NPCA components than for the neurons of \cite{Singla2022salient}.}%
    \vspace{10cm}
\end{figure*}

%% file: spurious_examples_2.tex
\begin{figure*}[h!]
    \setlength{\tabcolsep}{0.1em}
    \centering
    \normalsize
    \begin{tabular}{c c c c c p{0.02\textwidth} c c c c c}
       \multicolumn{5}{l}{{\blue\textbf{Gondola}} -  Random train. images (\textbf{confidence}\,\,/\,$\alpha_k$)} 
         &&\multicolumn{5}{c}{Images with spurious \textbf{houses/river} but \textbf{no gondola}}\\ 
         {\includegraphics[width=0.095\textwidth]{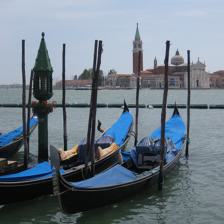}} 
         &{\includegraphics[width=0.095\textwidth]{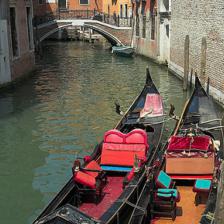}}
         &{\includegraphics[width=0.095\textwidth]{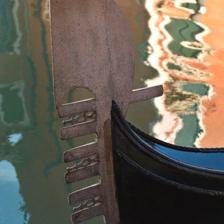}}
         &{\includegraphics[width=0.095\textwidth]{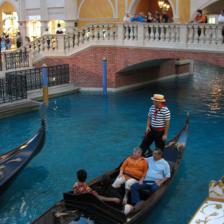}} 
         &{\includegraphics[width=0.095\textwidth]{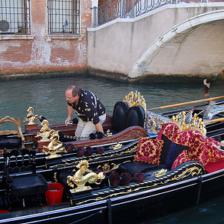}}
         &&%
         {\includegraphics[width=0.095\textwidth]{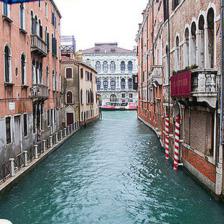}}
         & {\includegraphics[width=0.095\textwidth]{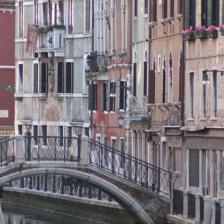}}
         &{\includegraphics[width=0.095\textwidth]{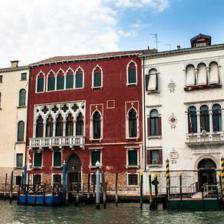}}
         &{\includegraphics[width=0.095\textwidth]{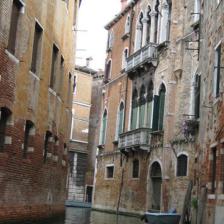}}
         &{\includegraphics[width=0.095\textwidth]{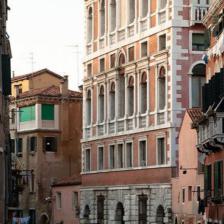}}
         \\
         $\mathbf{1.00}$\,/\,$3.2$
         &$\mathbf{1.00}$\,/\,$0.6$
         &$\mathbf{0.98}$\,/\,$-3.8$
         &$\mathbf{1.00}$\,/\,$1.2$
         &$\mathbf{0.98}$\,/\,$-2.0$
         &&$\mathbf{0.92}$\,/\,$6.4$
         &$\mathbf{0.82}$\,/\,$4.4$
         &$\mathbf{0.73}$\,/\,$3.1$
         &$\mathbf{0.85}$\,/\,$4.9$
         &$\mathbf{0.82}$\,/\,$4.1$\\
         \multicolumn{1}{c}{\NPFV-1} & \multicolumn{4}{c}{Max. activating train. images - NPCA Comp. 1} && \multicolumn{5}{c}{all classified as \textbf{gondola} by four ImageNet models}\\
         {\includegraphics[width=0.095\textwidth]{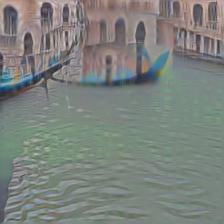}}& 
         {\includegraphics[width=0.095\textwidth]{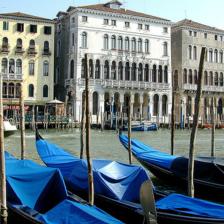}}&  {\includegraphics[width=0.095\textwidth]{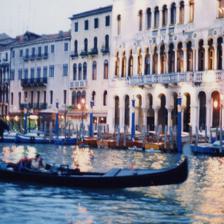}}&  {\includegraphics[width=0.095\textwidth]{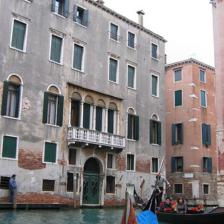}} &
         {\includegraphics[width=0.095\textwidth]{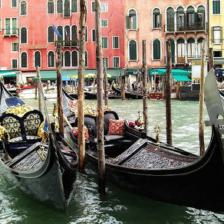}}&&
         {\includegraphics[width=0.095\textwidth]{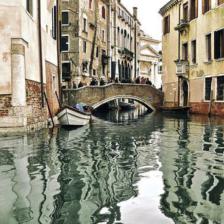}}
         &  {\includegraphics[width=0.095\textwidth]{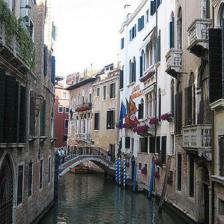}}
         &  {\includegraphics[width=0.095\textwidth]{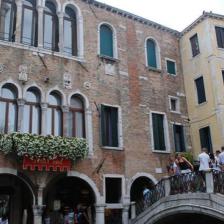}}
         &  {\includegraphics[width=0.095\textwidth]{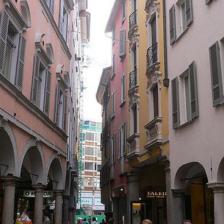}}
         &{\includegraphics[width=0.095\textwidth]{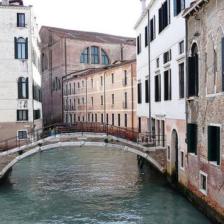}}\\
         $\mathbf{1.00}$\,/\,$6.9$
         &$\mathbf{1.00}$\,/\,$7.7$
         &$\mathbf{1.00}$\,/\,$6.9$
         &$\mathbf{0.96}$\,/\,$6.9$
         &$\mathbf{1.00}$\,/\,$6.9$
         &&$\mathbf{0.88}$\,/\,$4.5$
         &$\mathbf{0.90}$\,/\,$5.7$
         &$\mathbf{0.74}$\,/\,$5.4$
         &$\mathbf{0.79}$\,/\,$5.6$
         &$\mathbf{0.90}$\,/\,$5.6$\\
         \hline
         \multicolumn{5}{l}{{\blue\textbf{Racket}} - Random train. images (\textbf{confidence}\,\,/\,$\alpha_k$)} 
        &&\multicolumn{5}{c}{Images with spurious \textbf{tennis court/player} but \textbf{no racket}}\\ 
        {\includegraphics[width=0.095\textwidth]{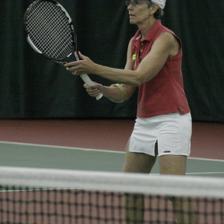}} 
        &{\includegraphics[width=0.095\textwidth]{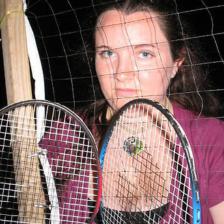}}
        &{\includegraphics[width=0.095\textwidth]{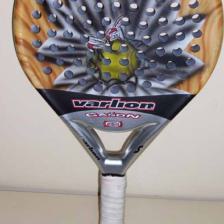}}
        &{\includegraphics[width=0.095\textwidth]{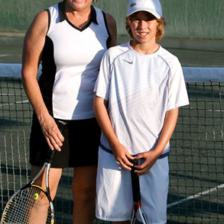}} 
        &{\includegraphics[width=0.095\textwidth]{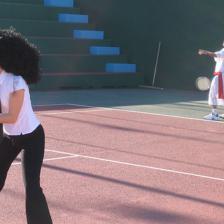}}
        &%
        &{\includegraphics[width=0.095\textwidth]{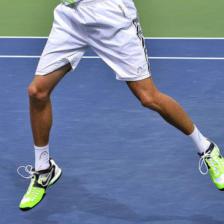}}
        &{\includegraphics[width=0.095\textwidth]{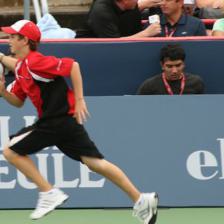}}
        &{\includegraphics[width=0.095\textwidth]{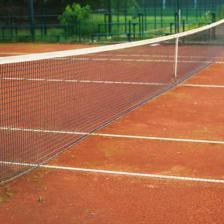}}
        &{\includegraphics[width=0.095\textwidth]{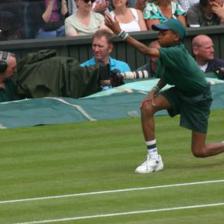}}
        &{\includegraphics[width=0.095\textwidth]{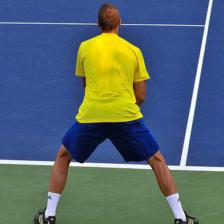}}\\
        $\mathbf{0.93}$\,/\,$0.7$
        & $\mathbf{0.38}$\,/\,$-3.0$
        & $\mathbf{0.62}$\,/\,$-2.4$
        & $\mathbf{0.12}$\,/\,$-3.1$
        & $\mathbf{0.97}$\,/\,$2.1$
        &&  $\mathbf{0.82}$\,/\,$6.2$ %
        & $\mathbf{0.94}$\,/\,$4.3$
        &$\mathbf{0.90}$\,/\,$3.8$
        &$\mathbf{0.97}$\,/\,$3.7$
        &$\mathbf{0.76}$\,/\,$5.9$\\
        \multicolumn{1}{c}{\NPFV-5} & \multicolumn{4}{c}{Max. activating train. images - NPCA Comp. 5} && \multicolumn{5}{c}{all classified as \textbf{racket} by four ImageNet models}\\
        {\includegraphics[width=0.095\textwidth]{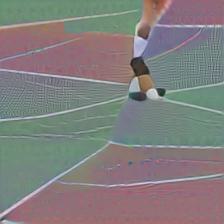}}& 
        {\includegraphics[width=0.095\textwidth]{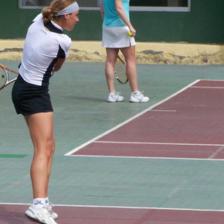}}&  {\includegraphics[width=0.095\textwidth]{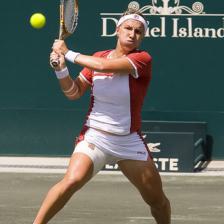}}&  {\includegraphics[width=0.095\textwidth]{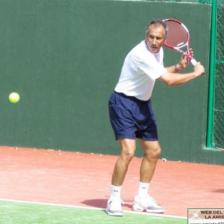}} &
        {\includegraphics[width=0.095\textwidth]{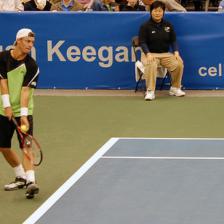}}&&
        {\includegraphics[width=0.095\textwidth]{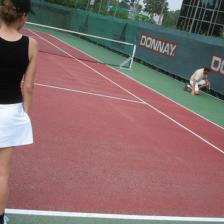}}
        &{\includegraphics[width=0.095\textwidth]{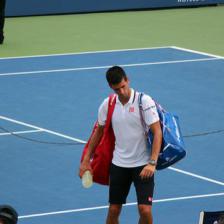}}
        &{\includegraphics[width=0.095\textwidth]{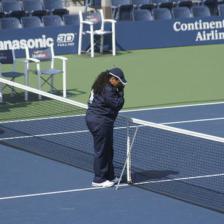}}
        &{\includegraphics[width=0.095\textwidth]{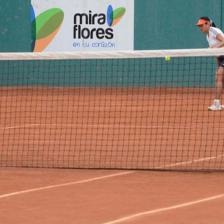}}
        &{\includegraphics[width=0.095\textwidth]{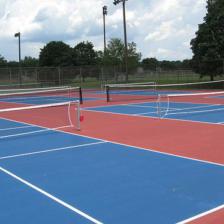}}\\
        $\mathbf{1.00}$\,/\,$17.8$
        & $\mathbf{0.78}$\,/\,$7.5$
        & $\mathbf{1.00}$\,/\,$7.2$
        & $\mathbf{1.00}$\,/\,$7.0$
        & $\mathbf{1.00}$\,/\,$7.0$
        && $\mathbf{0.94}$\,/\,$5.7$
        &$\mathbf{0.83}$\,/\,$4.7$
        &$\mathbf{0.76}$\,/\,$4.2$
        &$\mathbf{0.90}$\,/\,$3.8$
        &$\mathbf{0.56}$\,/\,$3.8$\\
        \hline
          \multicolumn{5}{l}{{\blue\textbf{Dam}} -    Random train. images (\textbf{confidence}\,\,/\,$\alpha_k$)} 
         &&\multicolumn{5}{c}{Images with spurious \textbf{waterfall} but \textbf{no dam}}\\ 
         {\includegraphics[width=0.095\textwidth]{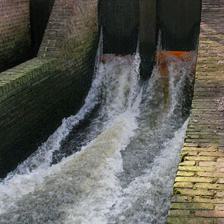}} 
         &{\includegraphics[width=0.095\textwidth]{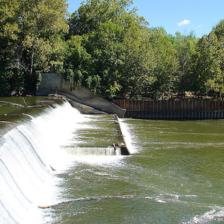}}
         &{\includegraphics[width=0.095\textwidth]{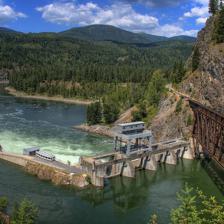}}
         &{\includegraphics[width=0.095\textwidth]{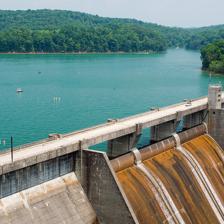}} 
         &{\includegraphics[width=0.095\textwidth]{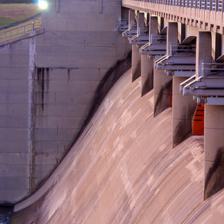}}
         &&{\includegraphics[width=0.095\textwidth]{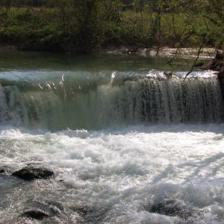}}
         &{\includegraphics[width=0.095\textwidth]{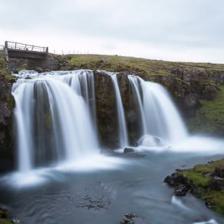}}
         &{\includegraphics[width=0.095\textwidth]{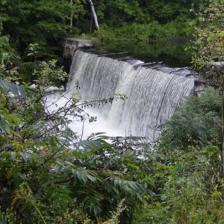}}
         &{\includegraphics[width=0.095\textwidth]{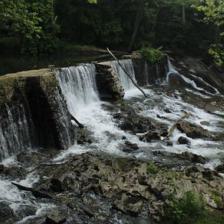}}
         &{\includegraphics[width=0.095\textwidth]{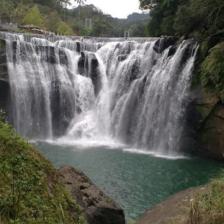}}\\
         $\mathbf{0.44}$\,/\,$0.1$
         &$\mathbf{0.52}$\,/\,$0.1$
         &$\mathbf{0.28}$\,/\,$0.0$
         &$\mathbf{1.00}$\,/\,$-0.0$
         &$\mathbf{0.85}$\,/\,$-0.0$
         &&$\mathbf{0.30}$\,/\,$0.1$
         &$\mathbf{0.99}$\,/\,$0.1$
         &$\mathbf{0.67}$\,/\,$0.1$
         &$\mathbf{0.55}$\,/\,$0.1$ 
         &$\mathbf{0.99}$\,/\,$0.1$\\
         \multicolumn{1}{c}{\NPFV-1} & \multicolumn{4}{c}{Max. activating train. images - NPCA Comp. 1} && \multicolumn{5}{c}{all classified as \textbf{dam} by four ImageNet models}\\
         {\includegraphics[width=0.095\textwidth]{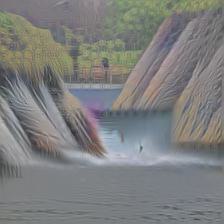}}& 
         {\includegraphics[width=0.095\textwidth]{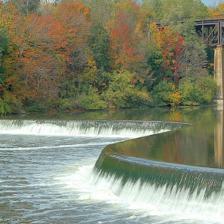}}&  {\includegraphics[width=0.095\textwidth]{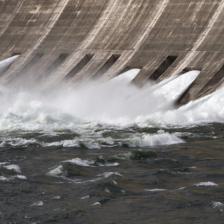}}&  {\includegraphics[width=0.095\textwidth]{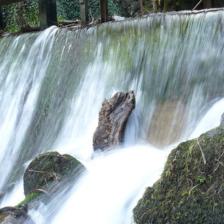}} &
         {\includegraphics[width=0.095\textwidth]{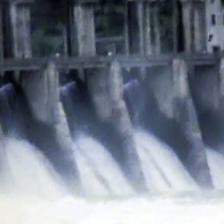}}&  &
         {\includegraphics[width=0.095\textwidth]{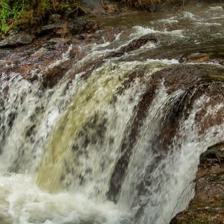}}&  {\includegraphics[width=0.095\textwidth]{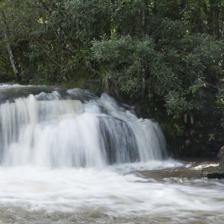}}&  {\includegraphics[width=0.095\textwidth]{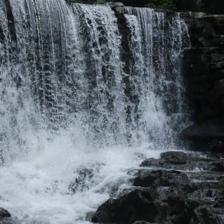}}&  {\includegraphics[width=0.095\textwidth]{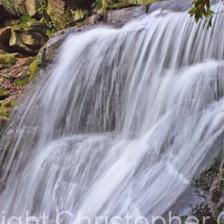}}&{\includegraphics[width=0.095\textwidth]{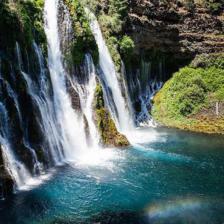}}\\
         $\mathbf{1.00}$\,/\,$0.2$
         &$\mathbf{1.00}$\,/\,$0.1$
         &$\mathbf{0.97}$\,/\,$0.1$
         &$\mathbf{1.00}$\,/\,$0.1$
         &$\mathbf{0.70}$\,/\,$0.1$
         &&$\mathbf{0.55}$\,/\,$0.1$
         &$\mathbf{0.63}$\,/\,$0.1$
         &$\mathbf{0.62}$\,/\,$0.1$
         &$\mathbf{0.76}$\,/\,$0.1$ 
         &$\mathbf{0.77}$\,/\,$0.1$ \\
         \hline
         \multicolumn{5}{l}{{\blue\textbf{Flagpole}}  -  Random train. images (\textbf{confidence}\,\,/\,$\alpha_k$)}  
         &&\multicolumn{5}{c}{Images with spurious \textbf{US flag} but \textbf{no flag pole}}\\ 
         {\includegraphics[width=0.095\textwidth]{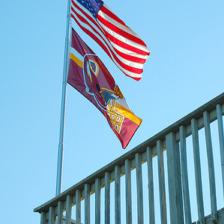}} 
         &{\includegraphics[width=0.095\textwidth]{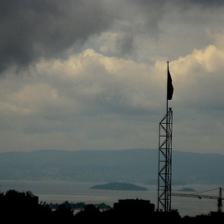}}
         &{\includegraphics[width=0.095\textwidth]{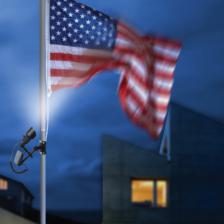}}
         &{\includegraphics[width=0.095\textwidth]{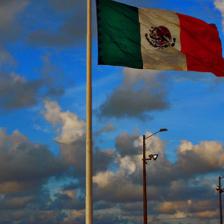}} 
         &{\includegraphics[width=0.095\textwidth]{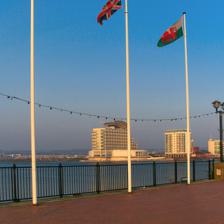}}
         &&{\includegraphics[width=0.095\textwidth]{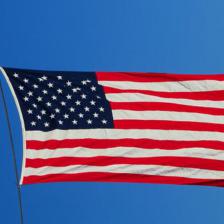}}
         &{\includegraphics[width=0.095\textwidth]{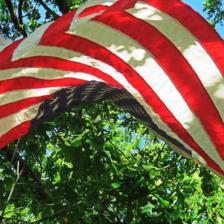}}
         &{\includegraphics[width=0.095\textwidth]{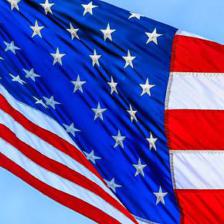}}
         &{\includegraphics[width=0.095\textwidth]{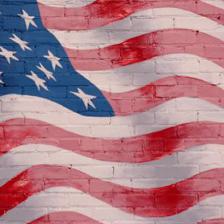}}
         &{\includegraphics[width=0.095\textwidth]{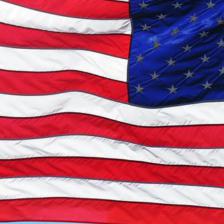}}\\
         $\mathbf{1.00}$\,/\,$4.9$
         &$\mathbf{0.25}$\,/\,$-3.9$
         &$\mathbf{1.00}$\,/\,$5.4$
         &$\mathbf{0.86}$\,/\,$-0.4$
         &$\mathbf{0.99}$\,/\,$-1.3$
         &&$\mathbf{1.00}$\,/\,$8.3$
         &$\mathbf{0.98}$\,/\,$3.6$
         &$\mathbf{0.96}$\,/\,$6.6$
         &$\mathbf{0.38}$\,/\,$3.4$
         &$\mathbf{0.99}$\,/\,$5.4$\\
         \multicolumn{1}{c}{\NPFV-1} & \multicolumn{4}{c}{Max. activating train. images - NPCA Comp. 1} && \multicolumn{5}{c}{all classified as \textbf{flag pole} by four ImageNet models}\\
         {\includegraphics[width=0.095\textwidth]{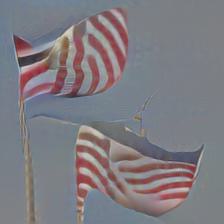}}& 
         {\includegraphics[width=0.095\textwidth]{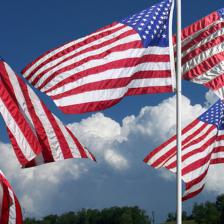}}&  {\includegraphics[width=0.095\textwidth]{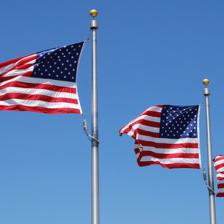}}&  {\includegraphics[width=0.095\textwidth]{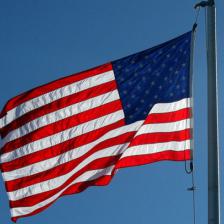}} &
         {\includegraphics[width=0.095\textwidth]{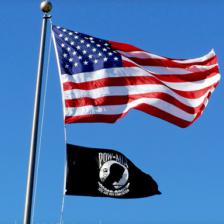}}&&
         {\includegraphics[width=0.095\textwidth]{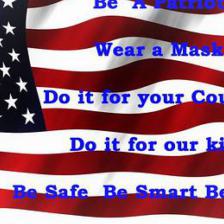}}&  {\includegraphics[width=0.095\textwidth]{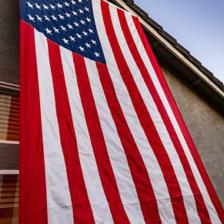}}&  {\includegraphics[width=0.095\textwidth]{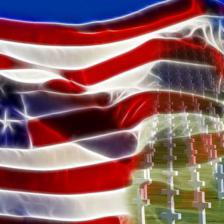}}&  {\includegraphics[width=0.095\textwidth]{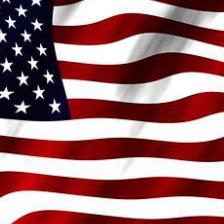}}&{\includegraphics[width=0.095\textwidth]{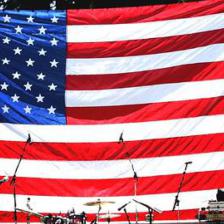}}\\
         $\mathbf{1.00}$\,/\,$16.1$
         &$\mathbf{1.00}$\,/\,$11.6$
         &$\mathbf{1.00}$\,/\,$11.1$
         &$\mathbf{1.00}$\,/\,$10.8$
         &$\mathbf{1.00}$\,/\,$10.8$
         &&$\mathbf{0.94}$\,/\,$5.1$
         &$\mathbf{1.00}$\,/\,$6.6$
         &$\mathbf{0.99}$\,/\,$5.1$
         &$\mathbf{1.00}$\,/\,$8.8$
         &$\mathbf{0.75}$\,/\,$4.2$
    \end{tabular}
    \caption{\label{fig:spurious-exp2} \textbf{Spurious features (ImageNet):} found by human labeling of our neural PCA components. For each class we show 5 random train. images (top left), the neural PCA Feature Visual. (NPFV) and 4 most activating train. images for the spurious feature component (bottom left). Right: four ImageNet models classify images \textbf{showing only the spurious feature but no class object} as this class.}
\end{figure*}

%% file: spurious_examples_3.tex
\begin{figure*}[h!]
    \setlength{\tabcolsep}{0.1em}
    \centering
    \normalsize
    \begin{tabular}{c c c c c p{0.02\textwidth} c c c c c}
          \multicolumn{5}{l}{{\blue\textbf{Hard disc}} -  Random train. images (\textbf{confidence}\,\,/\,$\alpha_k$)} 
         &&\multicolumn{5}{c}{Images with spurious \textbf{(serial) labels} but \textbf{no hard disc}}\\ 
         {\includegraphics[width=0.095\textwidth]{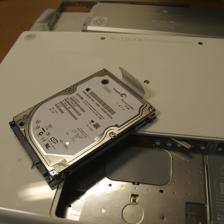}} 
         &{\includegraphics[width=0.095\textwidth]{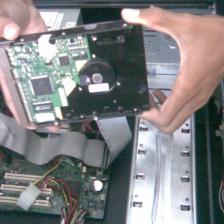}}
         &{\includegraphics[width=0.095\textwidth]{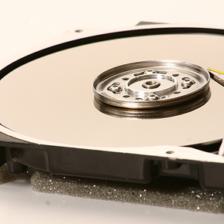}}
         &{\includegraphics[width=0.095\textwidth]{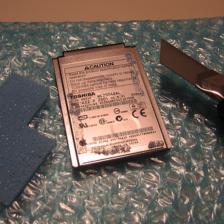}} 
         &{\includegraphics[width=0.095\textwidth]{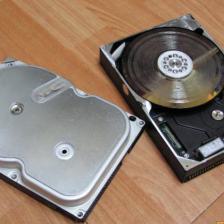}}
         &&{\includegraphics[width=0.095\textwidth]{figures/img/hard_disc/1.jpg}}
         &{\includegraphics[width=0.095\textwidth]{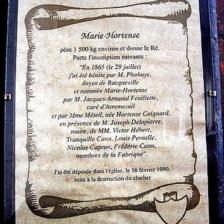}}
         &{\includegraphics[width=0.095\textwidth]{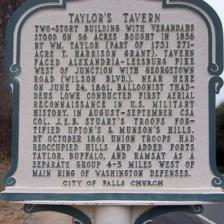}}
         &{\includegraphics[width=0.095\textwidth]{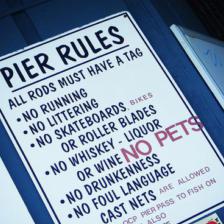}}
         &{\includegraphics[width=0.095\textwidth]{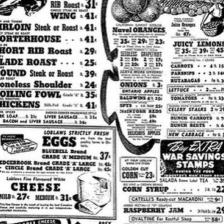}}\\
         $\mathbf{0.93}$\,/\,$0.3$
         &$\mathbf{0.03}$\,/\,$-3.3$
         &$\mathbf{1.00}$\,/\,$0.7$
         &$\mathbf{0.04}$\,/\,$-0.32$
         &$\mathbf{0.97}$\,/\,$-0.3$
         &&$\mathbf{0.74}$\,/\,$6.6$
         &$\mathbf{0.66}$\,/\,$6.5$
         &$\mathbf{0.68}$\,/\,$5.75$
         &$\mathbf{0.62}$\,/\,$5.4$
         &$\mathbf{0.65}$\,/\,$5.3$\\
         \multicolumn{1}{c}{\NPFV-1} & \multicolumn{4}{c}{Max. activating train. images - NPCA Comp. 1} && \multicolumn{5}{c}{all classified as \textbf{hard disc} by four ImageNet models}\\
         {\includegraphics[width=0.095\textwidth]{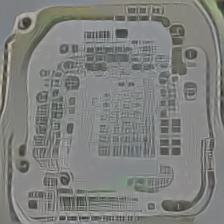}}& 
         {\includegraphics[width=0.095\textwidth]{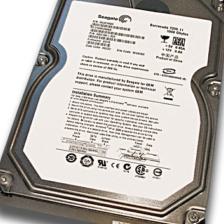}}&  {\includegraphics[width=0.095\textwidth]{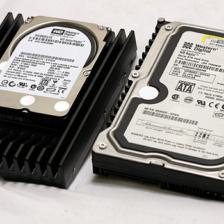}}&  {\includegraphics[width=0.095\textwidth]{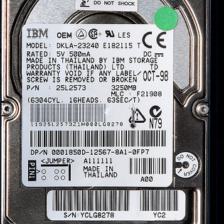}} &
         {\includegraphics[width=0.095\textwidth]{figures/img/hard_disc/act_pc_1_max_3.jpg}}&&
         {\includegraphics[width=0.095\textwidth]{figures/img/hard_disc/28.jpg}}&  {\includegraphics[width=0.095\textwidth]{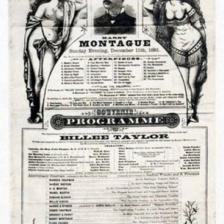}}&  {\includegraphics[width=0.095\textwidth]{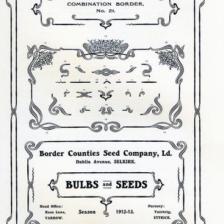}}&  {\includegraphics[width=0.095\textwidth]{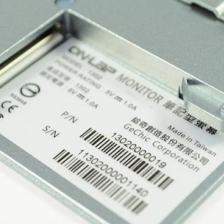}}&{\includegraphics[width=0.095\textwidth]{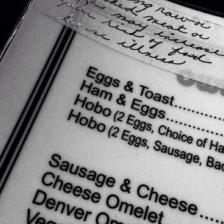}}\\
         $\mathbf{1.00}$\,/\,$14.0$
         &$\mathbf{1.00}$\,/\,$9.8$
         &$\mathbf{1.00}$\,/\,$8.8$
         &$\mathbf{1.00}$\,/\,$8.3$
         &$\mathbf{1.00}$\,/\,$8.3$
         &&$\mathbf{0.85}$\,/\,$4.9$
         &$\mathbf{0.59}$\,/\,$4.6$
         &$\mathbf{0.72}$\,/\,$4.3$
         &$\mathbf{0.83}$\,/\,$3.9$
         &$\mathbf{0.53}$\,/\,$3.9$\\
         \hline
         \multicolumn{5}{l}{{\blue\textbf{Snorkel}} -  Random train. images (\textbf{confidence}\,\,/\,$\alpha_k$)} 
        &&\multicolumn{5}{c}{Images with spurious \textbf{diver/human} but \textbf{no snorkel}}\\ 
        {\includegraphics[width=0.095\textwidth]{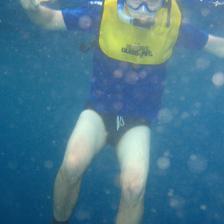}} 
        &{\includegraphics[width=0.095\textwidth]{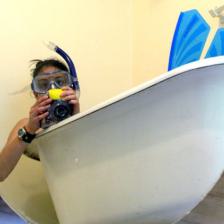}}
        &{\includegraphics[width=0.095\textwidth]{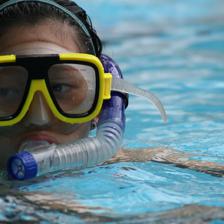}}
        &{\includegraphics[width=0.095\textwidth]{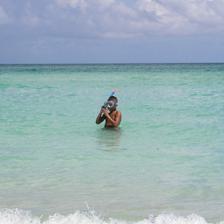}} 
        &{\includegraphics[width=0.095\textwidth]{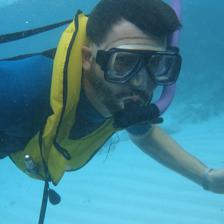}}
        &&{\includegraphics[width=0.095\textwidth]{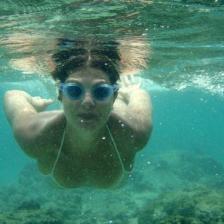}}
        &{\includegraphics[width=0.095\textwidth]{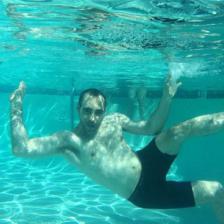}}
        &{\includegraphics[width=0.095\textwidth]{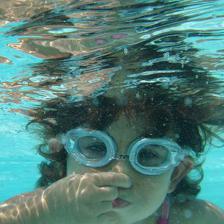}}
        &{\includegraphics[width=0.095\textwidth]{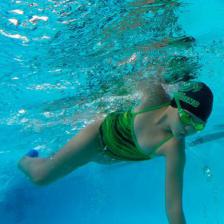}}
        &{\includegraphics[width=0.095\textwidth]{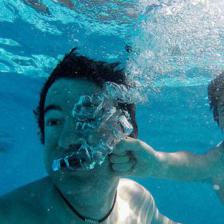}}\\
        $\mathbf{0.67}$\,/\,$-0.1$
         &$\mathbf{0.01}$\,/\,$-3.8$
         &$\mathbf{1.00}$\,/\,$2.4$
         &$\mathbf{0.17}$\,/\,$-2.3$
         &$\mathbf{0.84}$\,/\,$1.8$
         &&$\mathbf{0.90}$\,/\,$5.0$
         &$\mathbf{0.64}$\,/\,$4.3$
         &$\mathbf{0.83}$\,/\,$3.8$
         &$\mathbf{0.64}$\,/\,$3.8$
         &$\mathbf{0.74}$\,/\,$3.3$\\
        \multicolumn{1}{c}{\NPFV-1} & \multicolumn{4}{c}{Max. activating train. images - NPCA Comp. 1} && \multicolumn{5}{c}{all classified as \textbf{snorkel} by four ImageNet models}\\
        {\includegraphics[width=0.095\textwidth]{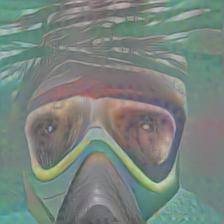}}& 
        {\includegraphics[width=0.095\textwidth]{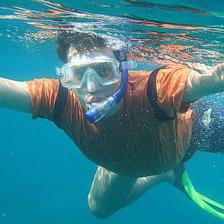}}&  {\includegraphics[width=0.095\textwidth]{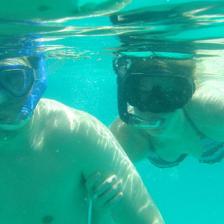}}&  {\includegraphics[width=0.095\textwidth]{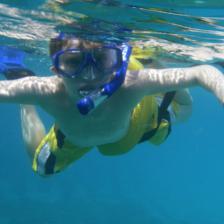}} &
        {\includegraphics[width=0.095\textwidth]{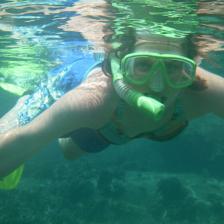}}&&
        {\includegraphics[width=0.095\textwidth]{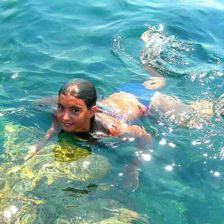}}&  {\includegraphics[width=0.095\textwidth]{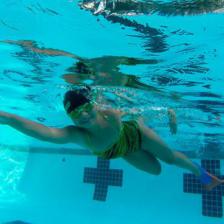}}&  {\includegraphics[width=0.095\textwidth]{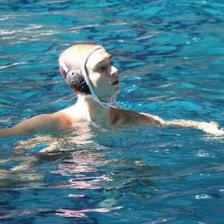}}&  {\includegraphics[width=0.095\textwidth]{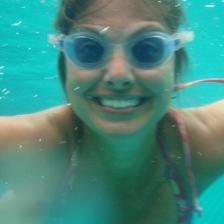}}&{\includegraphics[width=0.095\textwidth]{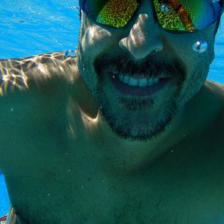}}\\
        $\mathbf{1.00}$\,/\,$9.6$
         &$\mathbf{1.00}$\,/\,$6.9$
         &$\mathbf{1.00}$\,/\,$5.9$
         &$\mathbf{1.00}$\,/\,$5.9$
         &$\mathbf{0.97}$\,/\,$5.8$
        &&$\mathbf{0.71}$\,/\,$3.1$
         &$\mathbf{0.61}$\,/\,$3.1$
         &$\mathbf{0.59}$\,/\,$3.0$
         &$\mathbf{0.59}$\,/\,$2.8$
         &$\mathbf{0.55}$\,/\,$2.8$\\
        \hline
        \multicolumn{5}{l}{{\blue\textbf{Mountain bike}} - Random train. images (\textbf{confidence}\,\,/\,$\alpha_k$)} 
        &&\multicolumn{5}{c}{Images with spurious \textbf{forest} but \textbf{no mountain bike}}\\ 
        {\includegraphics[width=0.095\textwidth]{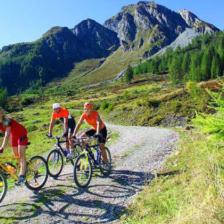}} 
        &{\includegraphics[width=0.095\textwidth]{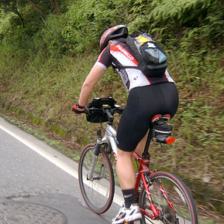}}
        &{\includegraphics[width=0.095\textwidth]{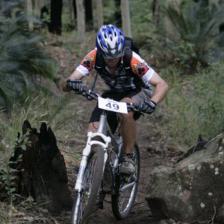}}
        &{\includegraphics[width=0.095\textwidth]{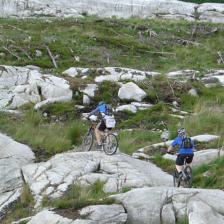}} 
        &{\includegraphics[width=0.095\textwidth]{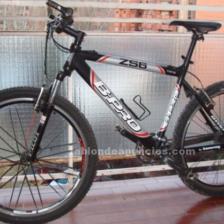}}
        &&{\includegraphics[width=0.095\textwidth]{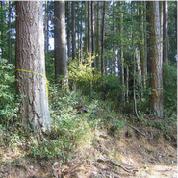}}
        &{\includegraphics[width=0.095\textwidth]{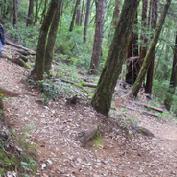}}
        &{\includegraphics[width=0.095\textwidth]{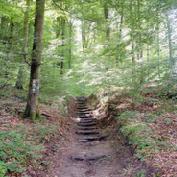}}
        &{\includegraphics[width=0.095\textwidth]{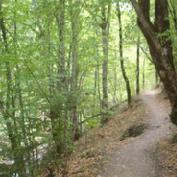}}
        &{\includegraphics[width=0.095\textwidth]{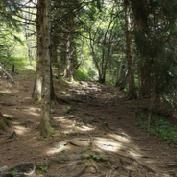}}\\
        $\mathbf{0.46}$\,/\,$-0.2$
        & $\mathbf{0.70}$\,/\,$0.1$
        & $\mathbf{0.99}$\,/\,$0.1$
        & $\mathbf{0.00}$\,/\,$0.0$
        & $\mathbf{0.55}$\,/\,$-0.1$
        && $\mathbf{0.31}$\,/\,$0.3$
        & $\mathbf{0.60}$\,/\,$0.3$
        & $\mathbf{0.33}$\,/\,$0.3$
        & $\mathbf{0.36}$\,/\,$0.3$
        & $\mathbf{0.35}$\,/\,$0.3$\\ 
        \multicolumn{1}{c}{\NPFV-1} & \multicolumn{4}{c}{Max. activating train. images - NPCA Comp. 1} && \multicolumn{5}{c}{all classified as \textbf{mountain bike} by four ImageNet models}\\
        {\includegraphics[width=0.095\textwidth]{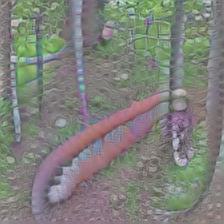}}& 
        {\includegraphics[width=0.095\textwidth]{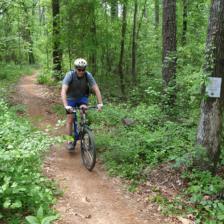}}&  {\includegraphics[width=0.095\textwidth]{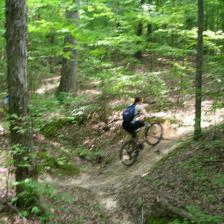}}&  {\includegraphics[width=0.095\textwidth]{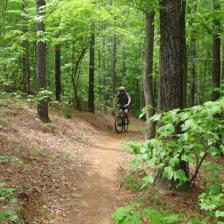}} &
        {\includegraphics[width=0.095\textwidth]{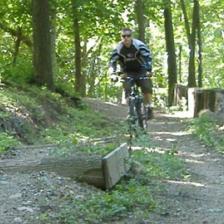}}&&
        {\includegraphics[width=0.095\textwidth]{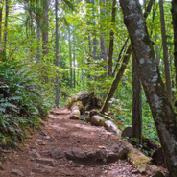}}&  {\includegraphics[width=0.095\textwidth]{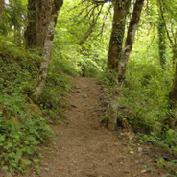}}&  {\includegraphics[width=0.095\textwidth]{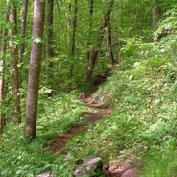}}&  {\includegraphics[width=0.095\textwidth]{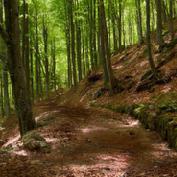}}&{\includegraphics[width=0.095\textwidth]{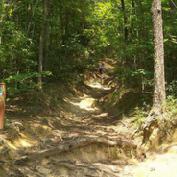}}\\
        $\mathbf{0.93}$\,/\,$0.5$
        & $\mathbf{0.96}$\,/\,$0.3$
        & $\mathbf{0.31}$\,/\,$0.3$
        & $\mathbf{0.24}$\,/\,$0.3$
        & $\mathbf{0.42}$\,/\,$0.3$
        && $\mathbf{0.45}$\,/\,$0.3$
        & $\mathbf{0.37}$\,/\,$0.3$
        & $\mathbf{0.34}$\,/\,$0.3$
        & $\mathbf{0.52}$\,/\,$0.3$
        & $\mathbf{0.53}$\,/\,$0.3$\\
        \hline
         \multicolumn{5}{l}{{\blue\textbf{Indigo Bunting}} -  Random train. images (\textbf{confidence}\,\,/\,$\alpha_k$)} 
         &&\multicolumn{5}{c}{Images with spurious \textbf{twigs} but \textbf{no indigo bunting}}\\ 
         {\includegraphics[width=0.095\textwidth]{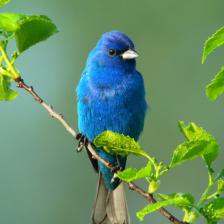}} 
         &{\includegraphics[width=0.095\textwidth]{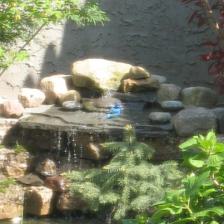}}
         &{\includegraphics[width=0.095\textwidth]{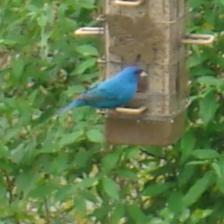}}
         &{\includegraphics[width=0.095\textwidth]{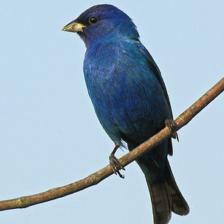}} 
         &{\includegraphics[width=0.095\textwidth]{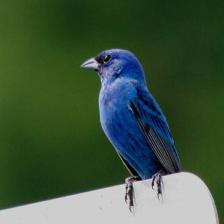}}
         &&{\includegraphics[width=0.095\textwidth]{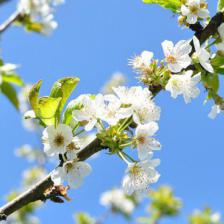}}
         &{\includegraphics[width=0.095\textwidth]{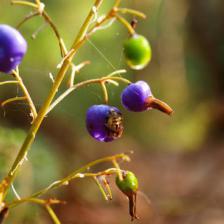}}
         &{\includegraphics[width=0.095\textwidth]{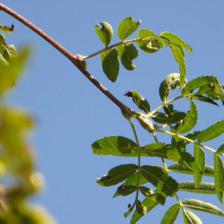}}
         &{\includegraphics[width=0.095\textwidth]{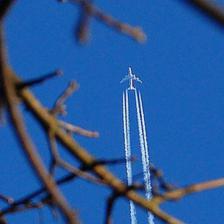}}
         &{\includegraphics[width=0.095\textwidth]{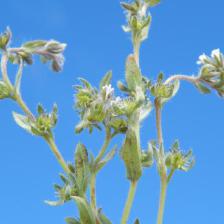}}\\
         $\mathbf{1.00}$\,/\,$2.1$
         &$\mathbf{0.00}$\,/\,$-2.1$
         &$\mathbf{1.00}$\,/\,$-1.1$
         &$\mathbf{1.00}$\,/\,$1.2$
         &$\mathbf{1.00}$\,/\,$0.4$
         &&$\mathbf{0.56}$\,/\,$2.8$
         &$\mathbf{0.49}$\,/\,$2.4$
         &$\mathbf{0.30}$\,/\,$2.4$
         &$\mathbf{0.16}$\,/\,$2.3$
         &$\mathbf{0.35}$\,/\,$2.2$\\
         \multicolumn{1}{c}{\NPFV-3} & \multicolumn{4}{c}{Max. activating train. images - NPCA Comp. 3} && \multicolumn{5}{c}{all classified as \textbf{indigo bunting} by four ImageNet models}\\
         {\includegraphics[width=0.095\textwidth]{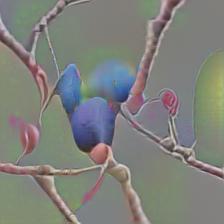}}& 
         {\includegraphics[width=0.095\textwidth]{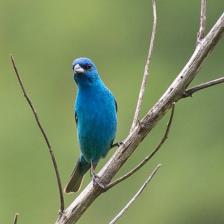}}&  {\includegraphics[width=0.095\textwidth]{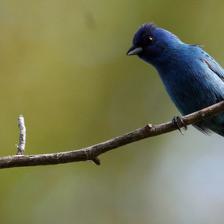}}&  {\includegraphics[width=0.095\textwidth]{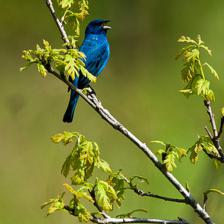}} &
         {\includegraphics[width=0.095\textwidth]{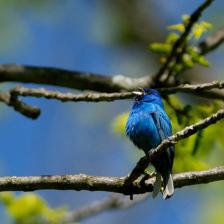}}&&
         {\includegraphics[width=0.095\textwidth]{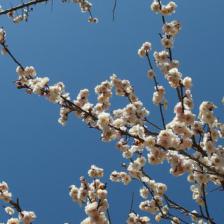}}&  {\includegraphics[width=0.095\textwidth]{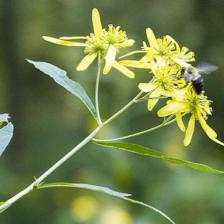}}&  {\includegraphics[width=0.095\textwidth]{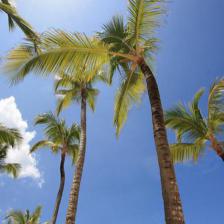}}&  {\includegraphics[width=0.095\textwidth]{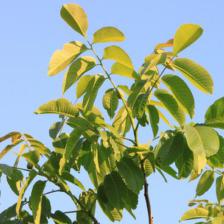}}&{\includegraphics[width=0.095\textwidth]{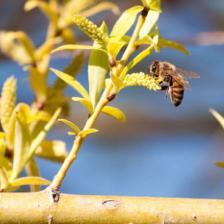}}\\
         $\mathbf{1.00}$\,/\,$7.7$
         &$\mathbf{1.00}$\,/\,$3.8$
         &$\mathbf{0.98}$\,/\,$3.6$
         &$\mathbf{1.00}$\,/\,$3.6$
         &$\mathbf{0.98}$\,/\,$3.5$
         &&$\mathbf{0.36}$\,/\,$2.2$
         &$\mathbf{0.37}$\,/\,$2.1$
         &$\mathbf{0.50}$\,/\,$1.8$
         &$\mathbf{0.44}$\,/\,$1.8$
         &$\mathbf{0.55}$\,/\,$1.7$
    \end{tabular}
    \caption{\label{fig:spurious-exp3}\textbf{Spurious features (ImageNet):} found by human labeling of our neural PCA components. For each class we show 5 random train. images (top left), the neural PCA Feature Visual. (NPFV) and 4 most activating train. images for the spurious feature component (bottom left). Right: four ImageNet models classify images \textbf{showing only the spurious feature but no class object} as this class.}
\end{figure*}

%% file: appendix_spurious_imagenet.tex
\begin{figure*}
\scriptsize
\begin{tabular}{c | c | c | c | c }
	{tench (humans,1)} & {g. white shark (water/foam, 2)} & {indigo bunting (twigs, 3)} & {agama (bark, 8)} & {a. alligator (vegetation, 10)}\\
	{\includegraphics[width=0.1800\textwidth]{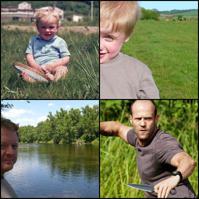}}&
	{\includegraphics[width=0.1800\textwidth]{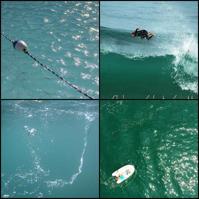}}&
	{\includegraphics[width=0.1800\textwidth]{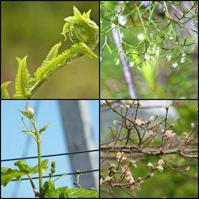}}&
	{\includegraphics[width=0.1800\textwidth]{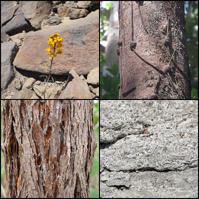}}&
	{\includegraphics[width=0.1800\textwidth]{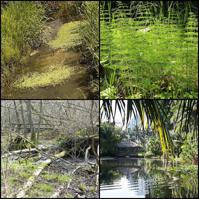}}\\
	\hline
	{water snake (water veg., 5)} & {black grouse (steppe/twigs, 6)} & {ptarmigan (snow, 2)} & {ruffed grouse (twigs/snow, 5)} & {s.-c. cockatoo (branches 3)}\\
	{\includegraphics[width=0.1800\textwidth]{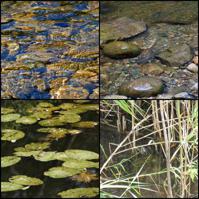}}&
	{\includegraphics[width=0.1800\textwidth]{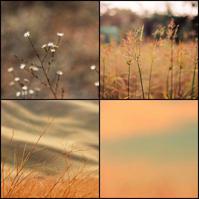}}&
	{\includegraphics[width=0.1800\textwidth]{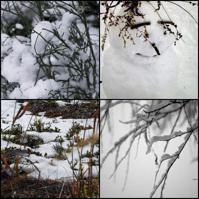}}&
	{\includegraphics[width=0.1800\textwidth]{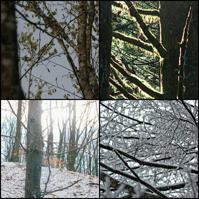}}&
	{\includegraphics[width=0.1800\textwidth]{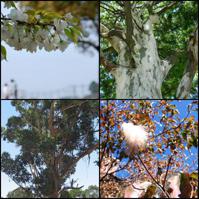}}\\
	\hline
	{coucal (branches/bush, 8)} & {hummingbird {\tiny (red feeder/flower, 2)}} & {platypus (water surface, 4)} & {koala (eucalyptus plants, 3)} & {grey whale (open sea, 6)}\\
	{\includegraphics[width=0.1800\textwidth]{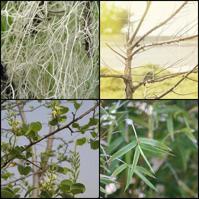}}&
	{\includegraphics[width=0.1800\textwidth]{figures/spurious_figure_0/11.jpg}}&
	{\includegraphics[width=0.1800\textwidth]{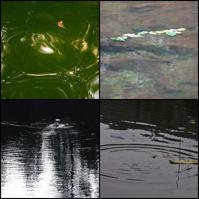}}&
	{\includegraphics[width=0.1800\textwidth]{figures/spurious_figure_0/13.jpg}}&
	{\includegraphics[width=0.1800\textwidth]{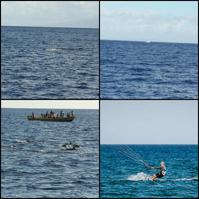}}\\
	\hline
	{leopard (tree/bark, 9)} & {sloth bear (stones/trunks, 3)} & {bee (violet flowers, 2)} & {walking stick (leaves, 3)} & {cabbage butterfly (flowers, 3)}\\
	{\includegraphics[width=0.1800\textwidth]{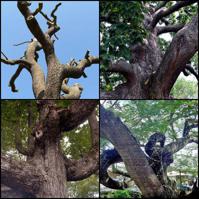}}&
	{\includegraphics[width=0.1800\textwidth]{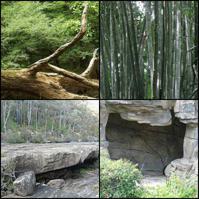}}&
	{\includegraphics[width=0.1800\textwidth]{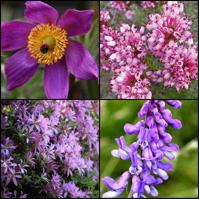}}&
	{\includegraphics[width=0.1800\textwidth]{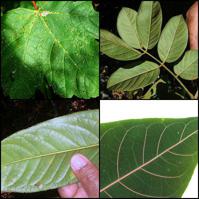}}&
	{\includegraphics[width=0.1800\textwidth]{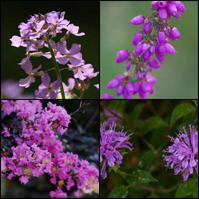}}\\
	\hline
	{sulphur butterfly (flowers, 2)} & {fox squirrel (tree/branches, 7)} & {hartebeest (steppe/straw, 9)} & {t.-t. sloth (branches/leaves, 6)} & {howler monkey (branches, 2)}\\
	{\includegraphics[width=0.1800\textwidth]{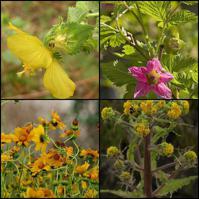}}&
	{\includegraphics[width=0.1800\textwidth]{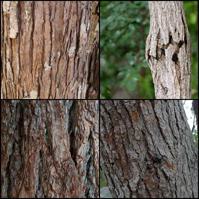}}&
	{\includegraphics[width=0.1800\textwidth]{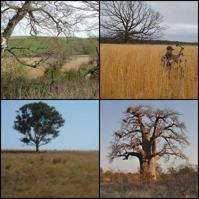}}&
	{\includegraphics[width=0.1800\textwidth]{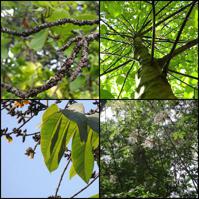}}&
	{\includegraphics[width=0.1800\textwidth]{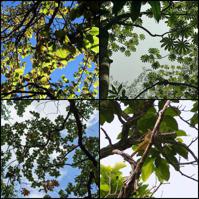}}\\
\end{tabular}
\caption{\label{fig:app_spurious_random1}Random selection of 4 images for classes 1-25 of our ``Spurious ImageNet'' dataset with class label (spurious feature, NPCA component). Note that the labels of spurious features are coarse and thus overlap e.g. several are leaves/branches/fowers. We are not able to identify if these are special trees or flowers which might be more specific. }
\end{figure*}

\begin{figure*}
\scriptsize\begin{tabular}{c | c | c | c | c }
	{indri (branches/leaves, 6)} & {barracouta (humans/hands, 8)} & {sturgeon (humans, 2)} & {gar (humans in white shirts, 5)} & {academic gown (woman, 7)}\\
	{\includegraphics[width=0.1800\textwidth]{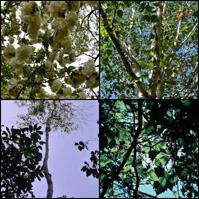}}&
	{\includegraphics[width=0.1800\textwidth]{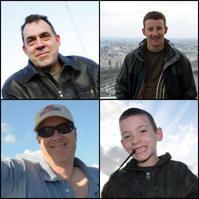}}&
	{\includegraphics[width=0.1800\textwidth]{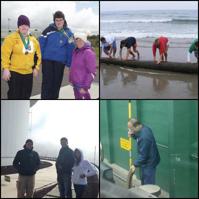}}&
	{\includegraphics[width=0.1800\textwidth]{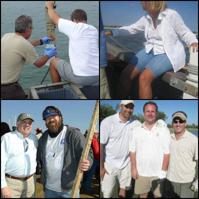}}&
	{\includegraphics[width=0.1800\textwidth]{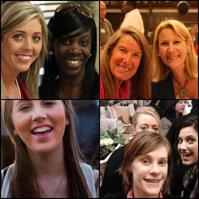}}\\
	\hline
	{bakery (storefront, 6)} & {balance beam (gymnasts, 1)} & {Band Aid (labels, 1)} & {barbershop (store front, 1)} & {barn (trees, 7)}\\
	{\includegraphics[width=0.1800\textwidth]{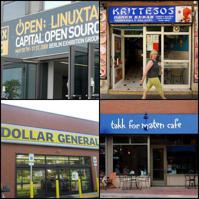}}&
	{\includegraphics[width=0.1800\textwidth]{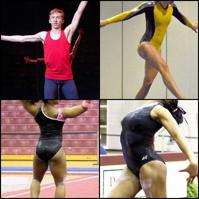}}&
	{\includegraphics[width=0.1800\textwidth]{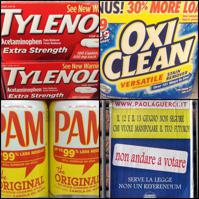}}&
	{\includegraphics[width=0.1800\textwidth]{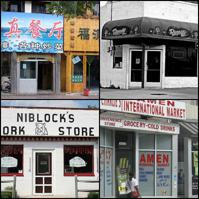}}&
	{\includegraphics[width=0.1800\textwidth]{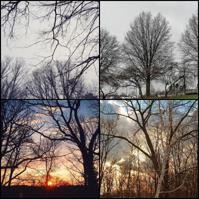}}\\
	\hline
	{bathing cap (humans/pool, 1)} & {bath towel (baby, 2)} & {bathtub (baby/children, 4)} & {beer bottle (colorful label, 1)} & {bikini (black-white images, 3)}\\
	{\includegraphics[width=0.1800\textwidth]{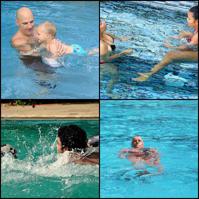}}&
	{\includegraphics[width=0.1800\textwidth]{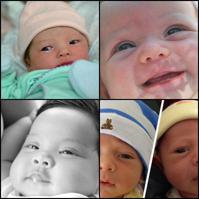}}&
	{\includegraphics[width=0.1800\textwidth]{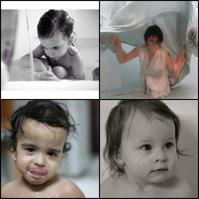}}&
	{\includegraphics[width=0.1800\textwidth]{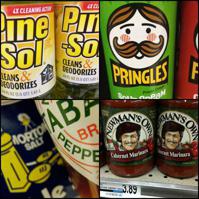}}&
	{\includegraphics[width=0.1800\textwidth]{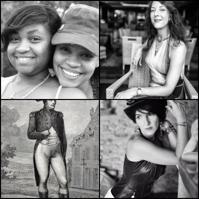}}\\
	\hline
	{bookshop (storefront, 1)} & {bulletproof vest (text, 5)} & {bullet train (train station, 7)} & {chain mail (face, 5)} & {chain saw (human worker, 2)}\\
	{\includegraphics[width=0.1800\textwidth]{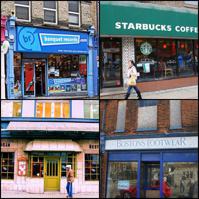}}&
	{\includegraphics[width=0.1800\textwidth]{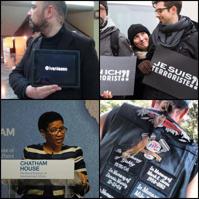}}&
	{\includegraphics[width=0.1800\textwidth]{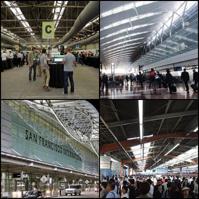}}&
	{\includegraphics[width=0.1800\textwidth]{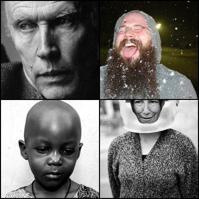}}&
	{\includegraphics[width=0.1800\textwidth]{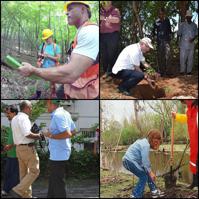}}\\
	\hline
	{cowboy hat (human/face, 4)} & {cradle (baby, 7)} & {dam (waterfall, 1)} & {dogsled (snow, 7)} & {dumbbell (athlete, 5)}\\
	{\includegraphics[width=0.1800\textwidth]{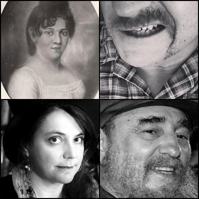}}&
	{\includegraphics[width=0.1800\textwidth]{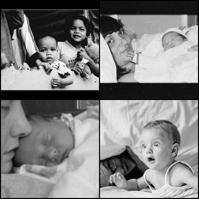}}&
	{\includegraphics[width=0.1800\textwidth]{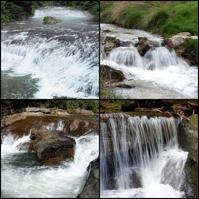}}&
	{\includegraphics[width=0.1800\textwidth]{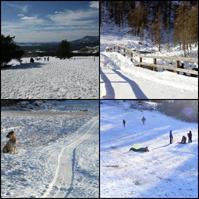}}&
	{\includegraphics[width=0.1800\textwidth]{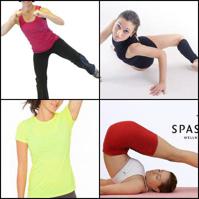}}\\
\end{tabular}

\caption{\label{fig:app_spurious_random2}Random selection of 4 images for classes 26-50 of our ``Spurious ImageNet'' dataset  with class label (spurious feature, NPCA component).}
\end{figure*}

\begin{figure*}
\scriptsize\begin{tabular}{c | c | c | c | c }
	{fireboat (water jet, 2)} & {flagpole (US flag, 1)} & {fountain pen (hand-written text, 10)} & {freight car (graffiti, 1)} & {gondola (houses, 1)}\\
	{\includegraphics[width=0.1800\textwidth]{figures/spurious_figure_2/0.jpg}}&
	{\includegraphics[width=0.1800\textwidth]{figures/spurious_figure_2/1.jpg}}&
	{\includegraphics[width=0.1800\textwidth]{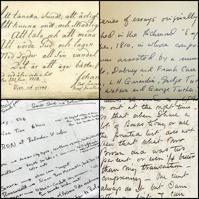}}&
	{\includegraphics[width=0.1800\textwidth]{figures/spurious_figure_2/3.jpg}}&
	{\includegraphics[width=0.1800\textwidth]{figures/spurious_figure_2/4.jpg}}\\
	\hline
	{hair spray (bend arms, 1)} & {hamper (flowers, 1)} & {hard disc (label, 1)} & {horizontal bar (gymnast, 8)} & {lighter (fire, 5)}\\
	{\includegraphics[width=0.1800\textwidth]{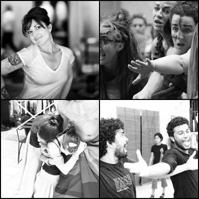}}&
	{\includegraphics[width=0.1800\textwidth]{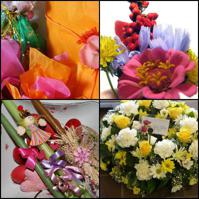}}&
	{\includegraphics[width=0.1800\textwidth]{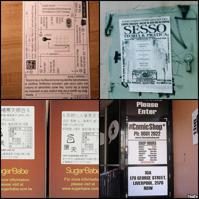}}&
	{\includegraphics[width=0.1800\textwidth]{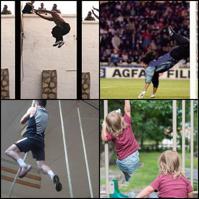}}&
	{\includegraphics[width=0.1800\textwidth]{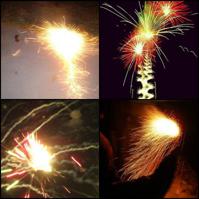}}\\
	\hline
	{miniskirt (woman, 5)} & {mortarboard (humans in suits, 3)} & {mountain bike (forest 1)} & {neck brace (humans, 3)} & {nipple (baby, 7)}\\
	{\includegraphics[width=0.1800\textwidth]{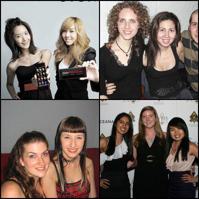}}&
	{\includegraphics[width=0.1800\textwidth]{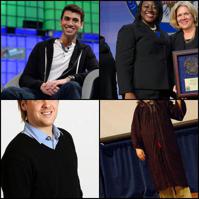}}&
	{\includegraphics[width=0.1800\textwidth]{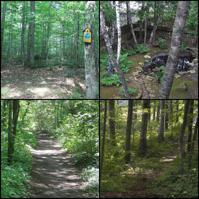}}&
	{\includegraphics[width=0.1800\textwidth]{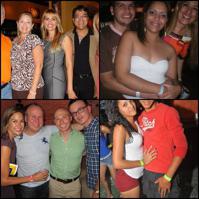}}&
	{\includegraphics[width=0.1800\textwidth]{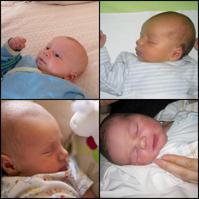}}\\
	\hline
	{obelisk (sky/clouds, 9)} & {ocarina (humans, 6)} & {padlock (wooden door, 2)} & {parallel bars (gymnasts, 3)} & {pencil box (pink/comic, 8)}\\
	{\includegraphics[width=0.1800\textwidth]{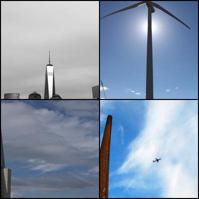}}&
	{\includegraphics[width=0.1800\textwidth]{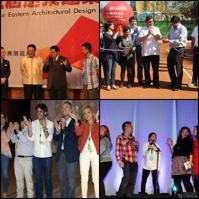}}&
	{\includegraphics[width=0.1800\textwidth]{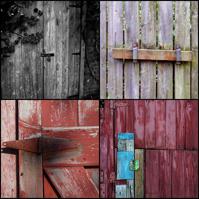}}&
	{\includegraphics[width=0.1800\textwidth]{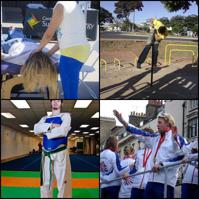}}&
	{\includegraphics[width=0.1800\textwidth]{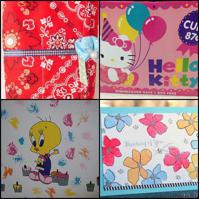}}\\
	\hline
	{pill bottle (pills, 1)} & {ping-pong ball (human/faces, 7)} & {plastic bag (twig/tree, 9)} & {plunger (humans with stick, 2)} & {pole (humans, bare arms/legs, 1)}\\
	{\includegraphics[width=0.1800\textwidth]{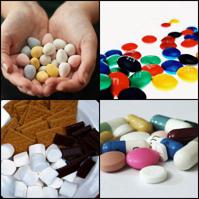}}&
	{\includegraphics[width=0.1800\textwidth]{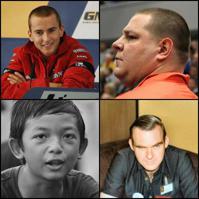}}&
	{\includegraphics[width=0.1800\textwidth]{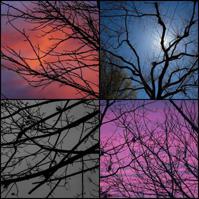}}&
	{\includegraphics[width=0.1800\textwidth]{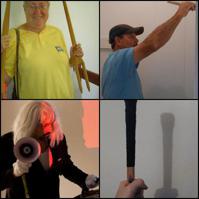}}&
	{\includegraphics[width=0.1800\textwidth]{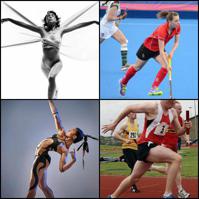}}\\
\end{tabular}

\caption{\label{fig:app_spurious_random3}Random selection of four images for classes 51-75 of our ``Spurious ImageNet'' dataset  with class label (spurious feature, NPCA component).}

\end{figure*}

\begin{figure*}
\scriptsize\begin{tabular}{c | c | c | c | c }
	{pop bottle (colorful label, 3)} & {pot (tree/bush, 8)} & {potter's wheel {\tiny (humans/bare arms, 5)}} & {puck (ice-hockey player, 1)} & {quill (hand-written text, 5)}\\
	{\includegraphics[width=0.1800\textwidth]{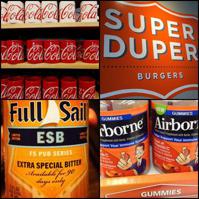}}&
	{\includegraphics[width=0.1800\textwidth]{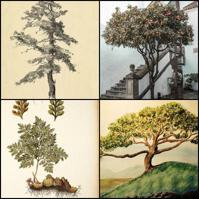}}&
	{\includegraphics[width=0.1800\textwidth]{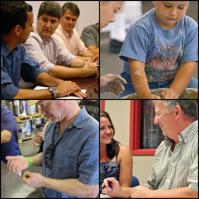}}&
	{\includegraphics[width=0.1800\textwidth]{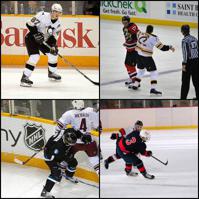}}&
	{\includegraphics[width=0.1800\textwidth]{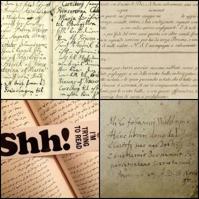}}\\
	\hline
	{racket (athlete/court, 5)} & {radio telescope (sky/hill, 4)} & {rain barrel (wooden wall, 4)} & {rotisserie (baking oven, 2)} & {rubber eraser (pen, 10)}\\
	{\includegraphics[width=0.1800\textwidth]{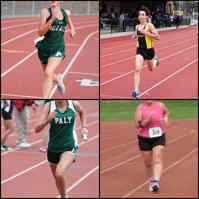}}&
	{\includegraphics[width=0.1800\textwidth]{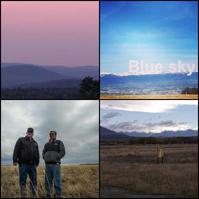}}&
	{\includegraphics[width=0.1800\textwidth]{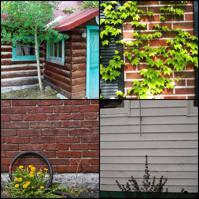}}&
	{\includegraphics[width=0.1800\textwidth]{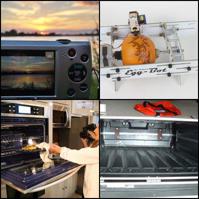}}&
	{\includegraphics[width=0.1800\textwidth]{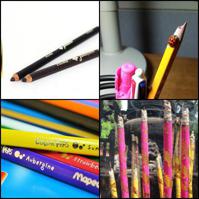}}\\
	\hline
	{safe (text, 8)} & {sax (face/dark background, 2)} & {seat belt (humans in car, 2)} & {shoe shop (humans indoor, 7)} & {shovel (wall/ground, 8)}\\
	{\includegraphics[width=0.1800\textwidth]{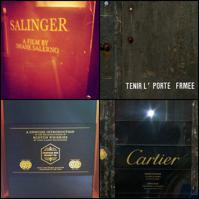}}&
	{\includegraphics[width=0.1800\textwidth]{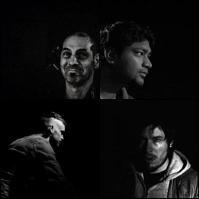}}&
	{\includegraphics[width=0.1800\textwidth]{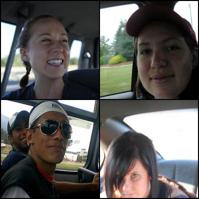}}&
	{\includegraphics[width=0.1800\textwidth]{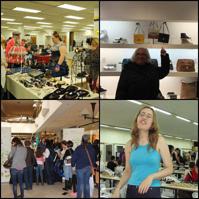}}&
	{\includegraphics[width=0.1800\textwidth]{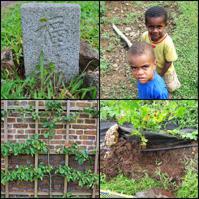}}\\
	\hline
	{shower cap (baby/humans, 4)} & {sleeping bag (text, 9)} & {snorkel (humans underwater, 1)} & {snowmobile (snowy forest, 1)} & {snowplow (snowy landscape 6)}\\
	{\includegraphics[width=0.1800\textwidth]{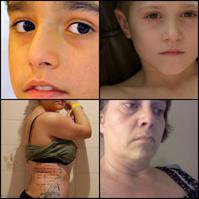}}&
	{\includegraphics[width=0.1800\textwidth]{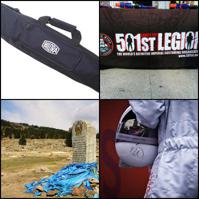}}&
	{\includegraphics[width=0.1800\textwidth]{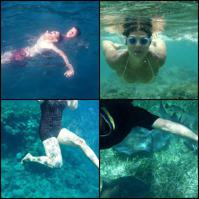}}&
	{\includegraphics[width=0.1800\textwidth]{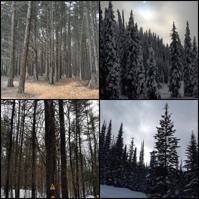}}&
	{\includegraphics[width=0.1800\textwidth]{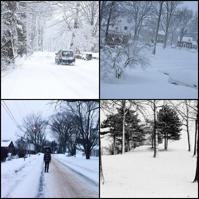}}\\
	\hline
	{steel drum (humans, 8)} & {stove (fire, 2)} & {trailer truck (sky/clouds, 5)} & {wing (clouds from above, 4)} & {cheeseburger (fries, 9)}\\
	{\includegraphics[width=0.1800\textwidth]{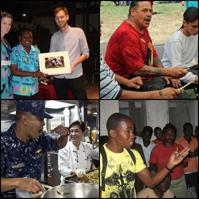}}&
	{\includegraphics[width=0.1800\textwidth]{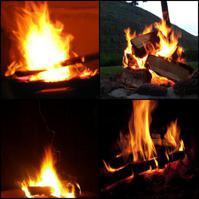}}&
	{\includegraphics[width=0.1800\textwidth]{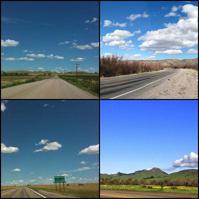}}&
	{\includegraphics[width=0.1800\textwidth]{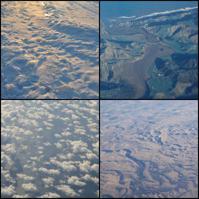}}&
	{\includegraphics[width=0.1800\textwidth]{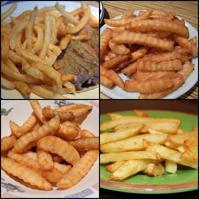}}\\
\end{tabular}
\caption{\label{fig:app_spurious_random4}Random selection of 4 images for classes 76-100 of our ``Spurious ImageNet'' dataset  with class label (spurious feature, NPCA component).}
\end{figure*}

%% file: panels_tables/dvces_flips.tex
\begin{figure*}[ht!]
    \setlength{\tabcolsep}{0.1em}
    \centering
    \normalsize
    \begin{tabular}{c c|c c|c c|c c|c c}
        \multicolumn{1}{c}{Original} &
        \multicolumn{1}{c|}{DVCE} &
        \multicolumn{1}{c}{Original} &
        \multicolumn{1}{c|}{DVCE} &
        \multicolumn{1}{c}{Original} &
        \multicolumn{1}{c|}{DVCE} &
        \multicolumn{1}{c}{Original} &
        \multicolumn{1}{c|}{DVCE} &
        \multicolumn{1}{c}{Original} &
        \multicolumn{1}{c}{DVCE} \\
        \hline
        \begin{subfigure}{0.095\textwidth}\centering     \caption*{\scriptsize \makecell{ fountain:\\ 0.36}}     \includegraphics[width=1\textwidth]{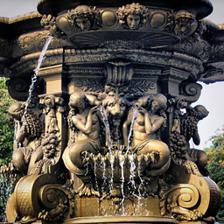}     \end{subfigure}&\begin{subfigure}{0.095\textwidth}\centering     \caption*{\scriptsize \makecell{ fireboat:\\ 0.53}}     \includegraphics[width=1\textwidth]{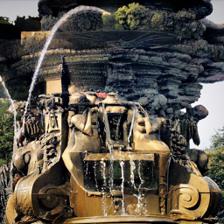}     \end{subfigure}&\begin{subfigure}{0.095\textwidth}\centering     \caption*{\scriptsize \makecell{ fountain:\\ 0.99}}     \includegraphics[width=1\textwidth]{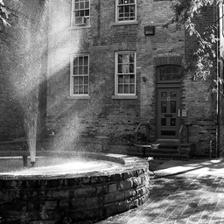}     \end{subfigure}&\begin{subfigure}{0.095\textwidth}\centering     \caption*{\scriptsize \makecell{ fireboat:\\ 0.86}}     \includegraphics[width=1\textwidth]{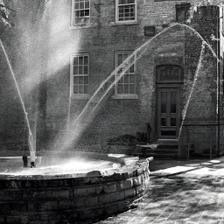}     \end{subfigure}&\begin{subfigure}{0.095\textwidth}\centering     \caption*{\scriptsize \makecell{ fountain:\\ 0.82}}     \includegraphics[width=1\textwidth]{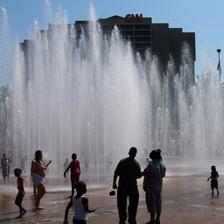}     \end{subfigure}&\begin{subfigure}{0.095\textwidth}\centering     \caption*{\scriptsize \makecell{ fireboat:\\ 0.98}}     \includegraphics[width=1\textwidth]{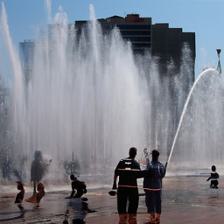}     \end{subfigure}&\begin{subfigure}{0.095\textwidth}\centering     \caption*{\scriptsize \makecell{ fountain:\\ 0.92}}     \includegraphics[width=1\textwidth]{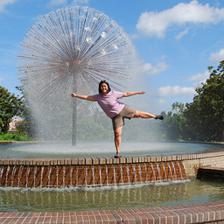}     \end{subfigure}&\begin{subfigure}{0.095\textwidth}\centering     \caption*{\scriptsize \makecell{ fireboat:\\ 0.98}}     \includegraphics[width=1\textwidth]{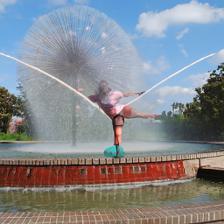}     \end{subfigure}&\begin{subfigure}{0.095\textwidth}\centering     \caption*{\scriptsize \makecell{ seashore:\\ 0.42}}     \includegraphics[width=1\textwidth]{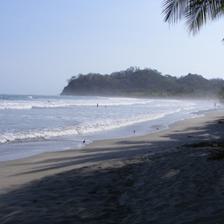}     \end{subfigure}&\begin{subfigure}{0.095\textwidth}\centering     \caption*{\scriptsize \makecell{ fireboat:\\ 1.0}}     \includegraphics[width=1\textwidth]{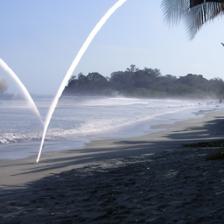}     \end{subfigure}\\\begin{subfigure}{0.095\textwidth}\centering     \caption*{\scriptsize \makecell{ stone wall:\\ 0.93}}     \includegraphics[width=1\textwidth]{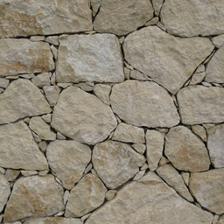}     \end{subfigure}&\begin{subfigure}{0.095\textwidth}\centering     \caption*{\scriptsize \makecell{ freight car:\\ 0.84}}     \includegraphics[width=1\textwidth]{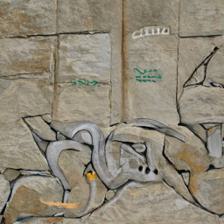}     \end{subfigure}&\begin{subfigure}{0.095\textwidth}\centering     \caption*{\scriptsize \makecell{ stone wall:\\ 0.92}}     \includegraphics[width=1\textwidth]{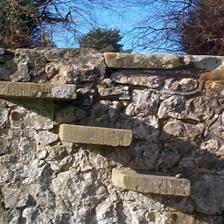}     \end{subfigure}&\begin{subfigure}{0.095\textwidth}\centering     \caption*{\scriptsize \makecell{ freight car:\\ 1.0}}     \includegraphics[width=1\textwidth]{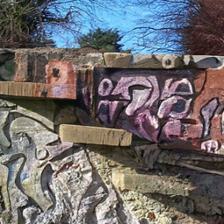}     \end{subfigure}&\begin{subfigure}{0.095\textwidth}\centering     \caption*{\scriptsize \makecell{ stone wall:\\ 0.93}}     \includegraphics[width=1\textwidth]{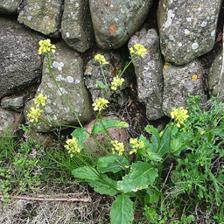}     \end{subfigure}&\begin{subfigure}{0.095\textwidth}\centering     \caption*{\scriptsize \makecell{ freight car:\\ 0.95}}     \includegraphics[width=1\textwidth]{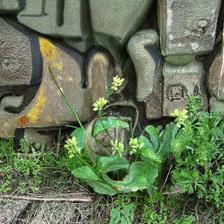}     \end{subfigure}&\begin{subfigure}{0.095\textwidth}\centering     \caption*{\scriptsize \makecell{ moving van:\\ 0.17}}     \includegraphics[width=1\textwidth]{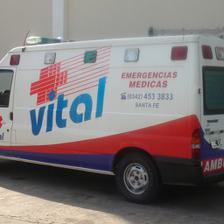}     \end{subfigure}&\begin{subfigure}{0.095\textwidth}\centering     \caption*{\scriptsize \makecell{ freight car:\\ 0.99}}     \includegraphics[width=1\textwidth]{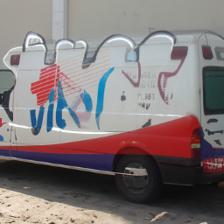}     \end{subfigure}&\begin{subfigure}{0.095\textwidth}\centering     \caption*{\scriptsize \makecell{ shopping cart:\\ 0.95}}     \includegraphics[width=1\textwidth]{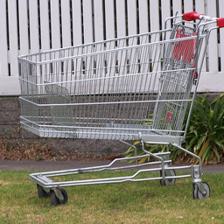}     \end{subfigure}&\begin{subfigure}{0.095\textwidth}\centering     \caption*{\scriptsize \makecell{ freight car:\\ 1.0}}     \includegraphics[width=1\textwidth]{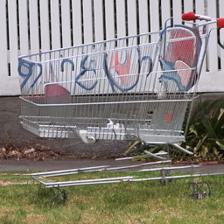}     \end{subfigure}\\\begin{subfigure}{0.095\textwidth}\centering     \caption*{\scriptsize \makecell{ geyser:\\ 0.99}}     \includegraphics[width=1\textwidth]{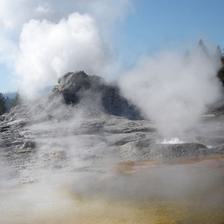}     \end{subfigure}&\begin{subfigure}{0.095\textwidth}\centering     \caption*{\scriptsize \makecell{ flagpole:\\ 1.0}}     \includegraphics[width=1\textwidth]{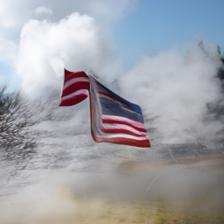}     \end{subfigure}&\begin{subfigure}{0.095\textwidth}\centering     \caption*{\scriptsize \makecell{ alp:\\ 0.33}}     \includegraphics[width=1\textwidth]{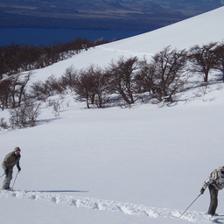}     \end{subfigure}&\begin{subfigure}{0.095\textwidth}\centering     \caption*{\scriptsize \makecell{ flagpole:\\ 1.00}}     \includegraphics[width=1\textwidth]{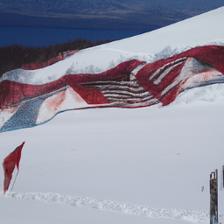}     \end{subfigure}&\begin{subfigure}{0.095\textwidth}\centering     \caption*{\scriptsize \makecell{ mosque:\\ 0.19}}     \includegraphics[width=1\textwidth]{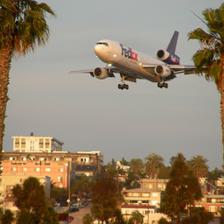}     \end{subfigure}&\begin{subfigure}{0.095\textwidth}\centering     \caption*{\scriptsize \makecell{ flagpole:\\ 1.0}}     \includegraphics[width=1\textwidth]{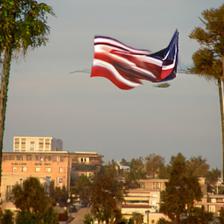}     \end{subfigure}&\begin{subfigure}{0.095\textwidth}\centering     \caption*{\scriptsize \makecell{ paddle:\\ 0.14}}     \includegraphics[width=1\textwidth]{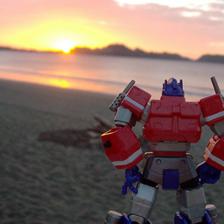}     \end{subfigure}&\begin{subfigure}{0.095\textwidth}\centering     \caption*{\scriptsize \makecell{ flagpole:\\ 0.99}}     \includegraphics[width=1\textwidth]{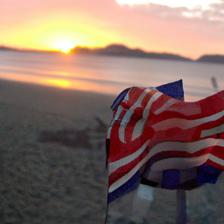}     \end{subfigure}&\begin{subfigure}{0.095\textwidth}\centering     \caption*{\scriptsize \makecell{ stone wall:\\ 0.21}}     \includegraphics[width=1\textwidth]{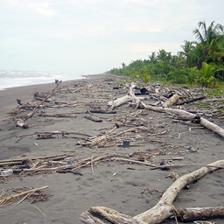}     \end{subfigure}&\begin{subfigure}{0.095\textwidth}\centering     \caption*{\scriptsize \makecell{ flagpole:\\ 1.0}}     \includegraphics[width=1\textwidth]{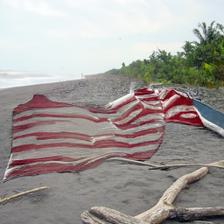}     \end{subfigure}\\\begin{subfigure}{0.095\textwidth}\centering     \caption*{\scriptsize \makecell{ car mirror:\\ 1.0}}     \includegraphics[width=1\textwidth]{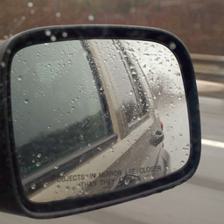}     \end{subfigure}&\begin{subfigure}{0.095\textwidth}\centering     \caption*{\scriptsize \makecell{ hard disc:\\ 0.92}}     \includegraphics[width=1\textwidth]{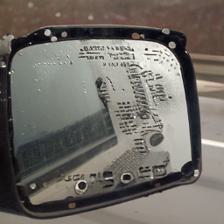}     \end{subfigure}&\begin{subfigure}{0.095\textwidth}\centering     \caption*{\scriptsize \makecell{ gas pump:\\ 0.42}}     \includegraphics[width=1\textwidth]{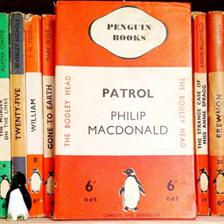}     \end{subfigure}&\begin{subfigure}{0.095\textwidth}\centering     \caption*{\scriptsize \makecell{ hard disc:\\ 0.99}}     \includegraphics[width=1\textwidth]{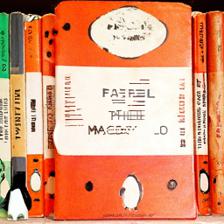}     \end{subfigure}&\begin{subfigure}{0.095\textwidth}\centering     \caption*{\scriptsize \makecell{ stone wall:\\ 0.93}}     \includegraphics[width=1\textwidth]{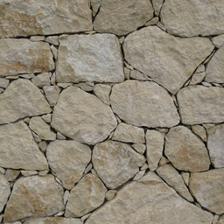}     \end{subfigure}&\begin{subfigure}{0.095\textwidth}\centering     \caption*{\scriptsize \makecell{ hard disc:\\ 0.93}}     \includegraphics[width=1\textwidth]{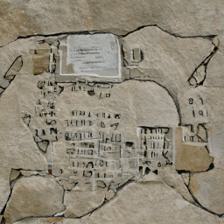}     \end{subfigure}&\begin{subfigure}{0.095\textwidth}\centering     \caption*{\scriptsize \makecell{ comic book:\\ 0.43}}     \includegraphics[width=1\textwidth]{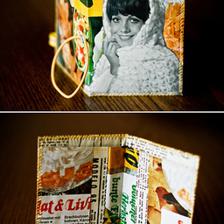}     \end{subfigure}&\begin{subfigure}{0.095\textwidth}\centering     \caption*{\scriptsize \makecell{ hard disc:\\ 1.0}}     \includegraphics[width=1\textwidth]{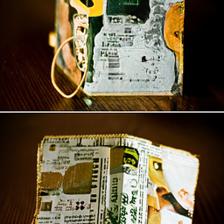}     \end{subfigure}&\begin{subfigure}{0.095\textwidth}\centering     \caption*{\scriptsize \makecell{ monitor:\\ 0.57}}     \includegraphics[width=1\textwidth]{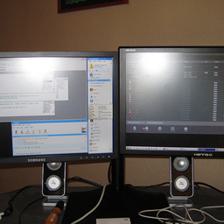}     \end{subfigure}&\begin{subfigure}{0.095\textwidth}\centering     \caption*{\scriptsize \makecell{ hard disc:\\ 0.97}}     \includegraphics[width=1\textwidth]{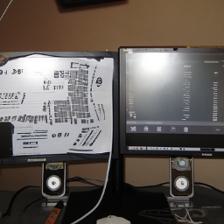}     \end{subfigure} \\
    \end{tabular}
    \caption{\label{fig:dvces_flips}Adding spurious features automatically with an adaptation of DVCEs \cite{Augustin2022Diffusion} changes the prediction of the classifier robust ResNet50. This happens, because, as has been shown qualitatively in Fig.~\ref{fig:spurious-exp} and quantitatively in Fig.~\ref{fig:barplot}, this classifier has learned to associate class ``fireboat'' with the spurious feature ``water jet'', ``freight car'' - with ``graffiti'', ``flagpole'' with a flag without the pole and mostly with ``US flag'', and ``hard disc'' - with ``label''. This again confirms that they are \textit{harmful} spurious features.}
\end{figure*}